# Foundation Models in Computational Pathology: A Review of Challenges, Opportunities, and Impact

Mohsin Bilal[1,*], Aadam[2], Manahil Raza[3], Youssef Altherwy[1], Anas Alsuhaibani[1], Abdulrahman Abduljabbar[1], Fahdah Almarshad[1], Paul Golding[4], Nasir Rajpoot[3]

[1] Information Systems Department, College of Computer Engineering and Sciences, Prince Sattam bin Abdulaziz University, Al-Kharj, Saudi Arabia
[2] Luddy School of Informatics, Computing, and Engineering, Indiana University Indianapolis, United States
[3] Tissue Image Analytics Centre, Department of Computer Science, University of Warwick, Coventry, United Kingdom
[4] Frontier AI, San Francisco Bay Area, United States
[*] Corresponding author: m.farooq@psau.edu.sa

## Abstract

From self-supervised, vision-only models to contrastive visual-language frameworks, computational pathology has rapidly evolved in recent years. Generative AI "co-pilots" now demonstrate the ability to mine subtle, sub-visual tissue cues across the cellular-to-pathology spectrum, generate comprehensive reports, and respond to complex user queries. The scale of data has surged dramatically, growing from tens to millions of multi-gigapixel tissue images, while the number of trainable parameters in these models has risen to several billion.

The critical question remains: how will this new wave of generative and multi-purpose AI transform clinical diagnostics? In this article, we explore the true potential of these innovations and their integration into clinical practice. We review the rapid progress of foundation models in pathology, clarify their applications and significance. More precisely, we examine the very definition of foundational models, identifying what makes them foundational, general, or multipurpose, and assess their impact on computational pathology. Additionally, we address the unique challenges associated with their development and evaluation. These models have demonstrated exceptional predictive and generative capabilities, but establishing global benchmarks is crucial to enhancing evaluation standards and fostering their widespread clinical adoption.

In computational pathology, the broader impact of frontier AI ultimately depends on widespread adoption and societal acceptance. While direct public exposure is not strictly necessary, it remains a powerful tool for dispelling misconceptions, building trust, and securing regulatory support.

## 1. Introduction

Foundation models (Bommasani et al. 2022; Alfasly et al. 2023; Z. Chen et al. 2024; OpenAI 2023b; 2023a; Kirillov et al. 2023; Radford et al. 2021) mark the dawn of a new AI era[a] with the power to transform computational pathology. Just as deep learning once disrupted the field (A. H. Song et al. 2023; Perez-Lopez et al. 2024), foundation models now promise to do the same, offering unprecedented capabilities in understanding, predicting, and diagnosing disease (Y. Xu et al. 2024; M. Y. Lu, Chen, Williamson, Chen, Zhao, et al. 2024). This review examines how these models could transform medical discovery, moving beyond merely augmenting human expertise to fundamentally reshaping it.

Foundation models, heralded as engines of "generalist AI", promise to unify knowledge across specialties like tissue histopathology and oncology (Vorontsov et al. 2024; R. J. Chen et al. 2024; M. Y. Lu, Chen, Williamson, Chen, Liang, et al. 2024). While they offer unprecedented synthesis of cross-domain insights, their ability to match specialists' deep understanding remains uncertain. As computational pathology adopts these models, we must evaluate two competing needs: adapting models for specific pathology tasks while maintaining their broad applicability.

Balancing these approaches unlocks their synergy: specialists ensure diagnostic precision while generalists accelerate innovation by identifying cross-disciplinary patterns. Building diagnostic foundation models requires rigorous attention to the nuances of data curation, model training, and validation (Bommasani et al. 2022). The stakes are high: biased or poorly curated datasets could lead to models that overgeneralize, potentially missing critical diagnostic subtleties. As we push towards clinical adoption (Saillard et al. 2023; Bilal, Tsang, et al. 2023; Campanella et al. 2019; Graham, Minhas, et al. 2023), it's essential to balance enthusiasm with caution, ensuring that these powerful tools are developed with the precision and care demanded. This review examines how foundation models can transform computational pathology, balancing their innovation potential against the imperatives of clinical safety and efficacy.

[a] The era of AI has just begun. Bill Gates, March 2023

These models generalize across diverse tasks, offering potential in diagnosing complex diseases and in predicting treatment responses by combining information from large-scale gigapixel histopathology images, oncology data, and genomics (Zimmermann et al. 2024; Sun, Zhu, et al. 2024; Jaume, Vaidya, et al. 2024). This holistic view was previously unattainable; however, these models pose dual challenges: architecting algorithms that can process pathology's complex high-dimensional datasets and developing rigorous frameworks to evaluate their performance.

The transition from task-specific (Saillard et al. 2023; Bilal, Tsang, et al. 2023; Campanella et al. 2019; Graham, Minhas, et al. 2023) and multi-task (Graham, Vu, et al. 2023) models towards generative (M. Y. Lu, Chen, Williamson, Chen, Zhao, et al. 2024; Sun, Zhu, et al. 2024; Sun, Wu, et al. 2024), task-agnostic (R. J. Chen et al. 2024; Zimmermann et al. 2024; Vorontsov et al. 2024) ones marks a paradigm shift, raising critical questions: How do these new models compare to specialized counterparts when evaluated fairly and rigorously? Are we genuinely capturing diagnostically useful subtleties of pathology, or are we risking a loss of precision in a quest for breadth? While significant progress has been made in striking this balance for analytical and predictive tasks, new problems have arisen in reasoning, especially in adhering to safety and devising benchmarks.

Processing gigapixel whole slide images (WSIs) poses a fundamental challenge: how to maintain contextual awareness across multiple scales while aggregating information from diverse spatial locations (Bilal, Jewsbury, et al. 2023; S. Chen et al. 2024; Campanella, Chen, et al. 2024). This multi-scale integration affects both computational costs and model performance. Beyond technical hurdles, clinical adoption faces practical barriers: workflow integration challenges, digitization costs, and the demands of scalable storage and computing infrastructure (Mayall et al. 2023).

Two recent surveys in computational pathology examine foundation and vision-language models by detailing their tools, datasets, and training methodologies. Chanda et al. (Chanda et al. 2024) provide an in-depth overview of these technological advances, albeit with limited discussion on the associated risks, clinical challenges, opportunities, and broader impact. In contrast, Ochi et al. (Mieko, Komura, and Ishikawa 2025) focus on five vision-only and five image-language aligned models, emphasizing both their applications and the obstacles encountered in translating these models into clinical practice. Collectively, the reviews underscore the need for a more rigorous evaluation of the opportunities, risks, and practical challenges that healthcare professionals face when adopting these emerging technologies.

Critical evaluations of recent studies reveal a mix of promising advances and persistent issues. Despite impressive successes in prediction, key barriers persist: limited data diversity, opaque model decisions, and poor generalization across populations. A critical question emerges: Is our progress toward clinical adoption truly revolutionary, or have we reached a plateau? This review comprehensively examines computational pathology's core challenges, from WSI context integration to clinical AI deployment. We examine emerging solutions while candidly assessing progress and persistent gaps.

## 2. Foundation Models: Definition, Importance, and Technical Foundations

The evolution of machine learning over the past three decades can be understood through three distinct phases, each marked by the "emergence" of new paradigms and the "homogenization" of techniques within the field (Bommasani et al. 2022). Figure 1 illustrates these evolution phases.

**Phase 1: Machine Learning**

In the early stages of machine learning, the focus shifted from explicitly programming how to solve a task to **inductive learning**—learning by examples. This period relied upon handcrafting of features, particularly for complex tasks in natural language processing and computer vision. The primary goal was to predict unseen ("out of distribution") examples based on patterns learned from known examples and their **manually designed representations**.

**Phase 2: Deep Learning**

The emergence of deep learning marked a fundamental shift toward data-driven approaches in artificial intelligence (AI). Instead of relying on manual feature engineering, systems could now **automatically** extract high-level features—known as **representation learning**—directly from raw data like image pixels. Convolutional neural networks (ConvNets) emerged as the leading architectural paradigm, bringing unprecedented standardization to model design. This architectural convergence, combined with access to larger datasets and GPU-accelerated parallel computing, drove dramatic performance improvements across computer vision applications. This success embodied Sutton's "bitter lesson": approaches that leveraged computation and large amounts of data consistently outperformed carefully crafted, domain-specific solutions, setting the stage for the coming phase of foundation models.

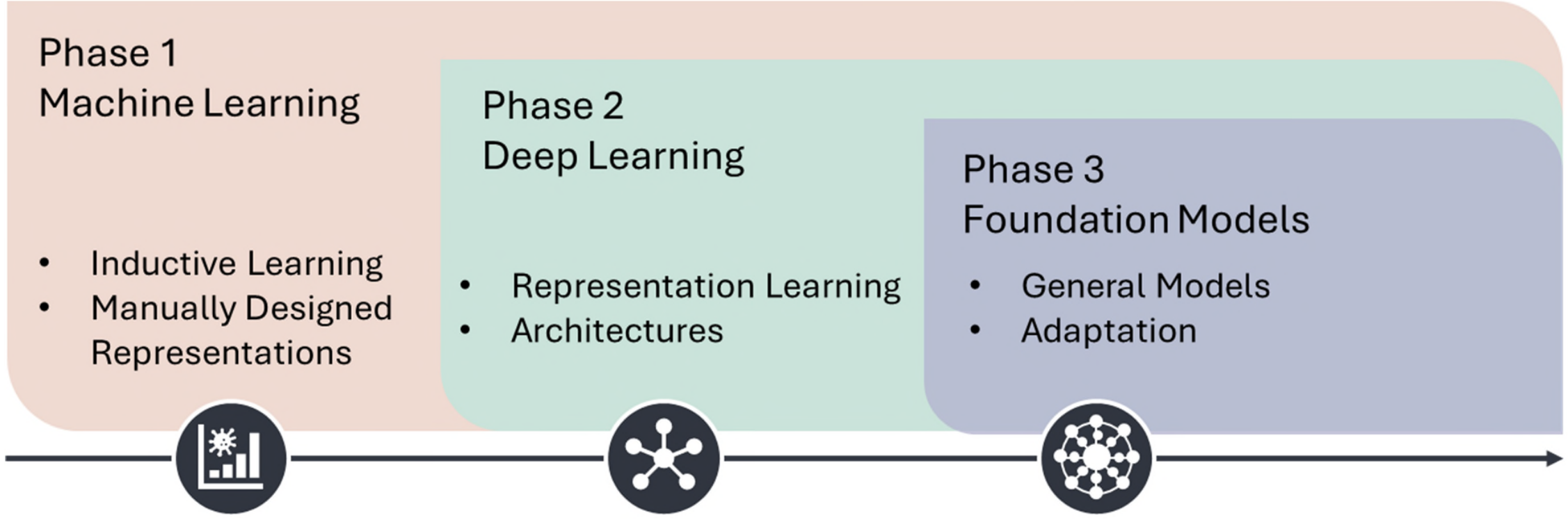

*Figure 1. Three marked phases of Machine Learning evolution: from machine learning, deep learning to foundation models.*

**Phase 3: Foundation Models**

The latest shift in machine learning marks an even deeper form of homogenization. While the deep learning era saw convergence around similar architectures like ConvNets, foundation models introduce homogenization at the model level itself—multiple applications now build upon the same pre-trained models rather than training separate models from scratch. These foundation models are designed with inherent generality, allowing them to adapt to diverse downstream tasks through techniques like fine-tuning. This capability is enabled by the unprecedented scale of training data and advances in self-supervised learning. Foundation models typically leverage transformer architectures (Vaswani et al. 2023), which through techniques such as autoregressive modeling and masked language modeling (as pioneered by BERT (Devlin et al. 2019)), have demonstrated remarkable ability to transfer knowledge across tasks. This versatility represents a significant inflection point: a single pre-trained model can serve as the foundation for a wide array of applications, fundamentally changing how we approach AI development.

The implications of this deeper homogenization are profound. On one hand, foundation models may offer substantial benefits in domains like computational pathology, particularly for scenarios involving rare diseases, low-cost data acquisition, or limited task-specific data. However, this shared foundation also introduces systemic risk, any flaws or biases in the foundation model can propagate to all downstream applications that build upon it. The uncertainty surrounding both the capabilities and limitations of foundation models makes their widespread adoption a complex challenge that requires careful consideration. Mitigating these inherited risks has become a central challenge for the ethical and safe development of AI in this field.

## 2.1. Definition and Characteristics of Foundation Models

Foundation models represent a fundamental shift in AI development, earning their name by serving as true foundations upon which diverse AI applications (Bommasani et al. 2022)can be built. These models are characterized by three key attributes: scale, self-supervision, and adaptability. Trained on massive, diverse datasets—often spanning multiple modalities including text, images, speech, structured data, and 3D signals—using self-supervised learning at unprecedented scale, they exhibit remarkable emergent properties. Despite being trained on seemingly simple tasks like next-token prediction, they develop capabilities that transcend their training objectives, demonstrating competence across **a wide range of downstream tasks** that weren't explicitly part of their training. This emergence of new capabilities, coupled with evidence that performance continues to improve with scale (the "scaling hypothesis"), has driven a persistent push toward larger datasets and model sizes.

The term "foundation model" was coined in August 2021 by Stanford's Institute for Human-Centered Artificial Intelligence's Center for Research on Foundation Models, marking the recognition of these models' transformative impact on both AI technology and society—i.e. they are foundational in very broad socio-technical terms, which has implications in terms of model reliability, safety and ethics (Bommasani et al. 2022). This paradigm emerged first in natural language processing with models like BERT (Devlin et al. 2019) and GPT-3 (Brown et al. 2020), before expanding to computer vision through innovations like Contrastive Language-Image Pretraining (CLIP (Radford et al. 2021))and the Segment Anything Model (SAM (Kirillov et al. 2023)). Now, **multimodal** foundation models can seamlessly operate across modalities, understanding relationships between text and images, generating images from text descriptions, or even processing medical imaging data alongside clinical notes. This versatility has begun to influence specialized domains like computational pathology, where the analysis of human tissue images could benefit from these powerful, adaptable architectures.

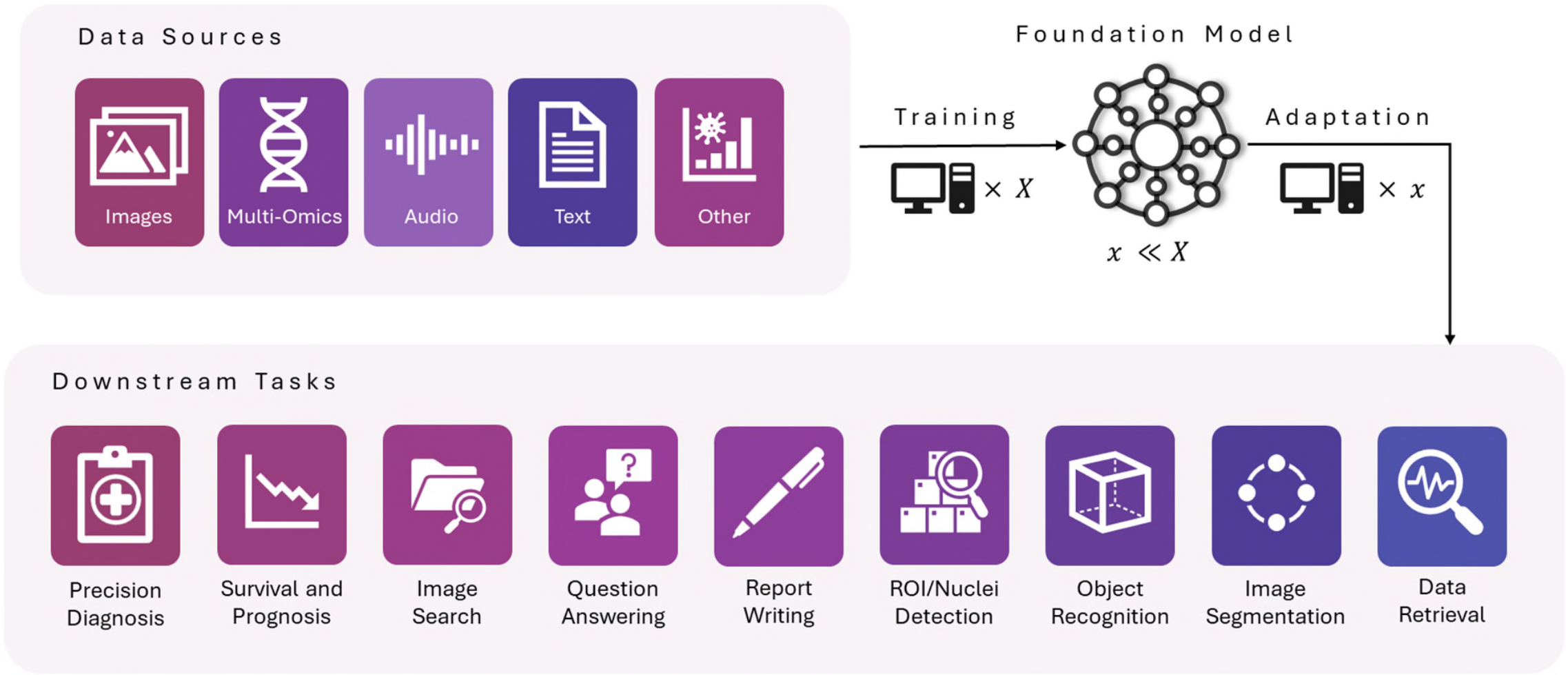

*Figure 2. Pathology (Multimodal) Foundation Models – Data sources, compute-extensive training as number of GPUs needed is much larger ($X$) than the adaptation (which can be done by a single modern GPU, $x$) and example downstream tasks.*

Traditional approaches in computational pathology have followed two main paradigms: classical predictive AI and contrastive learning methods. These approaches typically operate on small to moderate-sized datasets, using a combination of learning strategies tailored to different diagnostic and prognostic tasks (A. H. Song et al. 2023; Bilal et al. 2022). For analyzing individual image patches or tiles, fully supervised learning has been the dominant approach, requiring detailed annotations at the pixel level. When dealing with WSIs (A. H. Song et al. 2023; Bilal, Jewsbury, et al. 2023)—which are too large to process directly—weakly supervised learning techniques are often employed, combining predictions from multiple patches and requiring only slide-level labels. These methods, while effective for specific tasks, require careful engineering and task-specific training data, contrasting sharply with the more general-purpose nature of foundation models.

## 2.2. Pathology Foundation Models

The evolution of foundation models in histopathology reflects a rapid progression across three key dimensions: modality, scale, and learning approach. Initially developed for single-modality analysis of histopathology images, these models have undergone significant expansion in their capabilities. The progression from single-modality to dual-modality (combining images with text annotations) and ultimately to true multimodal integration (incorporating images, text, RNA-seq data, and immunohistochemistry findings) reflects the field's recognition that comprehensive disease understanding requires synthesizing multiple data types. This multimodal approach is proving essential for enhancing diagnostic accuracy and advancing precision medicine.

In parallel, the scale of training data has increased dramatically, with modern systems leveraging hundreds of thousands to millions of WSIs —a scale previously unprecedented in computational pathology. This expansion in data scale has been matched by evolution in learning approaches: from traditional classification models to more sophisticated techniques like contrastive learning, self-supervised learning, and most recently, generative models capable of producing synthetic pathology images and annotations. Figure 2 illustrates four key component of pathology foundation models – data sources and modalities, training, adaptation, their compute need, and exemplary downstream tasks.

The foundational nature of these models presents both unprecedented opportunities and significant challenges for the field of pathology. Their ability to adapt to diverse tasks could democratize advanced pathology analysis, particularly benefiting regions with limited access to specialist expertise. However, their widespread adoption also raises critical concerns about safety, reliability, and ethical deployment. Given that these models may influence critical diagnostic decisions, ensuring their robustness, avoiding biases, and maintaining transparency in their decision-making processes become increasingly paramount. The challenge lies in balancing their transformative potential with responsible development and deployment practices that prioritize patient safety and ethical considerations.

Table 1 present a review of **40 recent foundation models in pathology**, highlighting scale, data modalities, and various downstream tasks these have been adapted to. These models represent four distinct architectural paradigms: (1) image-only models focusing on visual feature learning, (2) multi-stain models incorporating different microscopy techniques, (3) cross-modal models that align images with text descriptions, and (4) generative multi-modal models capable of producing text, analyzing images, and sometimes integrating molecular data. While earlier models primarily used **contrastive** learning for representation learning, recent approaches increasingly adopt **generative** architectures, enabling more interactive and interpretable outputs like automated reporting and diagnostic reasoning.

*Table 1. Pathology Foundation Models—The scale and modalities and general, multipurpose or generative potential.*

| Model | # WSIs | Text | Molecular | MultiTissue | MultiResolution | MultiStain | Multiscanner | # Task | Classification | Segmentation | Detection | WSI | Cell | Generative |
|---|---|---|---|---|---|---|---|---|---|---|---|---|---|---|
| COBRA | 3048 | No | No | 6 | Yes | No | Yes | 15 | Yes | No | No | Yes | No | No |
| Phikon-v2* | 58359 | No | No | 30 | Yes | Limited | Yes | 8 | Yes | Yes | No | Yes | Yes | No |
| TissueConcepts | 7042 | No | No | 14 | Yes | No | Yes | 16 | Yes | Yes | Yes | Yes | Yes | No |
| HIBOU* | 1141581 | No | No | 12 | Yes | Limited | Yes | 12 | Yes | Yes | No | Yes | Yes | No |
| Virchow2 & 2G | 3,134,922 | No | No | 25 | Yes | Limited | Yes | 37 | Yes | Yes | No | Yes | Yes | No |
| OmniScreen | 30511 | No | No | 27 | Yes | No | Yes | 70 | Yes | No | No | Yes | No | No |
| H-Optimus-0* | 500000 | No | No | 32 | Yes | No | Yes | 11 | Yes | Yes | No | Yes | Yes | No |
| BEPH | 11760 | No | No | 32 | No | No | Yes | 11 | Yes | No | No | Yes | No | No |
| Prov-GigaPathV* | 171189 | No | No | 31 | Yes | Limited | Yes | 26 | Yes | Yes | No | Yes | Yes | No |
| Kaiko-ai* | 29000 | No | No | 32 | Yes | No | Yes | 8 | Yes | Yes | No | Yes | Yes | No |
| UNI* | 100000 | No | No | 20 | Yes | No | Yes | 34 | Yes | Yes | No | Yes | Yes | No |
| Virchow | 1,488,550 | No | No | 17 | Yes | No | Yes | 33 | Yes | Yes | No | Yes | Yes | No |
| BROW | 11,206 | No | No | 6 | Yes | No | Yes | 11 | Yes | Yes | No | Yes | Yes | No |
| Phikon* | 6,093 | No | No | 16 | Yes | No | Yes | 17 | Yes | Yes | No | Yes | Yes | No |
| Lunit* | 36,666 | No | No | 32 | Yes | No | Yes | 8 | Yes | Yes | No | Yes | Yes | No |
| HIPT | 10,678 | No | No | 33 | Yes | No | Yes | 9 | Yes | No | No | Yes | No | No |
| CtransPath | 32,320 | No | No | 32 | No | No | Yes | 11 | Yes | Yes | Yes | Yes | Yes | No |
| HistGen_v | 55000 | No | No | 60 | Yes | No | Yes | 10 | Yes | No | No | Yes | No | No |
| REMEDIS | 29018 | No | No | 32 | No | No | Yes | 15 | Yes | No | No | Yes | No | No |
| Madeleine | 16281 | No | No | 1 | No | Yes | Yes | 21 | Yes | No | No | Yes | No | No |
| PathoDuet | 14896 | No | No | 32 | Yes | Yes | Yes | 14 | Yes | No | No | Yes | No | No |
| RudolfV | 133,998 | No | No | 58 | Yes | Yes | Yes | 50 | Yes | Yes | No | Yes | Yes | No |
| PLUTO | 158852 | No | No | 28 | Yes | Yes | Yes | 13 | Yes | Yes | No | Yes | Yes | No |
| CHIEF | 60530 | Yes | No | 19 | No | No | Yes | 27 | Yes | No | No | Yes | No | No |
| THREADS | 47171 | No | Yes | 39 | No | Yes | Yes | 54 | Yes | No | No | Yes | No | No |
| MUSK | 33000 | Yes | No | 33 | Yes | Yes | Yes | 52 | Yes | Yes | No | Yes | No | Visual question answering |
| TITAN | 335645 | Yes | No | 20 | Yes | Limited | Yes | 61 | Yes | No | No | Yes | No | Report generation |
| SlideChat | 4181 | Yes | No | 10 | Yes | No | Yes | 22 | Yes | No | No | Yes | No | WSI-Language assistant |
| PMPRG | 5195 | Yes | No | 2 | No | No |  | 5 | Yes | No | No | Yes | No | Multi-organ report |
| mSTAR | 14621 | Yes | Yes | 32 | Yes | No | Yes | 43 | Yes | No | No | Yes | No | Multimodal report |
| HistGen | 7800 | Yes | No | 32 | Yes | No | Yes | 10 | Yes | No | No | Yes | No | Report generation |
| PathAlign | 350855 | Yes | No | 32 | Yes | Limited | Yes | 10 | Yes | No | No | Yes | No | Report generation |
| PathGen | 7300 | Yes | No | 32 | No | No | Yes | 18 | Yes | No | No | Yes | No | WSI-Language assistant |
| HistoGPT | 15129 | Yes | No | 1 | Yes | No | Yes | 9 | Yes | No | No | Yes | No | WSI-Language assistant |
| PathChat | 999202 | Yes | No | 20 | Yes | Yes | Yes | 54 | Yes | No | No | No | Yes | AI assistant |
| PRISM | 587196 | Yes | No | 17 | No | Yes | Yes | 40 | Yes | No | No | Yes | No | Report generation |
| PathAsst | 180000 | Yes | No | 32 | Yes | Yes | Yes | 5 | Yes | Yes | Yes | No | Yes | AI assistant |
| Prov-GigaPathMM | 171189 | Yes | No | 31 | Yes | Limited | Yes | 26 | Yes | Yes | No | Yes | Yes | Report generation |
| CONCH | 1170000 | Yes | No | 19 | No | Yes | Yes | 14 | Yes | Yes | No | Yes | No | Captioning |
| PLIP | 28414 | Yes | No | 32 | Yes | Yes | Yes | 26 | Yes | No | No | No | No | Captioning |

In terms of capability, we distinguish between "General" and “Multipurpose” as follows, for visions only models. General typically refers to a model's ability to:

- Multi-tissue—work across tissue types (from multiple organs) without specialization

- Multi-resolution—handle various objective magnification levels
- Multi-scanner—process images from different scanner vendors
- Multi-staining—maintain performance across varied staining protocols and types (e.g. IHC and other immunostaining)

"Multipurpose" typically indicates the model's ability to:

- Perform different types of tasks (classification, segmentation, detection)
- Support various downstream applications by adaptation or minimal fine-tuning
- Handle different levels of analysis (cell-level, tissue-level, whole-slide-level, and patient-level)

Foundation models demonstrate extracting meaningful features without task-specific training, though we are now witnessing semi and fully supervised learning framework too. The multimodal foundation models which leverage vision only models in addition to learning WSIs-paired textual, molecular or genetic data. The co-pilots leverage multimodal foundation models to exhibit generative and interactive capabilities.

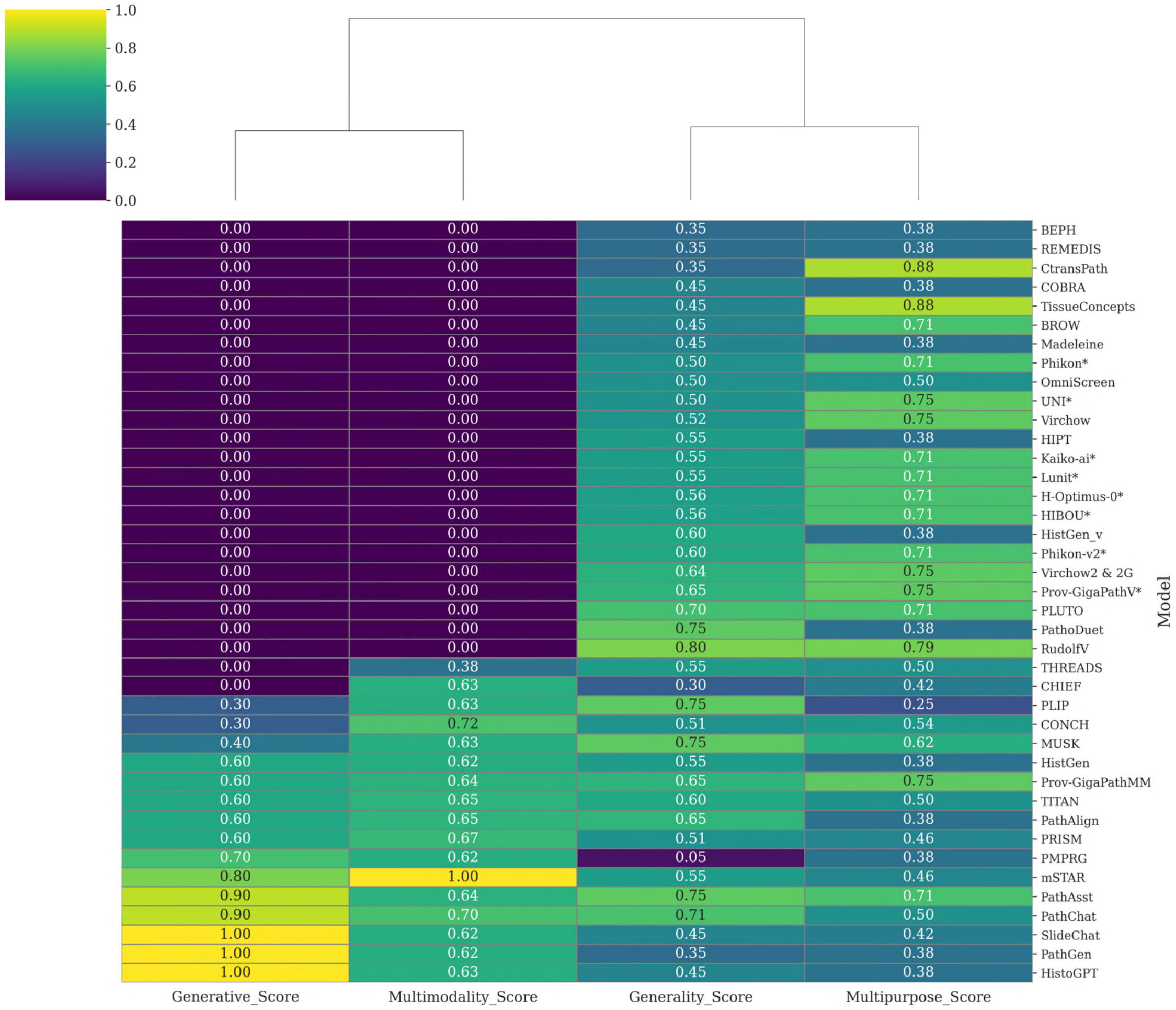

*Figure 3. Cluster heatmap of foundation models presenting multipurpose, generality, generative and multimodality scores. Multimodality score is sum of 0.2×WSIs, 0.5×Text (1/0), 0.3×Molecular. Generative score, assigned according to a lookup table: for example, "WSI-language assistant" → 1.0; "report generation" → 0.5; "no" → 0.0. Seaborn's clustermap clusters models (rows) and scores (columns) by similarity, visualizing scores from 0 to 1 in a heatmap.*

Analyzing WSIs of human tissue is central to computational pathology, which explains why early foundation model efforts focused solely on vision-based or image-only approaches. Analyzing WSIs of human tissue is central to computational pathology, which explains why early foundation model efforts focused solely on vision-based or image-only approaches. These models tackle a wide range of critical diagnostic tasks: from basic tissue classification (distinguishing tumor from normal tissue) to complex cellular analysis (identifying specific immune cell types in the tumor microenvironment), to clinically relevant predictions (cancer subtyping, biomarker status, and patient survival). The most recent and effective image-only foundation models are **Virchow2** (Zimmermann et al. 2024)**, Virchow2G** (Zimmermann et al. 2024)**, Phikon-v2** (Filiot et al. 2024)**, UNI** (R. J. Chen et al. 2024)**, Virchow** (Vorontsov et al. 2024)**, H-Optimus-0** (Saillard et al. 2024)**,** and **TissueConcepts** (Nicke et al. 2024) demonstrating strong performance across 37, 37, 8, 34, 33, 11, and 16 tasks, respectively.

For example, **Virchow** (Vorontsov et al. 2024) can simultaneously perform tumor detection, grade assessment, and molecular biomarker prediction, while **UNI**[13] can classify over 100 cancer types and perform nuclei segmentation across 20 different tissue types. **HIPT** (R. J. Chen et al. 2022), **BROW** (Y. Wu et al. 2023a), **Kaiko-ai** (ai et al. 2024) and **CtransPath** (Xiyue Wang et al. 2022a) have been outperformed by the aforementioned in terms of data scale, task coverage, and innovative use cases. **Phikon** (Filiot et al. 2023) has been trained on 40 million image tiles from TCGA cohorts of approximately 6000 WSIs used for training the foundation model. Similarly, BEPH (Z. Yang et al. 2024) was trained on over 11 million image tiles derived from TCGA cohorts of approximately 12000 WSIs. Figure 3 presents a cluster heatmap illustrating the multimodality, generative, generality, and multipurpose scores for each foundation model. General-Purpose (generality) score is an average of multi-tissue, multi-resolution, multi-staining, multi-scanner, and scaled number of WSIs scores. Multipurpose score is an average of six scores: scaled number of tasks, and five Boolean values, one each for classification, segmentation, and detection, WSI level, and cell level tasks.

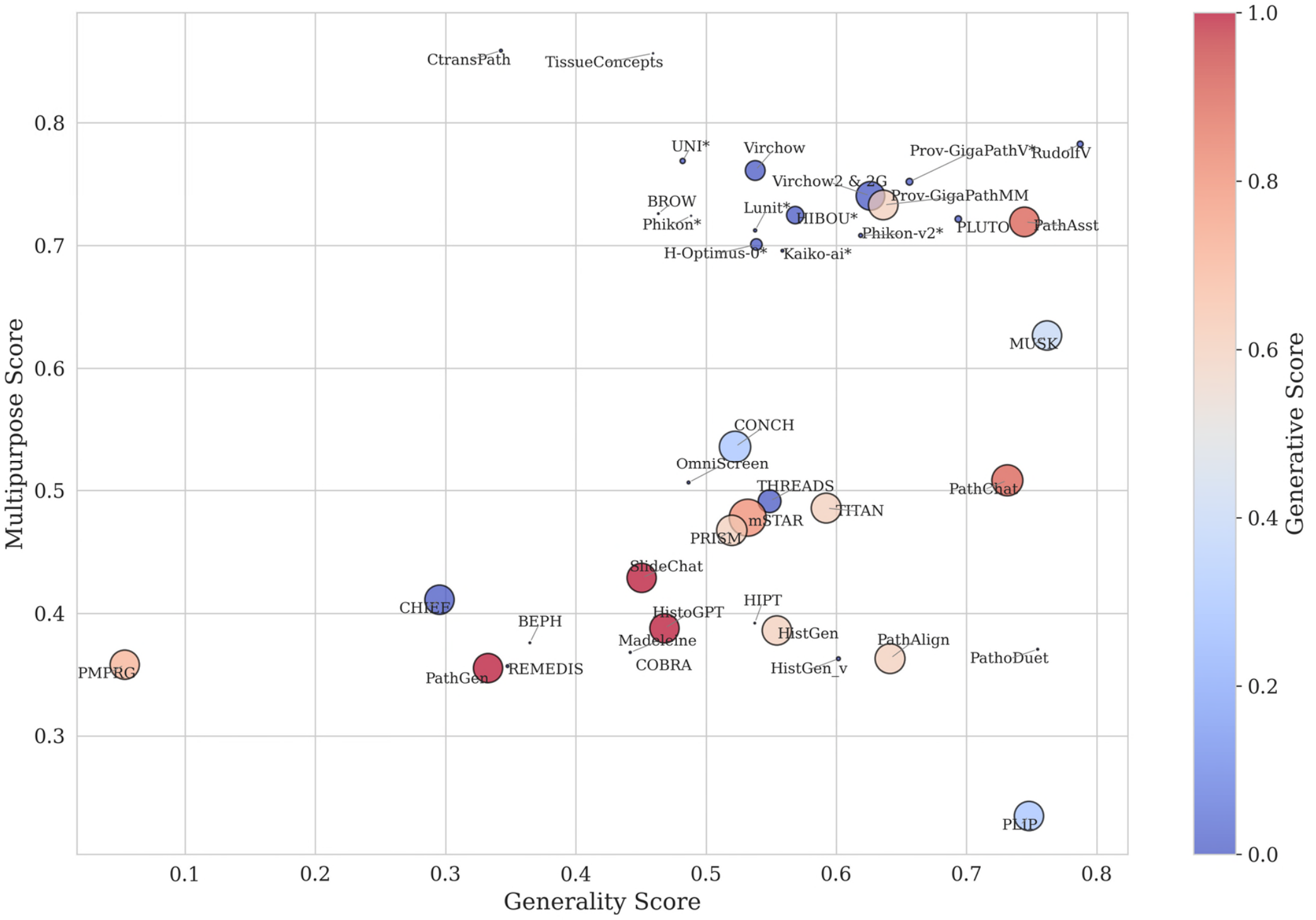

*Figure 4. Foundation models scatter plot positioning each model in terms of its generality score (x-axis), multipurpose score (y-axis), generative score (color bar), and multimodality score or a fallback based on #WSIs (bubble size).*

**UNI** (R. J. Chen et al. 2024), trained on 100,000 WSIs, is the most reused foundation model. **Virchow2** (Zimmermann et al. 2024) and **Virchow2G** (Zimmermann et al. 2024) were trained on the largest dataset of 3.1 million WSIs, with **Virchow2G** (Zimmermann et al. 2024) being the largest foundation model, featuring 1.85 billion trainable parameters. Utilizing Virchow2 (Zimmermann et al. 2024) for image tile representation—pretrained on 3 million WSIs—**OmniScreen** (Y. K. Wang et al. 2024) was further trained on new data from 30,511 patients with available NGS-based labels. This model simultaneously predicts 1,228 genomic biomarkers across 70 human cancers. **Phikon-v2** (Filiot et al. 2024) has been evaluated on 8 WSI-level tasks and shares an interesting finding to not set aside specialized models, e.g. for tasks like microsatellite instability. **TissueConcepts** (Nicke et al. 2024)**,** a unique and only **supervised** foundation model, presents an alternative approach to generality – adopting multi-tasks learning instead of task-agnostic self-supervision to demonstrate excellent performance on 16 tasks related to classification, detection, & segmentation. In addition to these large-scale models, COBRA (Lenz et al. 2024) represents an innovative approach to slide-level representation learning. Utilizing a dataset of only 3,048 WSIs and leveraging a self-supervised contrastive learning framework, COBRA (Lenz et al. 2024), excels in 15 downstream biomarker and mutation prediction tasks. COBRA (Lenz et al. 2024) is model-agnostic, meaning it can integrate patch embeddings from models like Virchow2[19] and UNI[13] to enhance their slide-level performance without additional fine-tuning. Figure 4 shows a scatter plot positioning each foundation model by its multipurpose and generality scores, with circles colored according to their generative scores.

**UNI** (R. J. Chen et al. 2024)**,** trained on 100,000 WSIs across 20 tissue types, represents a significant dataset milestone in pathology AI. Several subsequent large models, including **PathChat** (M. Y. Lu, Chen, Williamson, Chen, Zhao, et al. 2024), **mSTAR** (Y. Xu et al. 2024), and **HistoGPT** (Tran et al. 2024), have trained on datasets of similar scale (the 100,000 WSIs of **UNI** (R. J. Chen et al. 2024)), demonstrating the emergence of a common benchmark for large-scale pathology model training. **Virchow2** (Zimmermann et al. 2024) and **Virchow2G** (Zimmermann et al. 2024) represent a significant scaling leap, trained on the largest pathology dataset to date of 3.1 million WSIs from over 225,000 patients. While the 1.85 billion parameters of **Virchow2G** (Zimmermann et al. 2024) mark a milestone for pathology models, it remains relatively modest compared to general-purpose vision-language models like **GPT-4V** (estimated >1 trillion parameters) or PaLM-E (562 billion parameters). This difference in scale reflects the specialized nature of pathology models - they achieve strong performance with fewer parameters by focusing on a specific visual domain, unlike general VLMs that must handle arbitrary real-world images. Even compared to medical-specific VLMs like **Med-PaLM 2** (approximately 540 billion parameters), pathology models remain more compact while achieving expert-level performance in their domain. This efficiency suggests that domain-specific visual patterns in pathology, while complex, can be effectively learned with more focused architectures.

**Eleven** of the listed models are multimodal and generative. For instance, **mSTAR** (Y. Xu et al. 2024) integrates RNA-seq data with images and text, excelling in 32 tasks, and **THREADS** (Vaidya et al. 2025) integrates bulk RNA-seq and DNA, excelling in 54 oncology tasks. Models like **HistGen** (Zhengrui Guo et al. 2024)**, PathAlign** (F. Ahmed et al. 2024)**, PathGen** (Sun, Zhang, et al. 2024)**, HistoGPT** (Tran et al. 2024)**, PathChat** (M. Y. Lu, Chen, Williamson, Chen, Zhao, et al. 2024)**, slideChat** (Y. Chen et al. 2024)**, PRISM** (Shaikovski et al. 2024)**, PathAsst** (Sun, Zhu, et al. 2024)**, TITAN** (T. Ding et al. 2024), and **MUSK** (Xiang et al. 2025) operate on both image and text modalities, performing across 10, 5, 18, 5, 54, 22, 40, 8, 61, and 52 tasks, respectively. Notably, **TITAN** (T. Ding et al. 2024)**, PathChat** (M. Y. Lu, Chen, Williamson, Chen, Zhao, et al. 2024), **THREADS** (Vaidya et al. 2025)**,** and **MUSK** (Xiang et al. 2025) stand out with their extensive coverage of 61, 54, 54 and 52 tasks, **PRISM** (Shaikovski et al. 2024) excels in slide-level analysis across 40 tasks. **TITAN** (T. Ding et al. 2024) demonstrates strength in tasks like zero-shot classification, pathology report generation, and rare cancer retrieval. **PathChat** (M. Y. Lu, Chen, Williamson, Chen, Zhao, et al. 2024) and **PathAsst** (Sun, Zhu, et al. 2024) integrate advanced AI copilot and assistant capabilities, respectively. **MUSK** (Xiang et al. 2025) demonstrate strong performance in outcome prediction, relapse prediction, pan-cancer prognosis prediction and immunotherapy response prediction in addition to patch-level and slide-level benchmarks including multi and cross-modal retrieval, visual question answering, image classification and molecular biomarker prediction. **THREADS** (Vaidya et al. 2025) excels in generalizability demonstrating excellent performance in clinical subtyping, grading, mutation prediction, immunohistochemistry status determination, treatment response prediction and survival prediction.

**Six** of the selected models are image only but specialize in multiple modalities or staining - **REMEDIS** (Azizi et al. 2023)**, Madeleine** (Jaume, Vaidya, et al. 2024)**, PathoDuet** (Hua et al. 2024)**, Hibou** (Nechaev, Pchelnikov, and Ivanova 2024)**, RudolfV** (Dippel et al. 2024)**,** and **PLUTO** (Juyal et al. 2024). **REMEDIS** (Azizi et al. 2023) extends its capability to six medical imaging domains, including WSIs, and performs well on 15 tasks. **Madeleine** (Jaume, Vaidya, et al. 2024) and **PathoDuet** (Hua et al. 2024) specialize in processing multiple WSIs of the same sample, including those with immunohistochemistry and hematoxylin and eosin stains, achieving success in 21 and 14 tasks, respectively. The **Hibou** (Nechaev, Pchelnikov, and Ivanova 2024), **PLUTO** (Juyal et al. 2024)**,** and **RudolfV** (Dippel et al. 2024) models represent significant advancements in pathology-specific foundation models, each leveraging domain expertise and large-scale datasets to excel 12, 11, and 50 tasks, respectively.

**Hibou** (Nechaev, Pchelnikov, and Ivanova 2024), pretrained with the DINOv2 framework on over 1 million diverse WSIs, does well in both patch- and slide-level tasks, particularly in **cell-level segmentation** and classification, outperforming models like Cell-ViT (Hörst et al. 2024). **PLUTO** (Juyal et al. 2024) has its own distinctiveness in diverse pretraining dataset, backbone self-supervised modelling, adaptation, and evaluation. Its dataset contains 4-million cellular and sub-cellular annotations from broad-certified pathologists, includes samples from over 100 IHC stains and 6 special stains, it incorporated annotated data with self-supervised pretraining, and evaluation with specialized multiple heads for WSI level, tissue level, cellular and sub-cellular level prediction tasks. **RudolfV** (Dippel et al. 2024), developed with pathologist expertise and data curated from 15 laboratories, stands out in **tumor microenvironment profiling** and **biomarker evaluation** across 58 tissue types and 129 staining modalities.

**RudolfV** (Dippel et al. 2024) – developed specifically for computational pathology integrating pathologist expertise and semi-automated data curation to balance the diversity of tissue types, staining modalities, and disease profiles. Trained on 134,000 slides and 1.2 billion patches, clustered into 9 morphologically meaningful groups, RudolfV excels in nearly 50 tasks, including TME cell classification, a unique application, TME segmentation, IHC biomarker evaluation, and reference case search involving rare diseases. It shows robustness through pan-staining and pan-scanner consistency and evaluates foundation model characteristics across various settings. Notably, RudolfV outperformed UNI in 10 out of 12 benchmarks and 27 out of 31 datasets, Virchow and 2 out of 2 benchmarks. Its pathologist-driven design and ability to generalize across multiple scanners, stains, and tissue types make it a versatile and reliable tool for real-world pathology.

The remaining **four** foundation models— **CHIEF** (Xiyue Wang et al. 2024)**, Prov-GigaPath** (H. Xu et al. 2024)**, CONCH** (M. Y. Lu, Chen, Williamson, Chen, Liang, et al. 2024)**,** and **PLIP** (Huang et al. 2023)—are CLIP-like models designed for aligning pathology images with text, such as image captions or reports. Their performance spans of 27, 26, 14, and 26 tasks, respectively. **CHIEF** (Xiyue Wang et al. 2024) trained on 60,530 whole-slide images spanning 19 anatomical sitesand validated on 19,491 whole-slide images from 32 independent slide sets collected from 24 hospitals and cohorts internationally. It outperformed the state-of-the-art deep learning methods by up to 36.1% in tasks like cancer detection, tumor origin identification, genomic profile prediction (up to 53 genes 30 cancer types), IDH mutation status in brain cancer, MSI in CRC, survival prediction in 7 cancer types. Other CLIP based pathology models are not listed here considering their limited training data scale, nature of training approach (mostly adapting CLIP) and their evaluation on only less than 5 tasks.

Remarkable progress has been shown in developing pathology foundation models, with dramatic scaling in both data volume (from **HIPT**'s 10,678 WSIs to **Virchow2**'s 3.1 million WSIs) and task diversity (from basic tumor detection to complex molecular prediction). However, significant challenges remain in dataset composition and representation. Models show substantial imbalances in tissue representation - common cancers like breast, lung, and colon typically dominate the training sets, while rare cancers and non-neoplastic conditions are underrepresented. Geographic and demographic biases are equally concerning: most foundation models are trained predominantly on data from large academic medical centers in North America, Europe, or China, potentially limiting their generalizability across different healthcare settings and patient populations. For instance, while **PRISM** demonstrates impressive performance across 17 cancer types, and **PathAlign** leverages 350,855 WSIs, neither addresses the challenge of ethnic diversity in pathology patterns or variation in laboratory protocols across different regions. These obvious sources of training data biases warrant systematic investigation into model robustness across different patient populations, laboratory settings, and healthcare systems before widespread clinical deployment.

# 3. Foundation Models Technology

## 3.1. Modeling

Foundation models require several key computational properties to effectively learn and generalize from large-scale data. These properties, exemplified in medical imaging applications like digital pathology, include:

1. **Function Approximation Capacity** (traditionally and henceforth termed "**expressivity**" (Bommasani et al. 2022)): The model architecture must support universal function approximation through sufficient depth, width, and connectivity patterns. This capacity is formally characterized by the model's Vapnik-Chervonenkis (VC) dimension and approximation bounds for Lipschitz-continuous functions. In pathology applications, this manifests as the ability to learn complex mappings from gigapixel WSIs to diagnostic classifications across diverse tissue types, stains, and magnification levels. Previous works have shown that the expressivity and complexity of deep neural networks stem from their depth, width, connectivity, and structural patterns (Hanin and Rolnick 2019; Montúfar et al. 2014; Raghu et al. 2017).
2. **Computational and Statistical Scalability (“scalability”)**: The architecture must maintain stable optimization properties as both parameters and training data increase, typically measured through scaling laws that relate model performance to compute budget, parameter count, and dataset size. For pathology foundation models, this enables efficient processing of large-scale WSI datasets while maintaining consistent performance across different laboratory preparations and scanner vendors.

3. **Cross-Modal Processing Capabilities ("multimodality"** (Bommasani et al. 2022)**):** The model architecture must support joint processing of multiple input (Bommasani et al. 2022)modalities through unified representational spaces or modality-specific encoders with shared latent spaces. In pathology, this enables integration of WSI data with clinical metadata, molecular profiles, and structured report data, facilitating comprehensive diagnostic assessment. Multimodality, by incorporating large and complex distributions of images and text, plays a critical role in developing intelligence (Steyaert et al. 2023).
4. **Knowledge Storage and Retrieval Capacity ("memory capacity"** (Bommasani et al. 2022)**):** The model must effectively encode and access:
    a. Factual knowledge: Explicit medical facts, diagnostic criteria, and clinical guidelines (e.g., WHO classification criteria, TNM staging rules)
    b. Procedural knowledge: Recognition strategies and diagnostic workflows (e.g., systematic approaches to differentials)
    c. This capacity is measurable through:
    d. Knowledge probing tasks
    e. Fact completion accuracy
    f. Consistency of retrieved information across different prompting strategies
5. **Transfer Learning Capacity**: The pre-trained representations must serve as an effective initialization point for diverse downstream tasks, demonstrating both:
    a. Fine-tuning efficiency: Rapid adaptation to new tasks with minimal training data (e.g., adapting from common to rare cancer types)
    b. Zero/few-shot generalization: Task adaptation through prompting without weight updates (e.g., applying learned diagnostic criteria to novel disease presentations)
6. **Systematic Generalization** (traditionally termed **"compositionality"** (Bommasani et al. 2022)): The model must demonstrate the ability to combine learned primitives into novel compositions, formally measured through out-of-distribution generalization metrics and systematic compositionality tests. In pathology, this enables recognition of new disease patterns based on known cellular and architectural components, and application of diagnostic criteria across different manifestations of disease. Besides providing the ability to generalize from training data to unseen examples (Bengio, Courville, and Vincent 2014), compositionality boosts training efficiency and is linked to interpretability and multimodality (Bommasani et al. 2022; Dankers, Bruni, and Hupkes 2022).

The architecture of foundation models typically implements these properties through:

1. Multi-head self-attention mechanisms for capturing long-range dependencies across tissue regions and magnification levels
2. Deep, residual and hybrid architectures enabling hierarchical features learning from cellular to architectural patterns
3. Large parameters (now billions) enabling the emergence of sophisticated diagnostic capabilities
4. Pre-training objectives that promote task-agnostic representations applicable across different pathology applications

## 3.2. Training

Training the foundation models involves designing objectives that efficiently scale and generalize across various tasks and domains. Self-supervised learning plays a central role by mining vast, unlabeled datasets across modalities like text, images, and audio, enabling task-agnostic and generalized representations without the need for manual annotations. The key challenges lie in achieving domain completeness—ensuring models generalize across diverse tasks—and optimizing computational efficiency, with training objectives that predictably scale with architecture, data size, and compute resources to maximize model capabilities (Bommasani et al. 2022; Zhao et al. 2024).

Designing SSL methods involves balancing several trade-offs, particularly in input representation and model architecture. Representing raw input data, such as image pixels, preserves detailed information but slows down learning and increases computational costs. In contrast, methods like patch embeddings and tokenization enhance efficiency but risk losing valuable data. Balancing continuous and discrete input representations is another challenge, as both have unique benefits for downstream tasks. Furthermore, the choice between generative and discriminative models shapes foundation model performance: generative models, while flexible and interactive, are computationally intensive, whereas discriminative models offer faster, more efficient learning suited for high-dimensional data like images and audio (Bommasani et al. 2022; Zhao et al. 2024; Khan et al. 2024).

## 3.3. Training Pathology Foundation Models

The rapid advancement of AI in computational pathology has led to the development of numerous foundation models, each employing unique strategies for generating and utilizing embeddings—learned numerical representations that capture and compress the essential features of input data—from WSIs and associated textual data. This section provides a comprehensive analysis of the embedding strategies employed by 40 state-of-the-art models, highlighting key innovations, architectural designs, and their implications for pathological analysis. Table 2 and Table 4 present data scales and training details of single modality (image-only). Table 5 and Table 4 present data scales and training details of multimodal foundation models.

### 3.3.1. Self-Supervised Learning and DINO-based Models

*Table 2. Single Modality Image-only Pathology Foundation Models*

| Model | Architecture | Parameters | WSI | Tiles | Training Algorithm |
|---|---|---|---|---|---|
| Virchow | ViT-H | 632M | 1.5M | 2B | DINOv2 (SSL) |
| Virchow2 | ViT-H | 632M | 3.1M | 1.7B | DINOv2 (SSL) |
| Virchow2G | ViT-G | 1.9B | 3.1M | 1.9B | DINOv2 (SSL) |
| OmniScreen | Virchow2 | 632M | 48K | - | Weakly-Supervised (on Virchow2 embeddings) |
| H-Optimus-0 | ViT-G | 1.1B | >500K | - | DINOv2 (SSL) |
| Kaiko-ai | ViT-L | 303M | 29K | - | DINOv2 (SSL) |
| UNI | ViT-L | 307M | 100K | 100M | DINOv2 (SSL) |
| BROW | ViT-B | 86M | 11K | 180M | DINO (SSL) |
| Phikon | ViT-B | 86M | 6K | 43M | iBOT (Masked Image Modeling) |
| HIPT | ViT-HIPT | 10M | 11K | 104M | DINO (SSL) |
| CTransPath | Swin Transformer | 28M | 32K | 15M | MoCoV3 (SRCL) |
| Phikon-v2 | ViT-L | 307M | 58K | 456M | DINOv2 (SSL) |
| TissueConcepts | Swin Transformer | - | 7K | 912K | Supervised multi-task learning |
| PLUTO | FlexiVit-S | 22M | 158K | 195M | DINOv2 + MAE + Fourier-loss |
| Hibou-B | ViT-B | 86M | 1.1M | 1.2B | DINOv2 (SSL) |
| Hibou-L | ViT-L | 307M | 1.1M | 512M | DINOv2 (SSL) |
| Madeleine | CONCH | 86M | 23K | 48M | Multiheaded attention-based MIL |
| PathoDuet | ViT-B | 86M | 11K | 13M | MoCoV3 extension |
| RudolfV | ViT-L | 307M | 103K | 750M | Semi-supervised with DINOv2 (SSL) |
| REMEDIS | ResNet-152 | 232M | 29K | 50M | SimCLR (contrastive learning) |
| BEPH | BEiTv2 | 86M | 11K | 11M | BEiTv2 (SSL) |
| COBRA | Mamba-2 | 15M | 3,048 | - | Self-supervised contrastive learning |

Self-supervised learning, a paradigm where models learn meaningful representations from unlabeled data through automatically generated supervisory signals, has emerged as a powerful paradigm in computational pathology, with the DINO (self-**di**stillation with **no**-labels) framework that leverages (student-teacher) knowledge distillation between different views of the same image — and its derivatives playing a pivotal role. The Virchow series exemplifies this approach, with the original **Virchow** (Vorontsov et al. 2024) utilizing a ViT-H/14 architecture (632M parameters) and the DINOv2 (Oquab et al. 2024) self-supervised learning framework. This model processes WSIs as sequences of patches through self-attention mechanisms, enabling it to capture complex spatial relationships within pathological images.

The evolution of this approach is evident in **Virchow2** and **Virchow2G** (Zimmermann et al. 2024), which maintain the ViT-H architecture for Virchow2 and scale up to ViT-G (1.9B parameters) for Virchow2G. These models incorporate an extended-context translation (ECT) augmentation strategy, addressing the challenge of preserving cellular morphology during image processing—a critical consideration in pathology where fine structural details are often diagnostically significant.

*Table 3. Innovation and key features in training image-only pathology foundation models*

| Model | Key features/innovations |
|---|---|
| **Virchow** | Student-teacher paradigm with global and local cropping; extended-context translation (ECT) augmentation preserves cellular morphology |
| **Virchow2** | Scaled dataset size and increased diversity, trained on 3.1M WSIs, uses domain-inspired augmentation |
| **Virchow2G** | Scaled both data and model size, mixed magnification training; enhanced generalization across datasets |
| **OmniScreen** | Leveraging Virchow2 embeddings for weakly-supervised learning on MSK-IMPACT dataset. |
| **H-Optimus-0** | g/14 architecture, 4 registers and 40 transformer blocks; efficient handling of high-dimensional features. |
| **Kaiko-ai** | Modified DINO recipes; trained on multi-magnification TCGA WSIs; reduced GPU and batch size requirements because of Dynamic Patch Extraction |
| **UNI** | Combines self-distillation and masked image (resolution-agnostic) modeling; incorporates Sinkhorn-Knopp centering and KoLeo regularization for robustness. |
| **BROW** | Uses color augmentation, patch shuffling added to DINO framework, and multi-scale inputs |
| **Phikon** | Uses Masked Image Modeling with iBOT self-distillation; robust to image perturbations. |
| **HIPT** | Hierarchical approach for high-resolution image representation; Two-stage ViT approach capturing local and tissue-level features using DINO training. |
| **CTransPath** | Hybrid model using ConvNet for local features and Transformer for global context; semantically-relevant contrastive learning for feature richness. |
| **Phikon-v2** | Scaled ViT-L architecture; trained on 460M pathology tiles; robust ensembling for biomarker prediction. |
| **TissueConcepts** | Joint encoder utilizes transformer and convolution architectures trained with multi-task learning for classification, segmentation, and detection tasks |
| **PLUTO** | FlexiVit-multi-scale patching; Masked Autoencoder and Fourier-loss for out-of-distribution performance. |
| **Hibou-B/L** | Trained on over 1 million WSIs with RandStainNA augmentation for WSI-specific optimization. |
| **Madeleine** | Dual global-local cross-stain alignment using multi-head attention-based Multiple Instance Learning; Uses Graph Optimal Transport (GOT) framework for local patch alignment |
| **PathoDuet** | Custom self-supervised learning with cross-scale and cross-stain augmentations based on MoCoV3. |
| **RudolfV** | Trained on 134k slides across 58 tissue types and 129 staining methods; integrates pathologist expertise, uses stain-specific augmentations |
| **REMEDIS** | Utilizes SimCLR for contrastive learning to enhance feature representation. |
| **BEPH** | Lightweight self-supervised BEiT-based model pretrained using Masked Image Modeling. |
| **COBRA** | Foundation models -agnostic model leveraging contrastive self-supervised learning with multi-magnification training and efficient attention-based aggregation. |

Other models have built upon this foundation with novel enhancements. **UNI** (R. J. Chen et al. 2024), for instance, combines self-distillation (Caron et al. 2021) and masked image modeling (J. Zhou et al. 2021), incorporating improvements such as untying head weights and Sinkhorn–Knopp centering. These additions aim to improve the model's robustness to the variability inherent in pathology images, such as staining differences and artifacts.

The **BROW** (Y. Wu et al. 2023a) model took a different approach by introducing patch shuffling and color augmentation to the DINO framework. These innovations were specifically designed to improve robustness to the variability inherent in WSIs and different staining techniques, a persistent challenge in computational pathology. The **Hibou** (Nechaev, Pchelnikov, and Ivanova 2024) family of models (Hibou-B and Hibou-L) further exemplifies the potential of large-scale pretraining in pathology. These models leverage the DINOv2 framework on a dataset of over 1 million WSIs, incorporating advanced augmentations like RandStainNA (Shen et al. 2022) for WSI-specific optimization. This approach demonstrates the value of diverse, large-scale datasets in developing models capable of generalizing across various tissue types and staining methods. A notable departure from purely data-driven approaches is seen in the **RudolfV** (Dippel et al. 2024)model, which incorporated pathologist expertise in data curation. Trained on a carefully selected dataset of 134,000 slides across 58 tissue types and 129 staining methods, RudolfV excelled in tasks such as tumor microenvironment profiling and biomarker evaluation. This model's success highlights the potential synergy between expert domain knowledge and machine learning techniques in computational pathology.

The **Prov-GigaPath** (H. Xu et al. 2024) model introduced a novel two-stage approach, combining DINOv2 for tile-level pretraining with a LongNet architecture (J. Ding et al. 2023) for capturing slide-level context. By incorporating a masked autoencoder (K. He et al. 2022) pretraining step for the slide encoder and aligning vision and language using OpenCLIP (Ilharco et al. 2021), Prov-GigaPath enabled zero-shot prediction tasks. This capability is particularly valuable in pathology, where labeled data for specific conditions may be scarce. Building on similar SSL principles, **Kaiko-ai** (ai et al. 2024) introduces a scalable pipeline for training large pathology foundation models using SSL methods like DINO and DINOv2 (Oquab et al. 2024). Kaiko-ai stands out for its innovative online patching system, which allows dynamic extraction of patches from WSIs at arbitrary coordinates and magnifications. This approach contrasts with traditional offline patch extraction, improving training efficiency by reducing storage requirements and enabling more diverse patch sampling strategies. **OmniScreen** (Y. K. Wang et al. 2024) focuses on tile-level embeddings, which are critical for WSI processing. It divides WSIs into 224 × 224 pixel tiles and uses the Virchow2 (Zimmermann et al. 2024) model to embed these tiles into a 2,560-dimensional vector. These embeddings are aggregated into a global slide representation through an attention-based feed-forward network, facilitating accurate slide-level predictions. **COBRA's** (Lenz et al. 2024) model-agnostic architectures builds on SSL principles and feature space augmentations. COBRA (Lenz et al. 2024) utilizes patch embeddings from multiple foundation models and magnifications, aligning them in a shared feature space through contrastive learning. Its architecture employs Mamba-2 layers and multi-head gated attention for efficient slide-level aggregation.

### 3.3.2. Vision Transformer Based Architectures

The adoption of Vision Transformers (ViTs) in computational pathology has led to significant advancements in image understanding. The **Phikon** (Filiot et al. 2023) and **Phikon-v2** (Filiot et al. 2024) models exemplify this trend, combining Masked Image Modeling (MIM) with self-distillation techniques. Phikon-v2 demonstrated the benefits of scaling, with its ViT-L architecture trained on an impressive 460 million pathology tiles from over 100 cohorts. The model's use of robust ensembling for improved biomarker prediction showcases the potential of these architectures in clinically relevant tasks.

While Phikon-v2 focused on scaling through increased training data, **H-Optimus-0** emphasized architectural scale, employing a g/14 architecture (where 'g' denotes giant model size and '14' indicates 14x14 pixel patches) with 40 transformer blocks and 24 attention heads. This substantial increase in model capacity allowed for processing high-dimensional features in large pathology datasets, addressing the challenge of capturing fine-grained details crucial in pathological analysis. Similarly, **BEPH** (Z. Yang et al. 2024) also employed Masked Image Modeling (MIM) through the BEiTv2 ViT architecture (Peng et al. 2022) on a dataset of over 11 million pathology tiles from the TCGA. The model demonstrated significant improvements over previous architectures, outperforming DINO and ResNet in both whole-slide image classification accuracy and survival prediction.

The Hierarchical Image Pyramid Transformer (**HIPT**) (R. J. Chen et al. 2022) introduced a novel two-stage ViT model, processing patches at 256×256 pixel resolution before aggregating information at a 4096×4096 pixel region level. This hierarchical approach enables HIPT to capture both local and global features, a critical capability in representing complex histopathological structures. The model's ability to bridge different scales of analysis mirrors the way pathologists examine slides, moving between high and low magnifications to form a comprehensive assessment.

**PLUTO (**PathoLogy Universal TransfOrmer) (Juyal et al. 2024) took a different tack, focusing on creating a lightweight model that could still handle multi-scale analysis of WSIs. By incorporating FlexiViT (Beyer et al. 2023) for multi-scale patching and integrating Masked Autoencoder (MAE) (K. He et al. 2022) and Fourier-loss components, PLUTO achieved strong out-of-distribution performance despite having fewer parameters. This approach addresses an important consideration in the deployment of AI models in healthcare: the need for efficient, resource-conscious solutions that can perform robustly across varied clinical settings. Table 3 and Table 5 present key innovative features of for image only and multimodality pathology foundation models, respectively.

### 3.3.3. Hybrid and Multi-Task Models

Some models have opted for hybrid architecture, combining different neural network types to leverage their respective strengths.

**CTransPath** (Xiyue Wang et al. 2022a), for example, integrates a ConvNet with a multi-scale Swin Transformer (Ze Liu et al. 2021), employing a semantically-relevant contrastive learning (SRCL) framework. This hybrid approach aims to stabilize training and enhance the richness of extracted features, potentially capturing both low-level textural information and high-level semantic content. **TissueConcepts** (Nicke et al. 2024) takes a fully supervised multi-task learning approach, utilizing both transformer-based (Ze Liu et al. 2021) and convolution-based architectures (Zhuang Liu et al. 2022). By training on multiple tasks simultaneously (classification, segmentation, and detection), this model aims to learn more generalizable features that can be applied across various pathology tasks. This approach is particularly relevant in clinical settings where a single model capable of performing multiple analyses could streamline workflows and reduce computational overhead. Lastly, **REMEDIS** (Azizi et al. 2023) combines large-scale supervised learning (Kolesnikov et al. 2020) with contrastive self-supervised learning (T. Chen et al. 2020) to mitigate out-of-distribution (OOD) performance issues. This hybrid learning approach efficiently generalizes across unlabeled medical imaging data, offering robust performance with minimal retraining. REMEDIS is particularly adept at reducing the need for extensive labeled data, a key challenge in medical imaging.

### 3.3.4. Multimodal and Vision-Language Integration

In recent years, there has been significant progress in developing multimodal models that integrate both visual and textual data, as well as models capable of handling multiple histology stains and magnifications. These advancements have opened new avenues for more robust and comprehensive analysis in computational pathology. One such model, **PathoDuet** (Hua et al. 2024), is built on an SSL framework (Jing and Tian 2021) tailored for both H&E and IHC images. This model focuses on cross-scale positioning and cross-stain transferring pretext tasks, using a ViT (Dosovitskiy et al. 2020) architecture. PathoDuet effectively captures relationships across staining modalities and magnifications, demonstrating its superiority in downstream tasks such as colorectal cancer subtyping and tumor identification.

*Table 4. Multimodal Pathology Foundation Models*

| Model | Vision Modality | | | Text Modality | |
|---|---|---|---|---|---|
| | Model | Dataset | WSI | Model | Dataset |
| **CHIEF** | CTransPath | 14 Data Sources | 60K | CLIP | Anatomical site information |
| **mSTAR** | UNI | TCGA | 10K | BioBert | 11K Pathology Reports |
| **HistGen** | DINOv2 ViT-L | Multiple Public Data Sources | 55K | LGH Module | 7753 paired WSI Reports from TCGA |
| **PathAlign** | PathSSL | Proprietary Dataset | 350K | BLIP-2 | Diagnostic reports |
| **PathGen** | CLIP | TCGA | 7K | CLIP | 1.6M image-caption pairs |
| **HistoGPT** | CTransPath or UNI | Proprietary Dataset | 15K | BioGPT | Pathology Reports |
| **PathChat** | UNI | Multiple Data Sources | - | Llama 2 | Pathology-specific instructions |
| **PRISM** | Virchow | Virchow's dataset | 587K | BioGPT | 195K Clinical Reports |
| **PathAsst** | PathCLIP | PathCap dataset | 207K | Vicuna-13B | PathInstruct |
| **Prov-GigaPath** | ViT | Prov-Path | 171K | OpenCLIP | 17K WSI-Reports |
| **CONCH** | ViT | Multiple Data Sources | 21K | GPT-style | 1.17M image-caption pairs |
| **PLIP** | CLIP | OpenPath | - | CLIP | OpenPath |
| **SlideChat** | CONCH + LongNet | TCGA | 4915 | Qwen2.5-7B-Instruct | SlideInstruction |
| **PMPRG** | MR-ViT | Proprietary Dataset | 7422 | GPT-2 | Pathology Reports |
| **TITAN** | ViT | Mass-340K | 336K | CoCa | PAthChat – Synthetic + Medical reports |
| **MUSK** | MultiModal Transformer | TCGA | 33K | MultiModal Transformer | PubMed Central |

Similarly, **Madeleine** (Jaume, Vaidya, et al. 2024) leverages a multimodal pretraining strategy that integrates multiple histology stains as distinct views in SSL. Using a dual global-local cross-stain alignment strategy, Madeleine excels in various tasks, from molecular classification to prognostic prediction, outperforming single-stain models. The model's flexibility in incorporating additional stains makes it highly versatile for future applications in computational pathology.

Recent advancements in natural language processing have led to the development of models that integrate visual and textual information, enabling more comprehensive analysis of pathological data. **mSTAR** (Y. Xu et al. 2024) exemplifies this trend with its two-stage pretraining process. The first stage involves slide-level contrastive learning, using UNI (R. J. Chen et al. 2024) for patch-level features and TransMIL (Shao et al. 2021) for slide-level aggregation. The second stage employs self-taught training (Dosovitskiy et al. 2020) for patch-level feature learning, ensuring consistency between patch-level and slide-level representations.

**HistGen** (Zhengrui Guo et al. 2024) took a different approach, focusing on report generation through a Multiple Instance Learning (MIL) framework. Its local-global hierarchical encoder and cross-modal context module, pretrained on over 55,000 WSIs using ViT-L model with DINOv2 (Oquab et al. 2024) strategy for feature extraction, showcase how advanced architectures can be applied to the challenging task of generating human-readable reports from pathological images. This capability has significant implications for streamlining pathological workflows and improving communication between pathologists and other healthcare providers.

*Table 5. Innovation and key features in training multimodal pathology foundation models*

| Model | Key features/innovations |
|---|---|
| CHIEF | Combines self-supervised pretraining with weakly supervised learning; integrates CLIP's pretrained text encoder; uses attention-based pooling with intra and inter-WSI contrastive learning. |
| mSTAR | Two-stage pretraining integrating multimodalities - WSIs, pathology reports, and RNA-Seq data; uses inter-modality and inter-cancer contrastive learning. |
| HistGen | Applies MIL framework with DINOv2 pretrained on WSIs; focuses on report generation from pathological images. |
| PathAlign | Built on BLIP-2; uses Q-Former based WSI encoder with positional encodings for patches; enables text generation and visual QA. |
| PathGen | Integrates CLIP, LLaVA, and Llama-2; uses prompt-based retrieval and k-means clustering for patch extraction; generates captions and summaries. |
| HistoGPT | Combines CTransPath or UNI for image features with BioGPT for text; uses interleaved gated cross-attention for multimodal integration. |
| PathChat | Uses UNI pretrained on 1.18M pathology image-caption pairs with Llama 2; fine-tuned on 450K pathology-specific instructions. |
| PRISM | Combines Virchow for tile-level processing with Perceiver network for slide-level aggregation and BioGPT for text interaction. |
| PathAsst | Integrates PathCLIP with Vicuna-13B; FC layer for mapping embeddings; fine-tuned on PathInstruct. |
| Prov-GigaPath | Combines ViT with LongNet and OpenAI embeddings; employs contrastive and masked autoencoder strategies for enhanced feature representation. |
| CONCH | Uses iBOT for feature extraction and Coca (Yu et al. 2022) for contrastive and captioning objectives; integrates multiple histology stains as distinct views in SSL. |
| PLIP | Adapts CLIP for pathology-specific tasks; employs transformers for enhanced feature processing. |
| SlideChat | A multimodal (vision and text) projector; joint pretraining on image-caption pairs; fine-tuned on extensive pathology-specific instructions for interactive assistance. |
| PMPRG | Multi-scale regional representation; patient-level multi-organ report generation; explainable clinical-grade reports. |
| TITAN | Aligns (8K×8K) ROIs and WSIs with synthetic captions (obtained using PathChat) and reports, leverage CONCH (iBOT and Coca), and uses vision encoder, a text encoder, and a multimodal text decoder. |
| MUSK | Pretraining uses masked data modelling to leverage large-scale unpaired pathology images and text and then refines image and text alignment through contrastive learning |

**PathAlign** (F. Ahmed et al. 2024), built on the BLIP-2 framework (J. Li et al. 2023), focuses on aligning WSIs with diagnostic reports. Its use of a Q-Former based WSI encoder with positional encodings for patch coordinates enables precise localization of features within large pathology images. By enabling text generation and visual question-answering in pathology, PathAlign demonstrates the potential of these models to serve as interactive tools for pathologists, potentially enhancing diagnostic accuracy and efficiency.

The **PathGen** (Sun, Zhang, et al. 2024) model introduces a multi-stage process that integrates CLIP (Ilharco et al. 2021), LLaVA (H. Liu et al. 2023), and Llama-2 (Touvron et al. 2023) architectures. Its innovative use of prompt-based retrieval and k-means clustering for representative patch extraction, coupled with the ability to generate detailed captions and summaries for pathology images, showcases the potential of large language models in interpreting complex pathological data. This approach could significantly aid in the rapid interpretation of WSIs and in generating standardized reports.

**HistoGPT** (Tran et al. 2024) demonstrates another approach to multimodal integration, using either CTransPath (CTP) (Xiyue Wang et al. 2022b) or UNI (R. J. Chen et al. 2024) for image feature extraction (depending on model size) and BioGPT (Luo et al. 2022) for text processing. The model employs interleaved gated cross-attention (XATTN) blocks (Jaegle et al. 2021) to fuse visual and textual information, enabling tasks such as pathology report generation. **TITAN** (T. Ding et al. 2024) employs a three-stage by leveraging iBOT for the self-supervised learning of vision and CoCa for multimodal vision-language alignment, mixing synthetic region-level captions and slide-level pathology reports with ALiBi positional encoding to process long-context gigapixel WSIs. TITAN (T. Ding et al. 2024) outperforms earlier works in generating holistic embeddings of slides that support a few downstream tasks inclusive of zero-shot classification and rare cancer retrieval.

**PathChat** (M. Y. Lu, Chen, Williamson, Chen, Zhao, et al. 2024) and **PRISM** (Shaikovski et al. 2024) further exemplify the trend towards integrating large language models with vision encoders. PathChat's use of UNI (R. J. Chen et al. 2024) (pretrained on over 100 million histology patches) combined with a Llama 2 (Touvron et al. 2023) LLM, fine-tuned on pathology-specific instructions, demonstrates how domain-specific knowledge can be effectively incorporated into general-purpose language models. PRISM's approach of using Virchow (Vorontsov et al. 2024) for tile-level processing and a Perceiver network (Jaegle et al. 2021) for slide-level aggregation, integrated with BioGPT (Luo et al. 2022), showcases how different architectural components can be combined to handle the multi-scale nature of pathology images while enabling natural language interaction.

Another noteworthy addition is **MUSK** (Xiang et al. 2025), which employs a two-stage pretraining scheme combining masked image/language modeling on 50 million pathology patches and one billion text tokens, followed by contrastive learning on one million image–text pairs. By unifying visual and textual representations within a single multimodal transformer framework, MUSK demonstrates robust performance across diverse tasks—from image-to-text retrieval and VQA to molecular biomarker prediction and clinical outcome forecasting. Notably, it leverages large-scale, unpaired data to capture wide-ranging pathological features while still aligning vision-language features effectively in downstream tasks. This approach underscores the growing trend of integrating self-supervision with contrastive alignment to produce more versatile pathology foundation models.

**THREADS** (Vaidya et al. 2025)**—**a newly introduced slide-level foundation model that learns comprehensive WSI embeddings by aligning morphological features with genomic and transcriptomic data. It uses a ViT-L-based patch encoder (CONCHV1.5) to process 512×512 patches. On the molecular side, THREADS leverages scGPT for transcriptomic data and a multi-layer perceptron (MLP) for genomic profiles, unifying three domains via cross-modal contrastive learning on the MBTG-47K dataset (47,171 histomolecular pairs).

The development of models like **PathAsst** (Sun, Zhu, et al. 2024), **CONCH** (M. Y. Lu, Chen, Williamson, Chen, Liang, et al. 2024), **MUSK** (Xiang et al. 2025)**,** and **PLIP (**Pathology Language-Image Pretraining) (Huang et al. 2023) further illustrates the diverse approaches being explored in vision-language modeling for pathology. These models vary in their architectural choices and training strategies, from PathAsst's use of PathCLIP and Vicuna-13B (Chiang et al. 2023) to CONCH's incorporation of both contrastive and captioning objectives, MUSK's combination of unified masked modeling and large-scale contrastive alignment, and. PLIP's adaptation of the CLIP (Radford et al. 2021) architecture for pathology-specific tasks. Each demonstrates how general computer vision techniques can be effectively tailored to the unique challenges of pathology image analysis.

### 3.3.5. Conclusion and Future Directions

The diversity of embedding strategies employed by these foundation models reflects the complexity of pathological data and the multifaceted nature of pathology tasks. From self-supervised learning approaches that leverage vast amounts of unlabeled data to multimodal models that bridge visual and textual information, these advancements are reshaping how pathological analyses are conducted.

Key trends emerging from this review include:

1. The scaling of models and datasets, with architectures like Virchow2G and Hibou-L demonstrating the benefits of increased model capacity and diverse training data.
2. The importance of domain-specific adaptations, as seen in models like RudolfV that incorporate pathologist expertise in their development.
3. The rise of multimodal and vision-language models, exemplified by PathGen, HistoGPT, and PathChat, which promise more interpretable and interactive pathology AI systems.
4. The ongoing exploration of efficient architectures, such as PLUTO, that balance performance with computational requirements.

As the field continues to evolve, several challenges and opportunities emerge. These include the need for standardized benchmarks to compare model performance across diverse pathology tasks and datasets, the importance of model interpretability and explainability for clinical adoption, and the potential for these models to assist in rare disease diagnosis and personalized medicine approaches.

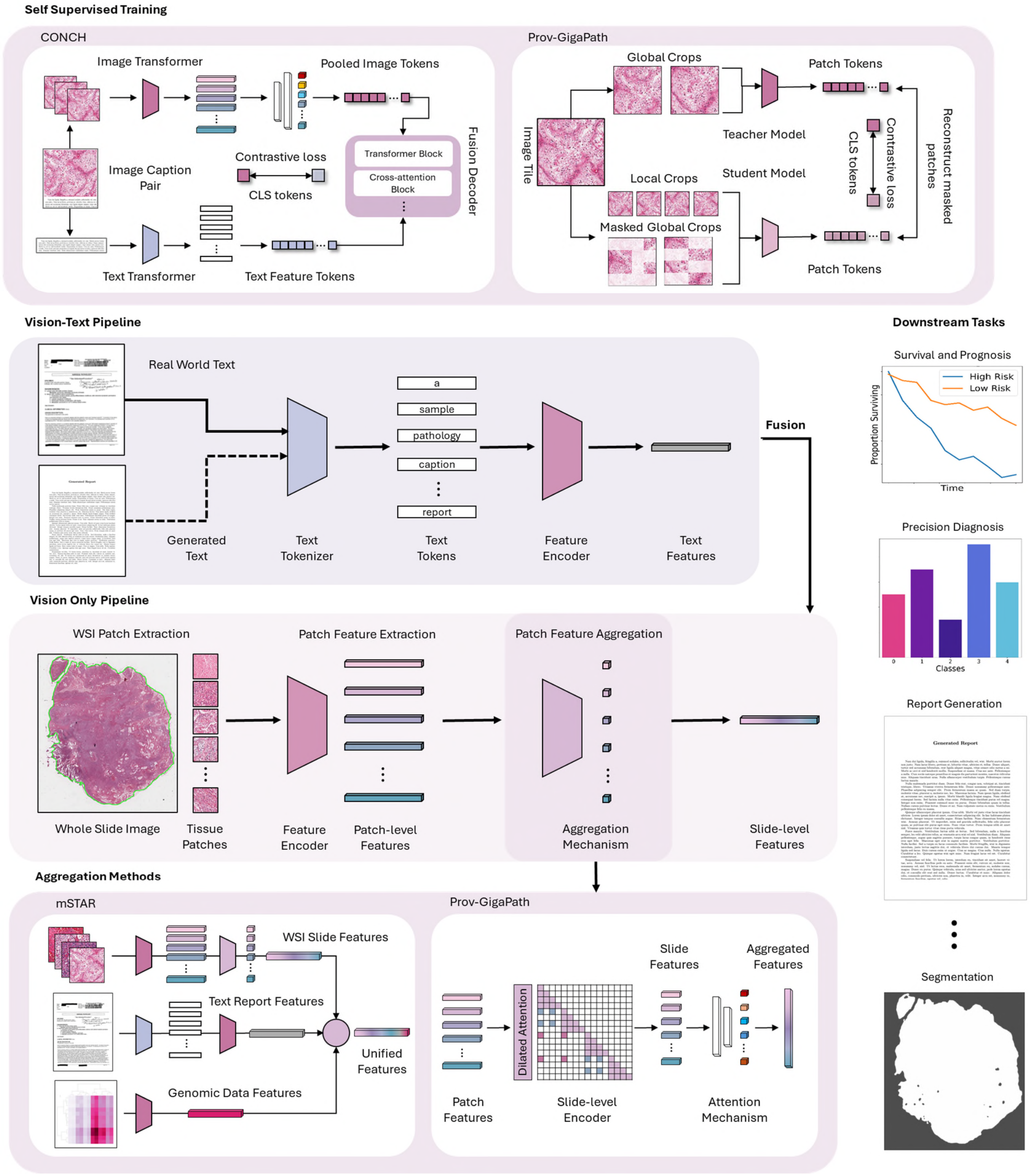

*Figure 5. Pathology Foundation Model Training: Vision-text and vision-only pipeline and aggregation methods – mSTAR's multimodal aggregation and Prov-GigaPath's patch into WSI aggregation.*

Figure 5 illustrates pathology pretraining workflow, for image-only and multiple modal pathology foundation models along with a few downstream task examples.

In conclusion, the embedding strategies employed by foundation models in computational pathology are rapidly advancing, promising to enhance diagnostic accuracy, improve workflow efficiency, and ultimately contribute to better patient outcomes. Future research directions may focus on further integration of multi-modal data sources, development of more efficient and interpretable models, and rigorous clinical validation to ensure the reliability and generalizability of these AI systems in real-world pathology practice.

As foundation models evolve, research must focus on improving multimodal learning, where models like CLIP (Radford et al. 2021) and ViLBERT (J. Lu et al. 2019) demonstrate different ways of encoding and reasoning across modalities. The future of SSL lies in creating more generalized training methods that transcend domain specificity, enabling application across fields like medical imaging and multimodal tasks. Additionally, discovering richer training signals that maximize model efficiency while minimizing computational resources is essential. Moving forward, explicit goal-directed training could enable models to autonomously adapt to real-world tasks through multitask, multiagent, and multimodal interactions, ultimately advancing the capabilities and versatility of foundation models (Bommasani et al. 2022; Zhao et al. 2024; Khan et al. 2024; J. Wu et al. 2023; Tamkin, Wu, and Goodman 2021; Ferber, Nahhas, et al. 2024).

## 3.4. GPU Needs for Pretraining

Pertaining large-scale single and multi-modality foundation models requires several GPU computers making the task unapproachable for academic research groups and labs. In Table 6, we present a few examples.

*Table 6. GPU-resource needs of pretraining pathology foundation models*

| Foundation model | Pretraining hardware resources |
|---|---|
| **Virchow2 (Zimmermann et al. 2024)**<br>**Virchow2G (Zimmermann et al. 2024)** | 16 Nvidia V100 GPUs<br>16 Nvidia V100 GPUs |
| **H-Optimus-0 (Saillard et al. 2024)** | 8 x A100 GPUs with 80Gb of memory |
| **PRISM (Shaikovski et al. 2024)** | 16 NVIDIA V100 32GB GPU's |
| **PathChat (M. Y. Lu, Chen, Williamson, Chen, Zhao, et al. 2024)** | 8 × 80GB NVIDIA A100 GPU |
| **UNI (R. J. Chen et al. 2024), TITAN (T. Ding et al. 2024), THREADS (Vaidya et al. 2025)** | 4 × 80GB NVIDIA A100 GPU |
| **CONCH (M. Y. Lu, Chen, Williamson, Chen, Liang, et al. 2024)** | 8 NVIDIA A100 80-GB GPU |
| **Prov-GigaPath (H. Xu et al. 2024)** | 16 nodes with 4 × 80 GB A100 GPUs |
| **mSTAR (Y. Xu et al. 2024)** | 4 × 80 GB NVIDIA H800 GPUs. |
| **Hibou-B (Nechaev, Pchelnikov, and Ivanova 2024)**<br>**Hibou-L (Nechaev, Pchelnikov, and Ivanova 2024)** | 8 A100-80G GPUs<br>32 A100-40G GPUs |
| **Kaiko-ai (ai et al. 2024)** | 4 – 16 H100 GPUs |
| **BEPH (Z. Yang et al. 2024)** | 8 x 40 GB NVIDIA HGX A100 GPUs |

## 3.5. Aggregation

Aggregation, the process of aggregating representations and features from tile-level predictions into slide-level prediction, has always persisted as a major challenge in computational pathology. This difficulty arises primarily due to the larger sizes of gigapixel whole-slide images (WSIs), that can go up to 150, 000 square pixels. Despite advancements, the direct application of recent developments in computer vision to digital pathology remains constrained by hardware and software limitations. In this section, we explore how foundation models address the challenge of aggregation and their potential to overcome these limitations.

Seven foundation models did not incorporate aggregation in their training and evaluation procedures. These include PathChat (M. Y. Lu, Chen, Williamson, Chen, Zhao, et al. 2024), PathAsst (Sun, Zhu, et al. 2024), PathMMU (Sun, Wu, et al. 2024), RudolfV (Dippel et al. 2024), Virchow 2 (Zimmermann et al. 2024), Virchow 2G (Zimmermann et al. 2024), and PLIP (Huang et al. 2023). These models focus their predictions on regions of interest (ROIs) within WSIs rather than applying aggregation across entire slides. However, several models have made significant strides in aggregation, integrating advanced methodologies to address the issue.

### 3.5.1. Advancing aggregation

Nine foundation models develop new aggregation methodologies.

1. **mSTAR** (Y. Xu et al. 2024): This multimodal foundation model employs a two-stage whole-slide pretraining pipeline. In the first stage, it trains a Slide Aggregator (Teacher) to inject multimodal knowledge, followed by a Patch Extractor (Student) in a self-taught student-teacher framework. Leveraging the UNI (R. J. Chen et al. 2024) model for patch representations and a slide-level contrastive loss, mSTAR (Y. Xu et al. 2024) demonstrated superior performance across seven varied tasks, including diagnostic predictions, molecular analyses, survival predictions, zero-shot slide classification, and report generation. It consistently outperformed UNI (R. J. Chen et al. 2024), CONCH (M. Y. Lu, Chen, Williamson, Chen, Liang, et al. 2024), PLIP (Huang et al. 2023), and ResNet50 (K. He et al. 2016).

2. **PathAlign** (F. Ahmed et al. 2024): Built on BLIP-based image-text alignment, PathAlign encodes WSIs using the Q-Former from BLIP-2 (J. Li et al. 2023) image transformer, representing each slide as a sequence of up to 10,240 patch embeddings. Pathologist evaluations rated the generated WSI text as clinically accurate for 78% of cases, demonstrating the promise of generative AI in WSI reporting.

3. **PRISM** (Shaikovski et al. 2024): This foundation model integrates Virchow tile embeddings with clinical report text, using a Perceiver network (Jaegle et al. 2021) for slide-level encoding and BioGPT (Luo et al. 2022) for language decoding. PRISM (Shaikovski et al. 2024) has shown excellent performance in cancer detection and subtyping, including rare cancers, with AUC scores exceeding 0.9 in both linear probing and fine-tuning.

4. **Prov-GigaPath** (H. Xu et al. 2024): This model introduces GigaPath, a novel ViT architecture tailored for pretraining gigapixel pathology slides. GigaPath leverages LongNet's (J. Ding et al. 2023) dilated self-attention to process the long sequences of visual tokens generated from the image tiles. Pretraining follows a two-step process, first using DINOv2 for image-level self-supervised learning and then employing a masked autoencoder (K. He et al. 2021) with LongNet for slide-level learning.

5. **SlideChat** (Y. Chen et al. 2024): This is a first vision-language assistant that understands gigapixel WSIs. It leverages on CONCH (M. Y. Lu, Chen, Williamson, Chen, Liang, et al. 2024) for patch encoding, and LongNet (H. Xu et al. 2024)· (J. Ding et al. 2023) as a slide encoder to employ sparse attention to aggregate slide-level features.

6. **CHIEF** (Xiyue Wang et al. 2024): This multimodal model employs a two-stage training process. First, it uses self-supervised learning for patch-level feature extraction, followed by weakly supervised learning with an attention module for WSI-level representation. Its attention-based pooling strategy includes three modules: a main deep attention aggregation module that computes class-specific attention scores for each tile, and two auxiliary modules for inter-WSI and intra-WSI feature learning. The instance branch assigns scores of 1 to tiles with the highest attention and 0 to those with the lowest, while the WSI branch uses contrastive learning to integrate information and enhance category separation at the WSI level.

7. **COBRA** (Lenz et al. 2024) employs Mamba-2 layers and multi-head gated attention to efficiently aggregate long sequences of patch embeddings from models like Virchow2[19] and UNI[13] for slide representation.

8. **TITAN** (T. Ding et al. 2024) leverages a ViT to encode a WSI into a slide embedding, in a 3-stage WSI-language alignment pipeline. Stage 1 refers to vision-only part that uses self-supervised learning with student–teacher knowledge distillation followed by Stage 2 and 3 for vision-language modeling that aligns WSI embeddings with synthetic caption and medical reports. ROIs in Stag 2 and eventually WSI in Stage 3 are spatially arranged in a 2D feature grid as a slide embedding.

9. **THREADS** (Vaidya et al. 2025) employ cross-modal contrastive learning to align the slide representation with the corresponding molecular embedding. The slide encoder aggregates tile embeddings generated by ROI encoder model (CONCHv1.5) into a slide representation using attention-based modeling. The RNA embedder concatenates transcriptomic profiles encoded using a single-cell foundation model, and DNA embedder concatenates genomic profiles encoded using a multi-layer perceptron model, respectively.

### 3.5.2. Reusing Attention-based aggregation

Following seven foundation models adopted attention-based MIL aggregation methodology.

1. **Madeleine** (Jaume, Vaidya, et al. 2024): Madeleine introduces multi-stain-guided slide representation learning, employing CONCH (M. Y. Lu, Chen, Williamson, Chen, Liang, et al. 2024) for patch embedding, followed by pre-attention and stain encoding. This is then passed through a multi-head ABMIL network (Ilse, Tomczak, and Welling 2018), yielding superior results compared to other models.

2. **PLUTO** (Juyal et al. 2024): PLUTO is a light weight foundation model introduces multi-stain, multi-resolution slide representation learning, that capture different biological contexts at slide, tissue, cellular/sub-cellular level. It also uses multi-head ABMIL network (Ilse, Tomczak, and Welling 2018) on frozen and trainable PLUTO featurizer backbone, for slide-level classification tasks.

3. **OmniScreen** (Y. K. Wang et al. 2024): This model uses an attention-based feed-forward network (Ilse, Tomczak, and Welling 2018) to aggregate tile-level embeddings into slide-level predictions. OmniScreen uses tile embeddings generated by Virchow 2 (Zimmermann et al. 2024), demonstrating its ability to perform large-scale aggregation effectively.

4. **UNI** (R. J. Chen et al. 2024) and **Virchow** (Vorontsov et al. 2024) are image-only foundation models that utilize DINOv2 for self-supervised pretraining on image patches, employing a student-teacher knowledge distillation approach for large ViT architectures. Both models adopt a two-stage multiple-instance learning (MIL) paradigm for WSI-level classification tasks but differ in their aggregation methods. In **UNI** (R. J. Chen et al. 2024), the process involves pre-extraction of ROI-level features and using a trainable permutation-invariant pooling operator to aggregate patch-level (or instance) features into a single slide-level (or bag) feature, similar to the CLAM (M. Y. Lu et al. 2021) model. For both the WSI Classification and Survival Prediction tasks, **BEPH** (Z. Yang et al. 2024) also aggregated patch-level features into WSI-level representations using a MIL framework inspired by the CLAM model.
5. In **Virchow** (Vorontsov et al. 2024), the **Agata** aggregator model (Raciti et al. 2023) is employed for weakly supervised learning, using a MIL approach to aggregate features for slide-level predictions, enhancing the model's performance in WSI tasks through more effective information integration. Agata aggregator learns to attend the selective tiles that contribute toward the label decision using cross-attention instead of full attention.
6. **MUSK** (Xiang et al. 2025) is patch-level vision-language foundation model that leverages ABMIL network (Ilse, Tomczak, and Welling 2018) to yield slide level predictions.

### 3.5.3. Challenges and Future Directions

Despite these advancements, the field still faces several hurdles. Aggregation remains a computationally intensive task due to the size and complexity of WSIs. Most foundation models currently do not incorporate aggregation techniques, limiting their ability to deliver slide-level predictions. Additionally, models that do support aggregation vary in their approaches, leading to inconsistencies in performance across different pathology tasks.

To fully unlock the potential of foundation models in digital pathology, research must focus on integrating and evaluating aggregation techniques into the foundation models and improving the integration of multimodal data. Advances in both hardware and software are essential to overcome current limitations, enabling more efficient and scalable solutions.

In summary, aggregation is a critical component in advancing pathology foundation models to clinical-grade performance. The diverse approaches taken by models like mSTAR3, PathAlign, PRISM, and Prov-GigaPath illustrate the progress being made, but there remains significant room for growth. Continued innovation in this area is key to realizing the full potential of AI in pathology.

# 4. Evaluation

Pathology foundation models are likely to be adapted to perform specific tasks for countless applications, leading to emergent abilities. This leads to unique challenges in evaluating foundation models to serve in tracking progress, fostering understanding, and providing documentation (Bommasani et al. 2022; Dehghani et al. 2021; W. Yang et al. 2024).

Evaluating foundation models can be broken down into two categories: **intrinsic and extrinsic evaluation**[1]. Intrinsic evaluation focuses on assessing the model itself and its **foundational capabilities**, independent of any specific task (Tamkin et al. 2023). Extrinsic evaluation involves assessing task-specific models that have been adapted from the foundation model, to evaluate their **emergent abilities** (Bommasani et al. 2022; Y. Wang et al. 2024). The task-specific evaluation via meta-benchmarks (Balachandran et al. 2024) relies on adaptation processes making it difficult to attribute performance gains either to the foundation model or to the adaptation method. To address this, intrinsic evaluation could focus on directly measuring capabilities like **bias, vision, and/or linguistic competence**, **independent of any specific task** (Gallegos et al. 2024).

Evaluating foundation models should also involve **accounting for the resources involved in training and adaptation**. It requires considering all resources used in the adaptation process too, from the data required to choose adaptation methods to access constraints. This ensures that evaluations not only assess task-specific performance but also offer insights into the best adaptation strategies for different contexts (Longpre et al. 2024; Zishan Guo et al. 2023).

## 4.1. Adaptation

Adaptation refers to the process of adjusting a foundation model, originally trained on vast and diverse data, to perform more effectively in specialized tasks. Through adaptation, a foundation model's broad capabilities are refined to improve its precision and relevance for targeted specialized tasks (Bommasani et al. 2022; C. Shi et al. 2024).

Despite the wide-ranging abilities of foundation models, their general nature often falls short when handling downstream tasks requiring deep domain knowledge or a high level of contextual sensitivity[1]. Adapting these models to specific tasks not only improves their performance but is also more computationally efficient than training a model from scratch. This task-specific adaptation is essential for enhancing the model's effectiveness in real-world applications, ensuring that it meets the required level of accuracy and efficiency (Bommasani et al. 2022; Z. Chen et al. 2024; W. Li et al. 2024).

The methods of adapting foundation models vary from **prompting** to **fine-tuning** and **continual learning techniques**. In prompt-based adjustments, task-specific instructions guide the model's output without changing its internal structure. Fine-tuning involves updating the model's parameters using domain-specific data to improve the model's focus and accuracy on each downstream task. The resource-efficient methods, such as low-storage adaptations, minimize the need for computational and storage overhead by selectively adjusting only certain parameters of the model (Bommasani et al. 2022; Firoozi et al. 2023; J. Liu et al. 2024).

Adapting a foundation model comes with its own set of challenges, as it contrasts with the original promise of a generalist AI capable of handling a wide range of tasks without modification. The reality is that no model can be equally effective across all domains, and adaptation is crucial to bridge the gap between generalization and specialization. While adaptation enhances task performance, it also requires additional resources, raising questions about efficiency and usability. Nevertheless, adaptation remains a critical component in making foundation models more applicable to specific applications (Bommasani et al. 2022; Alowais et al. 2023; X. Zhang et al. 2024).

Looking toward the future, a significant goal for foundation models is the development of effective continual learning techniques, where models can continuously update their knowledge in response to evolving research and development. This is a challenging task, as continual learning can lead to catastrophic forgetting, where new information overwrites previously learned knowledge. To address this, innovations in memory mechanisms and parameter updates are being explored. One such example is QPMIL-VL (Gou et al. 2024), a Vision-Language framework for incremental WSI classification. It addresses the challenge of catastrophic forgetting by leveraging CONCH and a queryable prototypes (pool of representative features) that capture diverse visual patterns across WSIs. This allows the model to match new images with these stored prototypes, minimizing the risk of "forgetting" previously learned information.

Achieving continual learning would reduce the need for resource-intensive retraining while keeping the models aligned with current socio-cultural expectations. However, the risks of misalignment or feedback loops emphasize the need for caution in developing these systems. Adaptation, whether through fine-tuning or continual learning, remains key to unlocking the full potential of foundation models in specialized and evolving environments (Bommasani et al. 2022; Verwimp et al. 2024; T. Wu et al. 2024).

## 4.2. Evaluation of pathology foundation models

Extrinsic evaluation has been as a major focus in pathology foundation models universally, where the models have been evaluated after adapting them to very specific downstream tasks.

### 4.2.1. Evaluation of image only models

Here we list downstream tasks on which image only pathology foundation model has been evaluated, all of which serve extrinsic evaluation.

**UNI** (R. J. Chen et al. 2024) demonstrated performance on several downstream tasks from ROI classification, ROI segmentation, slide classification, few shot classification, retrieval and prototyping or prompt-based evaluation. The generalizability of **CHIEF** (Xiyue Wang et al. 2024) foundation model have been evaluated with slide-level downstream tasks—cancer cell detection, tumor origin identification, genomic profile characterization and survival outcome prediction on the diverse external validation cohorts. **BEPH** (Z. Yang et al. 2024) was evaluated on patch-level classification, slide-level classification and slide-level survival prediction. **Virchow** (Vorontsov et al. 2024) excelled on tile-level linear probing benchmarks and whole slide level tasks which include pan-cancer detection and subtyping, tissue-agnostic cancer detection and subtyping, and biomarker prediction tasks, adapted with the training of an aggregator model.

Similar to **Virchow** (Vorontsov et al. 2024), tile-level classification benchmarks via linear probing used to evaluate **Virchow 2** (Zimmermann et al. 2024) capabilities. The robustness of **Virchow** (Vorontsov et al. 2024) and **Virchow 2** (Zimmermann et al. 2024) embeddings has been evaluated with out-of-distribution data (i.e. data obtained from institutions other than the training cohort). The evaluation of **OmniScreen** (Y. K. Wang et al. 2024)—built on tile embeddings of **Virchow 2**—involved its evaluation on a test set from the development cohort and on TCGA as an external validation cohort. **OmniScreen** has been evaluated on pan-cancer biomarker prediction (for screening of up to 505 genes for 15 most commonly treated cancers), biomarkers associated with histologic subtypes of cancers, biomarkers associated with targeted therapeutic hotspots, biomarkers associated with signaling pathways and genome instability.

**Kaiko-ai** (ai et al. 2024) adopts two groups of metrics to evaluate the performance of the foundation models. The first group of metrics evaluates the quality of the representations directly without labels using RankMe (Garrido et al. 2023) and off-diagonal correlation (ODCorr); whereas the second group of metrics benchmarks the performance of the representations on the downstream patch-level prediction tasks using **eva** framework with a lightweight head network, where labels are necessary to perform linear probing evaluation. **Eva**[b] is a pioneering global leaderboard for benchmarking of pathology foundation models in downstream six tile-level and two slide-level tasks. Downstream tile-level prediction tasks included nuclei segmentation evaluated with the Colorectal Nuclear Segmentation and Phenotypes (CoNSeP) and multi-organ nuclei segmentation (MoNuSAC) datasets.

**PLUTO** (Juyal et al. 2024) demonstrated its generic capabilities by multi-head adaptation – adding task-specific heads and adapt these heads through supervised fine-tuning. These adaptations have been carried out for WSI classification tasks with a multiple-instance learning, tile classification (tissue level) and instance segmentation (cellular- and subcellular-level). **Phikon-V2** (Filiot et al. 2024) evaluation comprises eight slide-level tasks that goes downstream training procedure following an ensemble strategy. It has also been evaluated in one-shot retraining settings. **CtransPath** (Xiyue Wang et al. 2022a) has been validated with patch retrieval, patch classification, weakly-supervised WSI classification, mitosis detection, and colorectal adenocarcinoma gland segmentation. **BROW** (Y. Wu et al. 2023a) demonstrated its emergent capabilities in downstream adaptions to slide-level subtyping, patch-level classification and nuclei instance segmentation tasks.

**PathoDuet** (Hua et al. 2024) has been evaluated on patch-level tissue subtyping and WSI-level classification as downstream tasks with H&E images and IHC expression level assessment, cross-site tumor identification, qualitative analysis of IHC slide as downstream tasks with IHC images implemented as a linear protocol. IHC expression level assessment refers to classification between IHC patches of different expression levels and cross-site tumor identification refers to cancer cell identification with data from two sites as an advanced challenge to models' capabilities. IHC  slide-level quantitative analysis refers binary classification with MIL setting for biomarker prediction. The performance of **Madeleine** (Jaume, Vaidya, et al. 2024)—a multimodal pretraining strategy for slide representation learning —has been demonstrated on few-shot classification, and full classification on breast cancer and kidney cancer subtyping tasks.

**RodulfV** (Dippel et al. 2024) has been evaluated on pan-indication H&E-based tumor microenvironment (TME) characterization, pan-indication immunohistochemistry biomarker evaluation, histological and molecular prediction benchmarks and reference case search including rare oncological and non-neoplastic diseases. **RodulfV's** ability of cross-tissue disease understanding in order to demonstrate generalization to new indications and TME characterization through cell classification especially for immune and stromal cells such as fibroblasts, granulocytes, macrophages, and plasma cells the model improved by 10.8% on average over the closest contender are remarkable abilities. Image-based reference case search is an interesting benchmark to evaluate a foundation model for semantically meaningful representation of tissue. In this task, pathologist annotates a region of interest (ROI) that is queried against a database of slides to retrieve most similar slides.

---

[b] https://kaiko-ai.github.io/eva/main/

**TissueConcepts** (Nicke et al. 2024)**—**the only multitask **supervised** foundation model—drastically reduced very expensive training of foundation model in terms of data, computation, and time. The performance and generalizability of **TissueConcepts** (Nicke et al. 2024) has been demonstrated on classification of WSIs from four of the most prevalent solid cancers - breast, colon, lung, and prostate. Its unique training recipe is a promising alternative that achieves comparable performance to self-supervised foundation models with only 6% of the data and resources. Another unique aspect of its evaluation is its encoders, trained for 160 hours on a single Nvidia RTX A5000 in Europe, to emit an estimated 18.91 kg of CO2 as compared to order of magnitude emission of up to 2004 kg of CO2 by self-supervised encoder trained for 160 hours on 48 Nvidia V100s.

Majority of pathology foundation models we reviewed are task-agnostics which have been trained with unsupervised SSL. However, pathology foundation models like **TissueConcepts** (Nicke et al. 2024) and **RudolfV** (Dippel et al. 2024) are uniquely able to excel in performing multiple specialized tasks while trained as fully supervised and semi-supervised making them an excellent alternate to compare SSL-based image only pathology foundation model. To demonstrate the classification performance, area under the receiver operator curve (AUROC) and balanced accuracy, whereas for the nuclei segmentation Dice score has been used.

After a thorough review of the evaluation and results sections of image only foundation models, we have observed several novel contributions and research gaps in evaluating pathology foundation models:

1- **Kaiko-ai** (ai et al. 2024) is the only foundation model, which has also been evaluated with an intrinsic evaluation using RankMe (Garrido et al. 2023) and off-diagonal correlation (ODCorr). RankMe (Garrido et al. 2023) estimates the rank of embeddings of test data and ODCorr measures the average correlation coefficient between the embeddings of different samples in the evaluation dataset. Authors have found ODCorr highly correlates with the downstream performance making it a useful extrinsic evaluation metric.

2- **TissueConcepts** (Nicke et al. 2024)**—**only multi task supervised foundation model—drastically reduced very expensive training of foundation model in terms of data, computation, and time. It's training recipe is a promising alternative that achieves comparable performance to self-supervised foundation models with only 6% of the data and resources.

3- **TissueConcepts** (Nicke et al. 2024) further contributes uniquely by evaluating CO2 emission evaluation of its encoders, trained for 160 hours on a single Nvidia RTX A5000 in Europe, to emit an estimated 18.91 kg of CO2 as compared to order of magnitude emission of up to 2004 kg of CO2 by self-supervised encoder trained for 160 hours on 48 Nvidia V100s (e.g. **CtransPath** (Xiyue Wang et al. 2022a)).

4- As demonstrated by **RodulfV** (Dippel et al. 2024), Image-based reference case search is an interesting benchmark to evaluate a foundation model for semantically meaningful representation of tissue. In this task, pathologist annotates a region of interest (ROI) that is queried against a database of slides to retrieve most similar slides.

5- Comparative analysis of **RodulfV** (Dippel et al. 2024) with **UNI** (R. J. Chen et al. 2024)**, Phikon** (Filiot et al. 2023)**, Virchow** (Vorontsov et al. 2024) and **Virchow 2** (Zimmermann et al. 2024) suggests that leveraging pathologist domain knowledge and data diversity can have a similar effect as an order of magnitude more data available for training. This gives another dimension to the pathology foundation models—training on very large and highly diverse as well as curated and balanced datasets built with expertise knowledge will likely be necessary— which might be explored further to validate and confirm the true benefits for such models in terms of both the competence and usefulness.

6- Zero-shot evaluation of downstream tasks appears an advance evaluation standard which appears to be vital to demonstrate foundation model useability as an AI copilot. However, this generally gives lower performance than finetuned or adapted foundation model.

7- There have been a few novel attempts to implicitly evaluate the generalist performance of the pathology foundation models, however, there is still a need for adaptation of pathology foundation models that required task specific data and additional computational resources for these evaluations:

   a. OncoTree cancer classification by **UNI** (R. J. Chen et al. 2024) is a large-scale hierarchical classification task in pathology that follows the OncoTree (OT) cancer classification system.
   b. Prediction of rare caners has been demonstrated by several foundation models.

c. **OmniScreen** (Y. K. Wang et al. 2024) has been evaluated on pan-cancer biomarker prediction (for screening of up to 505 genes for 15 most commonly treated cancers), biomarkers associated with histologic subtypes of cancers, biomarkers associated with targeted therapeutic hotspots, biomarkers associated with signaling pathways and genome instability.

d. **RodulfV** (Dippel et al. 2024) has been evaluated on reference case search including rare oncological and non-neoplastic diseases. **RodulfV's** ability of cross-tissue disease understanding in order to demonstrate generalization to new indications and TME characterization through cell classification especially for immune and stromal cells such as fibroblasts, granulocytes, macrophages, and plasma cells the model improved by 10.8% on average over the closest contender are remarkable abilities. Image-based reference case search is an interesting benchmark to evaluate a foundation model for semantically meaningful representation of tissue.

e. **PathoDuet** (Hua et al. 2024) evaluated on IHC expression level assessment, cross-site tumor identification.

8- The analysis of **Virchow's** (Vorontsov et al. 2024) experimental analysis identified that the long-tailed distribution of pathologic entities and histological structures, the lack of object scale diversity and the restricted color space are the aspects which merit further exploration.

9- There has not been an evaluation explicitly designed to directly measure the capacities of foundation models like bias and generalist competence, independent of any specific task focused on assessing the model itself.

### 4.2.2. Evaluation of image and text aligned models

Here we include downstream tasks on which image and text aligned pathology foundation has been evaluated, all of which serve extrinsic evaluation.

**CONCH** (M. Y. Lu, Chen, Williamson, Chen, Liang, et al. 2024)—a vision and language aligned foundation model has been evaluated with image classification, segmentation, image retrieval, text-to-image and image-to-text retrieval tasks. **CONCH** has been evaluated with supervised, weakly supervised classification experiments, end-to-end fine-tuning for classification experiments, and captioning with fine-tuning. The supervised learning based adaptation has been evaluated to maximize the task-specific performance with labeled training examples from the official training set.

First introduced in histopathology by **PLIP** (Huang et al. 2023)**,** a **zero-shot classification** of diverse tissues and diseases, including rare diseases is an interesting evaluation for **CONCH** (M. Y. Lu, Chen, Williamson, Chen, Liang, et al. 2024) where the goal is to classify an image by matching it with the most similar text prompt in the model's shared image–text representation space, which has been created by predetermined text prompts for all classes. For each downstream task, both tile-level and slide-level, user first represents the set of class or category names using a set of predetermined text prompts, where each prompt corresponded to a class. The usefulness of zero-shot capability has yet to be qualified because of lower performance scores.

Zero-shot cross-modal retrieval or image search application—retrieving the corresponding text entry on the basis of an image query (image-to-text) or vice versa (text-to-image) —is an addition in the evaluation set of CONCH (M. Y. Lu, Chen, Williamson, Chen, Liang, et al. 2024). CONCH has also been valuated with coarse-grained tissue segmentation on WSIs without labeled examples using the demonstrated zero-shot retrieval and classification capabilities of our model.

Similar to image retrieval – image-text aligned models like **PLIP** (Huang et al. 2023) enable image-to-text and text-to-image retrievals are steps towards an advanced evaluation however there is yet a research gap of directly measuring foundation models capabilities like bias and generalist competence, independent of any specific task in addition to an evaluation focused on assessing the model itself, independent of any specific task. The current visual-language pretrained models, including **CONCH** (M. Y. Lu, Chen, Williamson, Chen, Liang, et al. 2024), performed inferior on challenging zero-shot problems as compared to their supervised learning counterparts, which suggests that building a generalized foundation model capable of truly universal zero-shot recognition or retrieval for histopathology is in its early days and there is yet a long path to meet its goal of truly generalist pathology foundation model.

**PathAlign** (F. Ahmed et al. 2024) has been evaluated for WSI classification, image-to-text retrieval and text generation capabilities. The text output went through automatic evaluation based on similarity scores and qualitative evaluation by two US-board certified pathologists for Top-K and image-to-text retrieval and text generation. A case prioritization

example—a theoretical case load of colon biopsies which PathAlign has to sort by "severity" of the pathology findings— has been used to demonstrate potential vision-language application.

**MUSK** (Xiang et al. 2025)—a vision-language foundation model aligns clinical notes with pathological characteristics for precision oncology. In zero-shot cross-modal retrieval MSUK demonstrated superior performance over the seven other foundation models in both image-to-text and text-to-image retrieval, with **CONCH** (M. Y. Lu, Chen, Williamson, Chen, Liang, et al. 2024) being the second-best model. With minimal training, **MUSK** outperformed other vision–language foundation models in visual-question answering including those which specifically designed for VQA purposes. To demonstrate image encoder capabilities, it has MUSK outperformed the other foundation models in image retrieval, zero-shot image classification, few-shot image classification, and supervised image classification across 12 datasets. MUSK also achieved significantly higher performance than other pathology foundation models, in molecular biomarker prediction, melanoma relapse prediction, and pan-cancer prognosis prediction (across 16 major cancer types in TCGA), objective response (to classify immunotherapy responder vs non-responders based on tumor PD-L1 expression) and progression-free survival.

After a thorough review of the evaluation and results sections of image and text aligned foundation models, we have observed several novel contributions and a few research gaps in evaluating pathology foundation models:

1- **Retrieval** is an advanced evaluation standard that evaluates the quality of embedding produced by foundation model acting as encoders for content-based image retrieval of histology images, where the goal is to retrieve similar images (same class label for downstream ROI-level classification task) to a given query (or test set) image.

2- Zero-shot cross-modal retrieval or image search application—retrieving the corresponding text entry on the basis of an image query (image-to-text) or vice versa (text-to-image).

3- Similar to image retrieval – image-text aligned models enable image-to-text and text-to-image retrievals are steps towards an advanced evaluation however there is yet a research gap of directly measuring foundation models capabilities like bias and generalist competence, independent of any specific task in addition to an evaluation focused on assessing the model itself, independent of any specific task.

4- The current visual-language pretrained models, including **CONCH** (M. Y. Lu, Chen, Williamson, Chen, Liang, et al. 2024), performed poorly on challenging zero-shot problems as compared to their supervised learning counterparts, which suggests that building a generalized foundation model capable of truly universal zero-shot recognition or retrieval for histopathology is in its early days and there is yet a long path to meet its goal of truly generalist pathology foundation model.

5- A case prioritization example—a theoretical case load of colon biopsies which **PathAlign** (F. Ahmed et al. 2024) has to sort by "severity" of the pathology findings— has been used to demonstrate potential vision-language application.

6- One key aspect lacking in evaluating language-image aligned models is the comparison of separated image and language embeddings with their standalone counterparts.

7- **MUSK** (Xiang et al. 2025) training data include TCGA WSIs, therefore, its downstream performance on TCGA could be overestimated considering the data leakage during the model training.

### 4.2.3. Evaluation of multimodal models

**mSTAR** (Y. Xu et al. 2024) **–** a whole slide multimodality foundation model leveraging on Image Only foundation model like **UNI** (R. J. Chen et al. 2024) for image embeddings, involved training a slide aggregator on pairs of WSI-report, WSI-gene, and gene-report pairs. It has been evaluated on diagnostic tasks, molecular predictions, pathology survival analysis, novel multimodal survival analysis, zero and few-shot slide classification, and report generation. Quantitative evaluation of report generation uses BLEU, METEOR, and ROUGE-L to assess precision of n-grams (contiguous sequences of words), order, alignment, and recall, etc. For qualitative evaluation, the reports of mSTAR were analyzed in comparison to doctors ground truths. **HistGen** (Zhengrui Guo et al. 2024) has also been evaluated with WSI report generation, cancer subtyping, and survival analysis.

A slide-level foundation model—**PRISM** (Shaikovski et al. 2024) that builds on **Virchow** (Vorontsov et al. 2024) tile embeddings aggregated through Perceiver network and leverages clinical report text for pretraining. It has been evaluated on sixteen cancer detection tasks including 7 rare cancers, three cancer subtyping, and nine biomarker prediction tasks with zero-shot, linear probing and finetuning based adaptations settings. To demonstrate language-vision capabilities, it has been evaluated for report generation via image caption generation for a slide or a specimen using autoregressive decoding.

**HistoGPT** (Tran et al. 2024) **–** a vision-language model that generates reports from a series of pathology images of dermatology patients to demonstrate AI assistance to dermatologist. It's architecture consists **UNI** (R. J. Chen et al. 2024) and **CtransPath** (Xiyue Wang et al. 2022a) for as patch encoders, and BioGPT base for HistoGPT-L as a position encoder, and HistoGPT-S/M as a slide encoder, and a textual prompt processing transformer with a text head. It has demonstrated the prediction tumor subtypes and tumor thickness in a zero-shot fashion, in addition to report generation, disease classification and text-to-image visualization. To evaluate generative performance analysis of HistoGPT, authors have used four semantic-based machine learning metrics as well as two blinded domain expert evaluations. Authors have also evaluated all models using traditional syntax-based measures of BLEU-4, ROUGE-L, METEOR, and BERTscore.

**ProvGigaPath** (H. Xu et al. 2024) also evaluates image only as well as image-text aligned foundation model have first been adapted to downstream, task-specific extrinsic evaluations of cancer subtyping, gene mutation prediction, and novel fine-tuned image-report alignment leveraging zero-shot cancer subtyping and zero-shot gene mutation prediction. To finetune ProvGigaPath with standard cross-modal contrastive loss in continual pretraining, image only ProvGigaPath have been used as vision encoder, GPT3.5 for preparing cleaned reports and PubMedBERT for text embeddings.

Report generation or image caption generation appears as a significant contribution of language-vision models. For report generation, both quantitative (BLEU, METEOR, and ROUGE-L etc.) and qualitative evaluation (in comparison to doctors ground truths) have been demonstrated by **mSTAR** (Y. Xu et al. 2024) and **HistoGPT** (Tran et al. 2024).

**THREADS** (Vaidya et al. 2025) is a molecular-driven multimodal slide-level foundation model capable of generating universal representations of whole-slide images. **THREADS** has been evaluated on variety of tasks belonging to—clinical subtyping and grading, gene mutation prediction, immunohistochemistry (IHC) status prediction, and patient prognostication including treatment response and survival prediction—12 of which are in-house in a cross-validation matter and 42 publicly available datasets to demonstrate generalization and transferability to external dataset. Comparative analysis using linear probing demonstrated supervisor performances against **PRISM** (Shaikovski et al. 2024) , **ProvGigaPath** (H. Xu et al. 2024), and **CHIEF** (Xiyue Wang et al. 2024)**.** Data and label efficiency of THREADS have been demonstrated superior in 4 of 7 to predict patient treatment response and resistance and 5 of 6 survival prediction tasks. **THREADS** finetuning has demonstrated superior performance on all 54 tasks against **CHIEF** (Xiyue Wang et al. 2024) finetuning and 40 of 54 against **GIGAPATH** finetuning. **THREADS** outperformed the baseline foundation models in 12 retrieval tasks. **THREADS** introduces "molecular prompting", a new class of assessment to evaluate transfer and generalization without supervision. Further insights have been attained by studying clustering capabilities of the latent space measured via Rand index and mutual information, tSNE visualization, and data and model scaling laws into **THREADS**. **THREADS** may produce an overestimated generalization performance on TCGA because of its inclusion as a pretraining cohort.

1- **mSTAR** (Y. Xu et al. 2024) evaluated on novel multimodal survival analysis, zero and few-shot slide classification, and report generation.
2- **HistoGPT** (Tran et al. 2024) **–** a vision-language model that generates reports from a series of pathology images of dermatology patients to demonstrate AI assistance to dermatologist. To evaluate generative performance analysis of HistoGPT, authors have used four semantic-based machine learning metrics as well as two blinded domain expert evaluations. Authors have also evaluated all models using traditional syntax-based measures of BLEU-4, ROUGE-L, METEOR, and BERTscore.
3- **HistGen** (Zhengrui Guo et al. 2024) has also been evaluated with WSI report generation, cancer subtyping, and survival analysis.
4- **ProveGigaPath** (H. Xu et al. 2024) evaluated on novel fine-tuned image-report alignment leveraging zero-shot cancer subtyping and zero-shot gene mutation prediction.
5- **THREADS** (Vaidya et al. 2025) introduces "molecular prompting", a new class of assessment to evaluate transfer and generalization without supervision. To gain more insights, clustering capabilities of the latent space measured via Rand index and mutual information, and tSNE visualization.
6- The evaluation methods and frameworks for multimodal models are all based on additional multiple fine tuning and adaptation steps making the evaluating generalist AI foundational competence and emergence specialized competence more challenging and complicated.

### 4.2.4. Evaluation of AI Copilots

Here we embrace downstream tasks on which AI copilot and AI assistants have been evaluated, all of which serve extrinsic evaluation.

**PathChat** (M. Y. Lu, Chen, Williamson, Chen, Zhao, et al. 2024)—a multimodal generative AI copilot for human pathology leverages vision only foundation model—**UNI** (R. J. Chen et al. 2024) that goes through vision-language pretraining similar to **CONCH** (M. Y. Lu, Chen, Williamson, Chen, Liang, et al. 2024)—vision-language foundation model, further connected to **Llama 2** to form a complete multi-model large language model (MLLM) architecture, which has been further finetuned using a curated dataset of over 450,000 instructions. The generic capabilities of PathChat has been evaluated on **PathQABench**—a high quality benchmark for expert-curated pathology questions on representative high-resolution ROI images from 105 H&E WSI in a zero-shot transfer setting and its performance have been compared with both **LLaVA**, and LLaVA-Med5 , and ChatGPT-4 (powered by GPT-4V).

More specifically, it has been evaluated on **multiple-choice diagnostic questions** covering 54 diagnoses from 11 different major pathology practices and organ sites, and **open-ended questions** targeting a broad spectrum of topics including microscopy image description, histologic grade and differentiation status, risk factors, prognosis, treatment, diagnosis, IHC tests, molecular alterations and other tests. For multiple choice questions (MCQs), the evaluation has been conducted in two setting, once PathChat has been queried with an image only and in the second setting it has been queried by an image with clinical context that included patient age, sex, clinical history and radiology findings are included with the histology image for the clinical case. For evaluation on 260 open-ended questions, a panel of seven pathologists were recruited to assess the model responses.

To explore further use cases, **PathChat** (M. Y. Lu, Chen, Williamson, Chen, Zhao, et al. 2024) has been evaluated to complement quantitative evaluation of **PathQABench** by a follow-up from users in the form of interactive, multi-turn conversations. It demonstrated its ability to summarized key morphological features in an histology image, reasonably infer the primary origin of the tumor based on clinical context, can potentially help by guiding IHC interpretations, can attempt to follow well-known guidelines on tumor grading, describe tumor tissue and cell morphology, infer a diagnosis and correctly suggest potential IHC findings grounded in relevant background knowledge about the suspected malignancy, can potentially be consulted to perform human-in-the-loop differential diagnosis that may require several rounds of an IHC workup.

**PathAsst** (Sun, Zhu, et al. 2024) integrates custom-trained **PathCLIP**—a specialized of CLIP for pathology, and **Vicuna-13B—**an LLM component for enhanced pathology analysis. **PathAsst** (Sun, Zhu, et al. 2024) went through two-phase training procedure, first phase to aligns the vision encoder and LLM using **PathInstruct** dataset whereas the second phase to generate higher-quality and more detailed responses on the data from books within the PathInstruct. **PathCLIP** has been evaluated on zero-shot classification and cross-modal retrieval validation. **PathAsst** (Sun, Zhu, et al. 2024) has been evaluated on **PathVQA** dataset (X. He et al. 2021) that contains open-ended questions typically beginning with what, where, and when, as well as close-ended questions requiring yes/no responses. To demonstrate PathAsst's robust capabilities in handling complex pathology tasks such as—**liquid-based cytology (LBC) cell generation** by invoking cell generation model, **positive cell counting** by invoking the PD-L1 detection model for assistance, and **interpret pathology images independently** which has been compared with LlaVA and MiniGPT-4.

Text-image aligned **PathGen** (Sun, Zhang, et al. 2024) has been evaluated on downstream tasks of zero-shot image classification, few-shot image classification with linear probing, and WSIs classification with MIL. To evaluate **PathGen-LlaVA's** multimodal description generation capabilities, PathMMU test set has been created comprising multimodal multi-choice QAs and over 100K rounds of dialogue data and benchmarked their performance against GPT-4V, Gemini-Pro Vision, Qwen-VL-Max, as well as previous pathology-specific LMMs such as LLaVA-Med and Quilt-LLaVA.

**SlideChat** (Y. Chen et al. 2024)—a vision-language whole slide assistant that consists on patch-level encoder (CONCH), the slide level encoder (LongNet), the multimodal projector, and the large language model. SlideChat has been evaluated for its capabilities of WSI captioning, visual question answering, in zero-shot setting, and processing whole-slide images through SlideBench-Caption, SlideBenchVQA (TCGA), SlideBenchVQa (BCNB), and VQA-WSI. SlideBenchVQA (TCGA) comprised histological changes, cytomorphological characteristics, tumor characteristics, tissue architecture and arrangement, prognostic assessment, risk factors, biomarker analysis, treatment guidance, differential diagnosis, grading, staging, disease detection and classification. SlideBenchVQA (BCNB) comprised tumor types, histological grading, molecular subtype, ER, PR and HER2 status prediction tasks. SlideChat has also been evaluated on closed-set VQA pairs from the public WSI-VQA (P. Chen et al. 2024). To assess the PathChat model interpretability patch-level attention scores has been calculated by the correlation between the text output and specific image patches. This allowed identifying the most significant patches associated to the model's response generation. A text generation examples confirms the association of patches of high attention scores to an increased nuclear-to-cytoplasmic ratio, hyperchromatic nuclei, and prominent nucleoli. Another examples confirms the response of dense collagen deposition and reduced cellularity associated to the highly attentive patches.

1- The generic capabilities of AI-copilot—**PathChat** (M. Y. Lu, Chen, Williamson, Chen, Zhao, et al. 2024) has been evaluated on **PathQABench,** that contained **multiple-choice diagnostic questions** covering 54 diagnoses from 11 different major pathology practices and organ sites, and **open-ended questions** targeting a broad spectrum of topics including microscopy image description, histologic grade and differentiation status, risk factors, prognosis, treatment, diagnosis, IHC tests, molecular alterations and other tests.
2- **PathChat** (M. Y. Lu, Chen, Williamson, Chen, Zhao, et al. 2024) has been evaluated to complement quantitative evaluation of **PathQABench** by a follow-up from users in the form of interactive, multi-turn conversations.
3- As reported by the authors of **PathChat** (M. Y. Lu, Chen, Williamson, Chen, Zhao, et al. 2024), the MLLM validation lacks evaluation of illumination and effectiveness of capturing certain nuances specific to pathology such as—determining when to seek further contextual information, test results and institutional-specific guidelines; and when certain morphologically similar diseases cannot be ruled out.
4- **PathChat** (M. Y. Lu, Chen, Williamson, Chen, Zhao, et al. 2024) cannot processes entire WSI. And retrospective training which may be subject to outdated knowledge and information leads to factually inaccurate responses from the mode.
5- **SlideChat** is the first vision-language AI assistant that processes entire WSI. Therefore, SlideChat can assist in clinical applications which necessitates processing entire slide or specimen.
6- SlideChat demonstrates some advanced and novel capabilities which include understanding histological changes, cytomorphological characteristics, tumor characteristics, tissue architecture and arrangement, prognostic assessment, risk factors, biomarker analysis, treatment guidance, and differential diagnosis.
7- **PathAsst** (Sun, Zhu, et al. 2024) has been evaluated on **PathVQA** dataset (X. He et al. 2021) that contains open-ended questions typically beginning with what, where, and when, as well as close-ended questions requiring yes/no responses.
8- To demonstrate **PathAsst's** (Sun, Zhu, et al. 2024) robust capabilities in handling complex pathology tasks such as—**liquid-based cytology (LBC) cell generation** by invoking cell generation model, **positive cell counting** by invoking the PD-L1 detection model for assistance, and **interpret pathology images independently** which has been compared with LlaVA and MiniGPT-4.
9- **PathGen** (Sun, Zhang, et al. 2024) evaluation on a new benchmark (**PathMMU** test set) comprising multimodal multi-choice QAs and over 100K rounds of dialogue data

Finally, **evaluation design must evolve alongside** the new demands. Evaluations that **go beyond standard metrics like accuracy, incorporating robustness, fairness, resources, costs, and even environmental impact are critical** (Ray 2023). Human-in-the-loop evaluation, for example, could provide deeper insights into generative models' capabilities, capturing subtleties that automated metrics might miss (Zhong et al. 2023). Through thoughtful design, evaluations can better reflect the true capabilities and limitations of foundation models, ultimately guiding both progress and decision-making in AI systems.

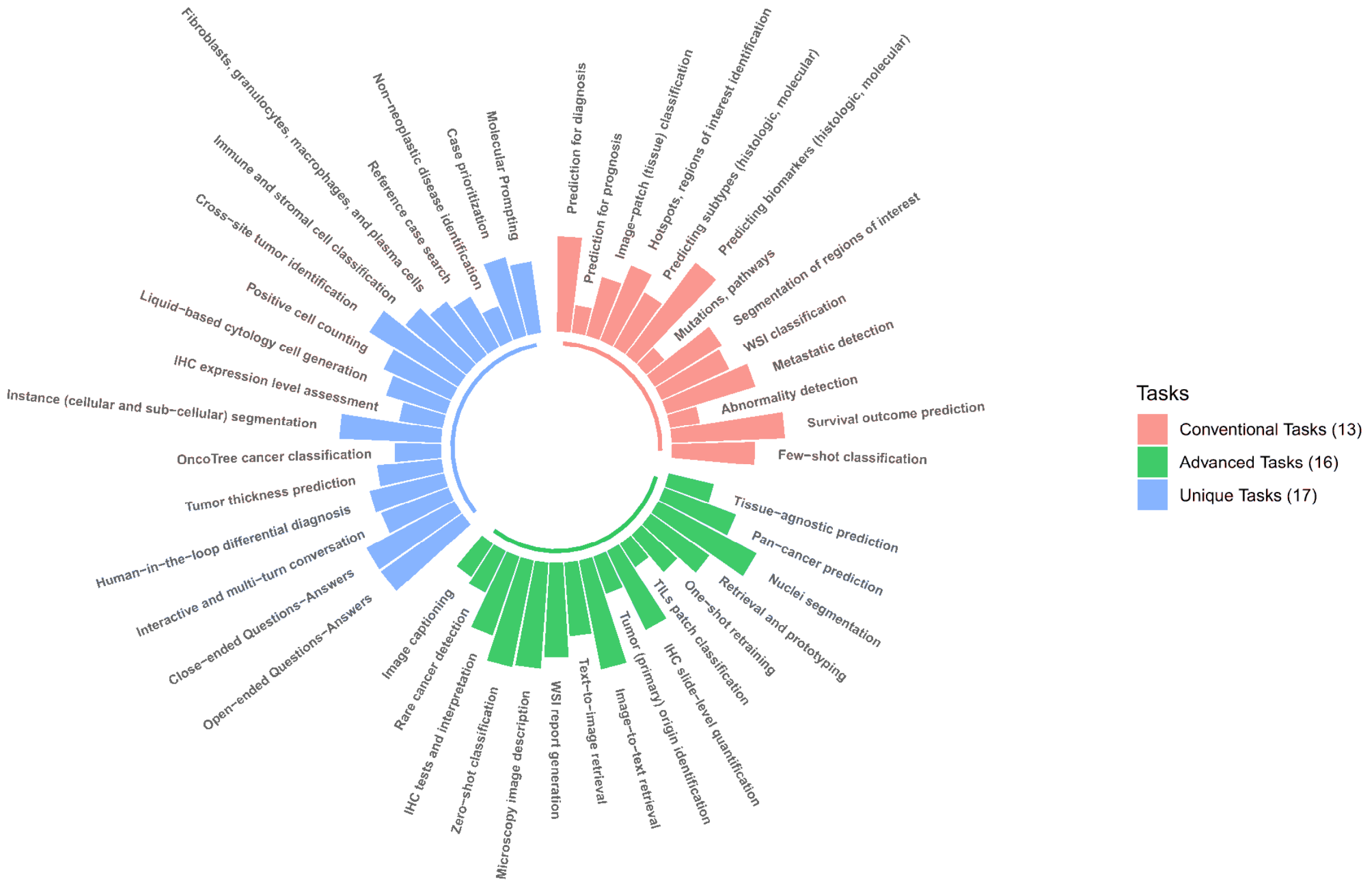

*Figure 6. Foundation model adapted to perform conventional, advanced, and unique computational pathology tasks*

In Figure 6, we categorized all downstream tasks performed by various foundation models as conventional, advanced and unique. The “conventional” here refers to traditional predictive analysis tasks that have been performed with image-only modalities, whereas denser and more detailed image only tasks and those which required image and text alignment during training emerged as “advanced” tasks for label-free self-supervised learning frameworks. The “unique” here refers to novel and specific tasks demonstrated by individual self-supervised foundation models or AI-copilots, uniquely.

## 4.3. Computational Pathology Advance

Over the past decade, computational pathology has advanced the diagnosis of various pathology tasks, including cancer detection (Campanella et al. 2019; Bilal, Tsang, et al. 2023), triaging and prescreening (Bilal, Tsang, et al. 2023; Graham, Minhas, et al. 2023), tissue classification (Kather et al. 2016; Javed et al. 2020), molecular pathway prediction (Bilal et al. 2021; Kather et al. 2019), genetic mutation identification (Kather et al. 2020; Saldanha et al. 2023), and the assessment of patient outcomes (Wahab et al. 2023; W. Lu et al. 2023) and treatment responses (Atallah et al. 2023; Hoang et al. 2024). This section explores how foundation models have excelled in these traditional downstream tasks and pioneered novel applications to reshape clinical research.

### 4.3.1. Cancer Detection and Subtyping

**Pan-Cancer Detection:** Foundation models have demonstrated remarkable success in pan-cancer detection including rare cancer types, models like **Virchow** (Vorontsov et al. 2024), **CTransPath** (Xiyue Wang et al. 2022a)**, UNI** (R. J. Chen et al. 2024)**, Virchow2** (Zimmermann et al. 2024) and **Virchow2G** (Zimmermann et al. 2024) setting new benchmarks. UNI (R. J. Chen et al. 2024) uniquely contributes in handling complex tasks such as **classifying up to 108 cancer types based on the OncoTree system**, surpassing previous models like CTransPath (Xiyue Wang et al. 2022a) and REMEDIS (Azizi et al. 2023) in tissue classification accuracy across various magnifications. **TITAN** (T. Ding et al. 2024) also advances in a slide-level OncoTree code classification task with 46 classes and dysplasia, IFTA status, and Gleason grading. Virchow advances in tissue-agnostic cancer detection and subtyping, H-Optimus-0 in pan-cancer tissue classification, and Phikon in histological type prediction, and CHIEF in cancer cell detection, and tumor origin identification. Virchow2, have achieved state-of-the-art performance in cancer detection and subtyping, in 12 tile-level tasks and in both in-domain and out-of-domain benchmarks, and ranks 1st Eva[c] leaderboard currently. Their scale, combined with domain-inspired adaptations to the DINOv2 training algorithm, enables these models to handle diverse cancer types and outperform traditional approaches.

Models like **PathGen-1.6M** (Sun, Zhang, et al. 2024) have introduced zero-shot transfer capabilities, enabling accurate subtyping of BRCA, NSCLC, and RCC without requiring additional labels. PathGen-1.6M's maintained an average median zero-shot accuracy of 70.2% across diverse cancer subtyping tasks. Additionally, **PRISM** (Shaikovski et al. 2024), designed for slide-level analysis, demonstrated proficiency in tasks like zero-shot cancer detection, subtyping, and biomarker prediction. **UNI** (R. J. Chen et al. 2024) also introduces few-shot class prototypes, enabling prompt-based slide classification and robust generalization for subtyping diverse cancers, further strengthening its role as a foundation model for challenging diagnostic workflows.

### 4.3.2. Tissue Classification and Image Analysis

**Tissue Subtypes and Composition:** The **HistoGPT** (Tran et al. 2024) model demonstrates the lead of foundation models in tissue classification in dermatology with zero-shot classification of tumor subtypes and thickness, generating detailed reports on tissue subtypes, cellular composition, and potential diagnoses. REMEDIS and Hibou-B/L demonstrate advance in pan-cancer tumor-infiltrating lymphocyte detection. PathoDuet and PathChat contribute uniquely in cross-stie tumor identification and describing microscopy image, respectively.

**Instance and nuclei segmentation:** BROW and Kaiko-ai advances self-supervised learning applied to nuclei segmentation on CoNSeP dataset. RodulfV excels in several advanced and unique contributions including cell classification especially for immune and stromal cells such as fibroblasts, granulocytes, macrophages, and plasma cells. Hibou-B/L promised advances are cell-level tasks – nuclei segmentation and classification in PanNuke dataset. PLUTO uniquely contributes to instance segmentation at cellular and subcellular level.

**Whole-Slide Image (WSI) Classification:** BLIP-based **PathAlign** (F. Ahmed et al. 2024) and **mSTAR** (Y. Xu et al. 2024) have set new benchmarks in WSI classification and multimodal survival analysis. PathAlign's performance in image-to-text retrieval, case prioritization, and WSI classification, with AUROCs ranging from 0.945 to 0.987, has been highly rated by pathologists, while mSTAR has demonstrated superior performance in zero-shot and few-shot slide classification, outperforming models like UNI (R. J. Chen et al. 2024) and ResNet50 (K. He et al. 2016). **Virchow2's** (Zimmermann et al. 2024) mixed-magnification capabilities offer slide-level analysis, tissue classification and molecular prediction. **OmniScreen** (Y. K. Wang et al. 2024) extends this by predicting a wide range of genomic biomarkers from WSIs, offering robust molecular insights with a mean AUROC of 0.89 across the 15 most common cancers.

### 4.3.3. Molecular Pathway Prediction and Genetic Mutation Identification

**Molecular Subtyping and Mutation Prediction:** The role of foundation models extends to molecular subtyping and mutation prediction, as demonstrated by **Madeleine** (Jaume, Vaidya, et al. 2024)**, Prov-GigaPath** (H. Xu et al. 2024)**, and TITAN** (T. Ding et al. 2024)**.** Prov-GigaPath (H. Xu et al. 2024)achieved state-of-the-art performance in 25 out of 26 digital pathology tasks, including zero-shot subtyping and mutation prediction. Its superior performance in the prediction of LUAD-specific five-gene mutations, highlights the quality of its pretraining data.

**OmniScreen** (Y. K. Wang et al. 2024) simultaneously predicts 1,228 genomic biomarkers across 70 human cancers, identifying 80 high-performing biomarkers. It has identified 391 genomic alteration biomarkers with AUC > 0.75 across the 15 most common cancer types, revealing 40 histologic biomarkers associated with phenotype-genotype correlations and 58 treatment-associated biomarkers predictive of response to FDA-approved drugs. This unified model surpasses supervised models in predicting molecular pathways, DNA repair defects, and genomic instability, making it a powerful tool for precision medicine.

### 4.3.4. Report Generation and Diagnostic Assistance

**Automated Report Generation:** Foundation models, **mSTAR** (Y. Xu et al. 2024) generates comprehensive reports, coupled with its diagnostic capabilities and molecular prediction accuracy, has made it a unique tool in precision diagnosis. **PRISM's** (Shaikovski et al. 2024) **and TITAN** (T. Ding et al. 2024) **demonstrate** capabilities in generating clinical reports, alongside their performance in cancer detection with rare cancers, and biomarker prediction highlight the potential of foundation models in automating diagnostics.

[c] https://kaiko-ai.github.io/eva/main/

**Diagnostic Accuracy and Assistance: PathAsst** (Sun, Zhu, et al. 2024)**, PathChat** (M. Y. Lu, Chen, Williamson, Chen, Zhao, et al. 2024)**, and SlideChat** integrate AI copilot capabilities, providing real-time assistance to pathologists by handling complex queries and zero-shot classification. PathAsst's performance in zero-shot classification and cross-modal retrieval tasks has made it a go-to model for complex diagnostic scenarios. PathChat, with its unique ability to handle multiple-choice questions (MCQs), interactive and multi-turn conversation, description, guardrails, human-in-the-loop diagnosis and open-ended queries, has outperformed models like GPT-4V and LLaVA in diagnostic accuracy, particularly when both images and clinical context are available.

### 4.3.5. Pathology-Specific Innovations

**Rare Disease Classification:** foundation models have also shown extraordinary promise in the classification of rare diseases. **CONCH** (M. Y. Lu, Chen, Williamson, Chen, Liang, et al. 2024) has demonstrated state-of-the-art performance across diverse benchmarks, including the recognition of up to 30 categories of brain tumors. Its success in applying zero-shot and few-shot learning to rare disease classification underscores the transformative potential of foundation models in pathology. RoduflV has contributed to reference case searching including rare oncological and non-neoplastic diseases.

**Multiple staining capabilities:** Several foundation models have demonstrated capabilities of handling multiple staining inputs like IHC and immunostaining. RodulfV contributed in pan-indication H&E-based tumor microenvironment (TME) characterization, and pan-indication immunohistochemistry biomarker evaluation. PathDuet unique contributions include IHC slide-level quantitative diagnostics, IHC expression level assessment, cross-site tumor identification.

**Metastatic and Survival analysis:** Phikon, HIPT, REMEDIS, CHIEF, HistGen, TITAN, and mSTAR excel in survival analysis. mSTAR uniquely contributes in multi-modal survival analysis gaining insights from WSIs, report, and RNA-seq. H-Optimous-0, Phikon and REMEDIS also perform metastatic analysis.

**Interpreting Complex Pathology Data:** The application of foundation models to complex pathology data is shown by models like **PathAsst** (Sun, Zhu, et al. 2024), LLaVA, and MiniGPT-4, which offer cell-level insights, PD-L1 detection, liquid-based cytology (LBC) cell generation, and positive cell counting, significantly enhancing the interpretation of pathology images.

These novel applications highlight the profound impact of foundation models in computational pathology.

## 4.4. Benchmarking Pathology Foundation Models

Recent advancements in pathology foundation models have driven numerous benchmarking efforts to assess their performance across a wide array of clinical and computational tasks. These benchmarks provide crucial insights into the strengths, limitations, and potential of different foundation models to revolutionize digital pathology. Below, we review twelve key benchmarking studies, organized by application and task, highlighting the foundation models involved, their performance, and any novel applications or extraordinary findings.

### 4.4.1. Evaluating the Versatility and Generalizability of Foundation Models

**Pathological Slide Analysis**

**OpenMEDLab** (Xiaosong Wang et al. 2024) is an open-source platform, includes two notable pathology foundation models—PathoDuet (Hua et al. 2024) and BROW (Y. Wu et al. 2023a)—which are pretrained using self-supervised learning (SSL) and self-distillation, respectively. PathoDuet focuses on the analysis of H&E and IHC-stained slides, utilizing local features and pretext tokens for global feature extraction, while BROW enhances WSI analysis. These models were benchmarked on datasets including TCGA, CAM17[d], and private datasets, demonstrating inspiring and competitive results across various downstream tasks.

[d] https://camelyon17.grand-challenge.org/Data/

**High-Resolution Slide-Level Aggregation**

**BeyondMIL-WSIModeling** (Campanella, Fluder, et al. 2024) introduced a novel approach to jointly train both a tile encoder and a slide-aggregator fully in memory and end-to-end at high resolution. This method has shown remarkable performance in tasks such as LUAD EGFR mutation prediction and breast cancer detection. The fine-tuned model achieved a validation AUC of 0.82, outperforming the Attention-based MIL model (Ilse, Tomczak, and Welling 2018), which reached an AUC of 0.76.

**Slide-Level MIL Classification**

**MILmeetFM** (C. Lu, Xu, Wang, Shi, Qin, et al. 2024) systematically compared six foundation models—CTransPath (Xiyue Wang et al. 2022a), PathoDuet (Hua et al. 2024), PLIP (Huang et al. 2023), CONCH (M. Y. Lu, Chen, Williamson, Chen, Liang, et al. 2024), and UNI (R. J. Chen et al. 2024)—and six recent multiple instance learning (MIL) methods across five WSI tasks, including breast cancer grading and biomarker status prediction. The study found that foundation models trained with more **diverse histological images** provided better patch-level feature embeddings and feature **re-embedding online** for slide-level aggregations improved WSI classification performance, emphasizing the value of continuous model adaptation and refinement.

**Consistency Assessment and Flexibility Assessment**

**Lee et al.** (Jeaung Lee et al. 2024) **benchmarked** the performance of four foundation models – CTransPath (Xiyue Wang et al. 2022a), Lunit-Dino (Kang et al. 2023), Phikon and UNI (R. J. Chen et al. 2024) – across 14 datasets spanning 5 organs under two different evaluation scenarios. In the consistency assessment, the authors evaluated the models' ability to adapt to different datasets within identical tasks using five distinct fine-tuning approaches: linear probing, full fine-tuning, partial fine-tuning, parameter-efficient fine-tuning (PEFT), and training from scratch. In the Flexibility Assessment, they examined the models' generalizability to new tasks and datasets under data-limited conditions. They employed five few-shot learning (FSL) approaches: ProtoNet, MatchingNet, Baseline, Baseline++, and K-Nearest Neighbors (KNN) across three different adaptation scenarios—near-domain, middle-domain, and out-domain. The study demonstrated that UNI consistently outperformed other models in both assessment scenarios. Additionally, pathology foundation models demonstrated effective generalization across datasets and tasks, particularly when adapted using PEFT or FSL methods like Baseline++.

### 4.4.2. Benchmarking Across Computational Tasks

**Pretraining of ViT Models**

**CPathatScale** (Campanella et al. 2023) leveraged the largest academic pathology dataset to date, comprising over 3 billion images from 423,000 WSIs, to compare the pretraining of ViT models using masked autoencoders (MAE) and self-distillation models (DINO). The DINO algorithm consistently outperformed other models, achieving high AUCs over 90% in detection tasks. However, **biomarker prediction tasks showed more variability**, and outcome prediction tasks yielded poor results across all tested models. This benchmark underscores the importance of selecting appropriate algorithms and tailoring models to address specific clinical challenges, when dealing with large-scale datasets.

**Universal Knowledge Distillation (UKD) and Model Generalization**

The **Generalizable Pathology Foundation Model (GPFM)** benchmarked by **UKD** (Ma et al. 2024) evaluated the performance of off-the-shelf foundation models across six distinct clinical task types, including WSI classification, ROI classification, survival modeling, retrieval, visual question answering (VQA), and report generation. GPFM, utilizing expert and self-knowledge distillation, achieved top performance across 29 out of 39 tasks, with an impressive average rank of 1.36. This far outpaced the second-best model, UNI (R. J. Chen et al. 2024), which attained an average rank of 2.96. GPFM's success demonstrates the significant **benefits of integrating expert knowledge with foundation models**, enhancing their robustness and generalizability across a diverse range of clinical tasks.

**Embedding Aggregation Methods**

**Benchmarking Embedding Aggregation Methods** (S. Chen et al. 2024) provided a comprehensive evaluation of widely used aggregation methods assessing the embeddings generated by various foundation models. This study covered nine clinically relevant tasks, including diagnostic assessment, biomarker classification, and outcome prediction, using private datasets related to breast cancer, inflammatory bowel disease, and several lung cancers. Key findings included:

1. **Domain-specific (histological image-based) foundation models** consistently outperformed ImageNet-based models across all aggregation methods and **spatial-aware aggregators** significantly enhanced performance when using ImageNet pre-trained models but not when using domain-specific foundation models.
2. **No single model excelled across all tasks**, and the expected superiority of spatially aware models was not uniformly observed.

**Self-Supervised Learning (SSL) in Pathology**
**Giga-SSL** (Lazard et al. 2023) introduced a self-supervised learning (SSL) method designed for TCGA to learn WSI representations without any annotations. When applied to a variety of downstream tasks, Giga-SSL substantially improved classification performance over fully supervised alternatives, particularly in tasks involving small datasets. In a pan-cancer setting, Giga-SSL **doubled the number of mutations predictable from WSIs** compared to previous methods.

**Benchmarking SSL on DPath** (Kang et al. 2023) conducted a principled largest-scale study comparing SSL methods in the context of pathology, focusing on four image classification tasks and one challenging **nuclei instance segmentation**. The unlabeled raining dataset comprises 36,666 WSIs including 20,994 from TCGA and 15,672 from TULIP (proprietary). This study revealed that **standard SSL with linear fine-tuning and low-label regimes**, consistently outperformed ImageNet pre-training, **in nuclei instance segmentation**.

### 4.4.3. Benchmarking Foundation Models on Clinically Relevant Tasks

**Clinical Relevance and Dataset Composition**
**A clinical Benchmark of Pathology foundation models** (Campanella, Chen, et al. 2024) systematically assessed public pathology foundation models across a variety of clinically relevant tasks using private MSHS cohorts. This study compared CTransPath (Xiyue Wang et al. 2022a), UNI (R. J. Chen et al. 2024), Virchow (Vorontsov et al. 2024), Prov-GigaPath (H. Xu et al. 2024), and ResNet50 (K. He et al. 2016). The findings were noteworthy:

1. Dino and DinoV2-based foundation models like UNI (R. J. Chen et al. 2024) and Prov-GigaPath (H. Xu et al. 2024) performed better in disease and biomarker detection.
2. ImageNet-based encoders and CTransPath (Xiyue Wang et al. 2022a) consistently underperformed.
3. The pretraining dataset was identified as a critical factor in model performance, particularly in disease detection tasks, where there was no evidence that downstream performance scaled with model size.
4. In biomarker prediction, larger models performed better for NGS lung tasks but not for IHC breast cancer tasks, with no evidence of improved performance associated with higher computational costs.

**Ovarian Cancer Subtype Classification**
Breen et al. (Breen et al. 2024) evaluated 17 feature extraction models for ovarian cancer subtype classification, including ImageNet-pretrained models (ResNet50, ResNet18, and ViT-L) and histopathology foundation models (RN18-Histo, RN50-Histo, Lunit[139], **CtransPath**[37], Hibou-B[39], Phikon, Kaiko-B8, GPFM[137], **UNI**[13], Hibou-L[39], Virchow, Virchow2-CLS, **H-Optimus-0**[51], and **Prov-GigaPath**[46]). The study demonstrated that histopathology foundation models significantly outperformed the ImageNet-pretrained models, with 13 out of 14 foundation models consistently exceeding the performance of the baselines. Among them, H-optimus-0 achieved the best overall performance. While larger models and those pretrained on larger datasets generally demonstrated superior performance, **UNI**[13] and GPFM models delivered results that exceeded expectations given their moderate pretraining dataset sizes.

A recent **study** (Neidlinger et al. 2024) **benchmarked ten histopathology foundation models across 13 patient cohorts**, involving 6,791 patients and 9,493 slides from lung, colorectal, gastric, and breast cancers, focusing on weakly-supervised tasks such as biomarker prediction, morphological property analysis, and prognostic outcome estimation.

1. The vision-language foundation model CONCH (M. Y. Lu, Chen, Williamson, Chen, Liang, et al. 2024) outperformed vision-only models, excelling in 42% of tasks, highlighting the effectiveness of integrating multimodal data in pathology.
2. Models trained with distinct cohorts learned complementary features for the same predictive tasks. An ensemble of these models surpassed CONCH (M. Y. Lu, Chen, Williamson, Chen, Liang, et al. 2024) in 66% of tasks, establishing a new benchmark for performance.

3. A critical insight was that data diversity is often more beneficial than data volume, indicating that a varied training dataset can lead to more robust models.

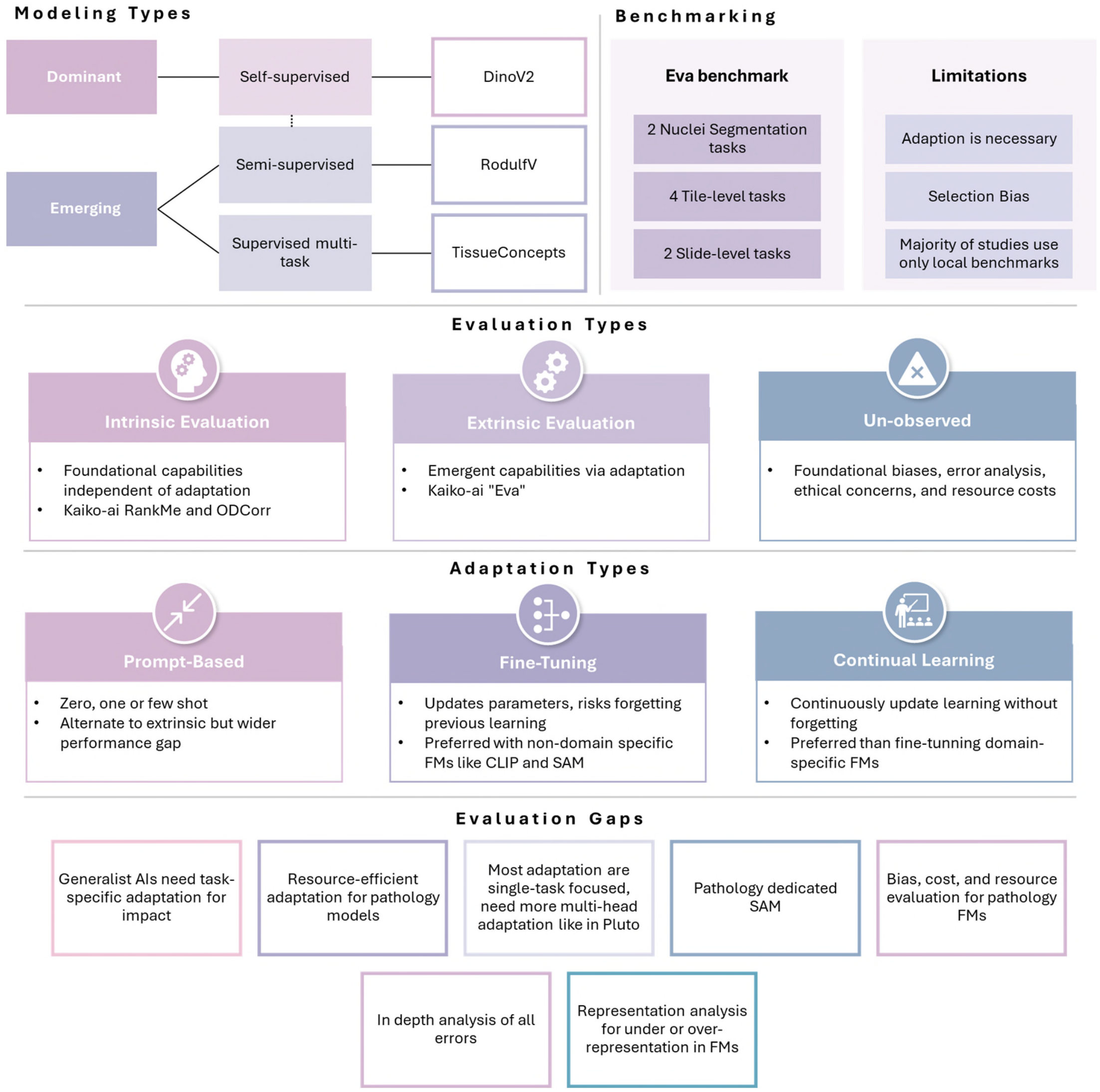

*Figure 7. Understanding Pathology Foundation Models Evaluation and Adaptation*

The study highlights the potential of generalist AI models in specialized pathological tasks. However, their effectiveness may be limited by the difficulties of pathology, indicating the need for further exploration of their adaptability in this domain. The findings raise concerns about the efficacy of stand-alone training for large general foundation models like Virchow (Vorontsov et al. 2024) and UNI (R. J. Chen et al. 2024) in specialized tasks. Hybrid approaches combining generalist models with task-specific refinements are essential to enhance performance and capture critical features unique to the domain. The ability of complementary learning to outperform CONCH (M. Y. Lu, Chen, Williamson, Chen, Liang, et al. 2024) suggests that multimodal models are still evolving. Comprehensive evaluation across diverse datasets is crucial to identify limitations and improve these models, paving the way for more robust tools in digital pathology.

These insights underscore the importance of refining AI models for pathology through specialized training and continuous evaluation to better address the complexities of clinical tasks in the future.

### 4.4.4. Critical benchmarking findings

The benchmarking analysis highlights the remarkable potential of generalist AI capabilities in computational pathology but also reveals significant limitations in specialized tasks. Generalist foundation models often require retraining or extensive fine-tuning to reach state-of-the-art performance in specific applications. Notably, while existing foundation models excel at certain task types, they struggle to effectively handle the full range of clinical tasks. For example, in the MSI prediction task within Giga-SSL (Lazard et al. 2023), an AUROC of 0.75 or below reflects underperformance compared to supervised models. The performance gains of foundation models, especially in clinical-grade evaluations, are notable but not yet transformative or ready for widespread clinical adoption.

Similar challenges are observed in the variability of biomarker prediction outcomes in CPathatScale (Campanella et al. 2023), the need for fine-tuning in MILmeetFM (C. Lu, Xu, Wang, Shi, Qin, et al. 2024), and the requirement for task-specific training of Virchow2 in PanCancerNGS (Y. K. Wang et al. 2024). Moreover, these studies underscore the importance of aggregation in computational pathology; even with advanced foundation models, full-resolution, all-in-memory, end-to-end slide modeling remains essential for achieving significant performance improvements and realistic, data-driven results. These comprehensive benchmarking efforts demonstrate the impressive capabilities of current pathology foundation models while also highlighting the importance of evaluating their robustness and generalization. Evaluations such as robustness to image corruption, as seen in PathCLIP (Zheng et al. 2024), are crucial for assessing the reliability of generalist models under real-world conditions.

Benchmarking foundation models on novel tasks such as zero-shot and few-shot learning, multiple-choice questions (MCQs), open-ended Q/A, and image-to-text and text-to-image retrieval is crucial for assessing their generalization capabilities. By refining these models and exploring innovative approaches tailored to clinical needs, digital pathology and healthcare can progress toward delivering more accurate and versatile solutions, ultimately improving patient outcomes. Figure 7 catalogs findings of pathology foundation models' evaluation.

Foundation models, serving as central repositories of medical knowledge, can be interactively queried and updated by healthcare professionals and researchers, enabling efficient adaptation for tasks like patient Q&A apps or clinical trial matching (Bommasani et al. 2022). However, their implementation comes with challenges, including the integration of multimodal data and adherence to ethical and legal standards, particularly in privacy, safety, and explainability (Bommasani et al. 2022). A dedicated benchmarking process is essential to ensure the safe and effective use of interactive foundation models, such as AI assistants (Sun, Zhu, et al. 2024) and co-pilots (M. Y. Lu, Chen, Williamson, Chen, Zhao, et al. 2024), in healthcare. Rigorous evaluation will be key to advancing these models in a way that meets clinical demands while ensuring ethical and responsible deployment. These benchmarking studies have constructed their own local benchmark limiting direct and fair comparison which is necessary to demonstrate real scientific advance.

## 4.5. Societal Impact

Foundation models have redefined traditional tasks in computational pathology, while their advancements in novel applications promise to revolutionize clinical diagnostics and research, contributing to more equitable and efficient healthcare outcomes.

Foundation Models have potential to make a significant societal impact, driving a paradigm shift in AI research and deployment. A clinical-grade AI co-pilot—**Pagie Alba**[e] built on OmniScreen and Virchow 2 for the research use only to integrate real-time insights for pathologists, oncologists, and clinical teams. Another research-use AI agent for biomedical image analysis— **Judith**[f] built on UNI, CONCH, and PathChat to accelerate scientific discovery. Emerging of AI copilots – **Paige Alba** and **Judith** can make significant societal impact in the world of pathology. However, their potential remains underutilized due to limited accessibility and least understood because of evaluation difficulties as detailed in Section 4 (Evaluation). Broadening the availability of these and such tools could democratize their use, fostering innovation and improving outcomes in precision medicine and pathology research and development.

Moreover, AI agents and copilots are not free from potential harm arising from intrinsic, adaptation, and representational biases to yield inequitable outcomes. Intrinsic and adaptation biases are the byproducts of training and adaptation data sources, modeler diversity, architectures, and objectives. Extrinsic harms impact the user experiences on the context of downstream use cases. These harms link to representational biases— misrepresentation, under and overrepresentation— and performance disparities—for individuals, population groups, and subgroups (Bommasani et al. 2022).

---

[e] https://www.businesswire.com/news/home/20240905686688/en/Paige-Unveils-Alba-The-AI-Clinical-Grade-Co-Pilot-Set-to-Revolutionize-Diagnostics-and-Treatment-in-Pathology-and-Oncology
[f] https://modella.ai/judith.html

# 5. Adaptation Studies on Foundation Models

Recent advancements in both pathology and non-pathology foundation models have demonstrated that adapting image-only models, as well as text-guided image-based models, can significantly enhance performance in specialized computational pathology tasks. The adaptation of foundation models has become a practical and accessible area of research for many academic groups and labs. This section explores how efficiently adapting these models, with a focus on integrating generalized feature representations, can improve specialized tasks in computational pathology.

## 5.1. Adaptation of Pathology Foundation Models

**Segmentation by Factorization**
**Gildenblat et al.** (Gildenblat and Hadar 2024) introduce Segmentation by Factorization (F-SEG), an unsupervised segmentation method for pathology that generates segmentation masks from pre-trained deep learning models, including pathology foundation models, without additional training. F-SEG factorizes spatial features into segmentation masks and their associated concept features. By clustering features from deep learning models trained on TCGA, F-SEG creates robust tissue phenotypes for H&E images. The results show improved segmentation quality using pathology foundation models.

**Vision-Language-based Survival Analysis**
**Liu et al.** (P. Liu et al. 2024) propose a Vision-Language-based Survival Analysis (VLSA) paradigm to improve prognostic learning from histopathology whole-slide images (WSIs). VLSA leverages pathology vision-language foundation models, enhancing data efficiency and overcoming limitations of traditional multi-instance learning (MIL) frameworks. It incorporates prognostic language priors to guide the aggregation of visual features, compensating for weak supervision in MIL. VLSA introduces ordinal survival prompt learning and an ordinal incidence function to refine predictions, which are interpreted using Shapley values. Experiments across five datasets demonstrate its effectiveness for survival analysis in computational pathology.

**Ovarian Cancer Bevacizumab Treatment Response Prediction**
**Mallya et al.** (Mallya et al. 2024) leveraged foundation models and MIL to extract ovarian tumor tissue features for predicting bevacizumab response from WSIs. They have compared 6 dation models founCTransPath (Xiyue Wang et al. 2022a), Lunit-Dino (Kang et al. 2023), Phikon, PLIP (Huang et al. 2023), UNI (R. J. Chen et al. 2024), and Virchow and 3 MIL methds ABMIL, CLAM, and VarMIL and found founCTransPath (Xiyue Wang et al. 2022a) with CLAM achieve highest AURCO of 0.86. Their Survival models achieved statistical significance ($p < 0.05$) to stratify high- and low-risk cases among the aggressive subtype of high-grade serous ovarian carcinoma patients.

**Robustness to Image Corruption**
**PathCLIP** (Zheng et al. 2024) evaluates the robustness of PathCLIP across various corrupted images, including datasets of osteosarcoma and WSSS4LUAD. The study examined the impact of 11 types of corruption, such as brightness, contrast, defocus, and resolution. Key findings included certain corruptions, like hue, markup, deformation, defocus, and resolution, caused significant performance fluctuations in PathCLIP.

## 5.2. Adaptations of Non-pathology Foundation Models for Performance Gains

This section explores the adaptation of non-pathology foundation models, focusing on the integration of text-based reasoning and image analysis, as well as the application of SAM-based methods to improve segmentation and classification outcomes.

### 5.2.1. Open AI's CLIP adaptation for pathology tasks

**MI-Zero** (M. Y. Lu et al. 2023) is zero-shot transfer MIL-based framework contrastively aligns image and text models on gigapixel WSIs without requiring any additional labels. MI-Zero performed better than OpenAI's CLIP (Radford et al. 2021) and when trained it on ARCH (Gamper and Rajpoot 2021) data but not the when compared to fully supervised baseline Attention-based MIL (ABMIL) (Ilse, Tomczak, and Welling 2018) on full data.

**Marini et al.** (Marini et al. 2024) introduce a CLIP-based multimodal architecture to construct a robust biomedical knowledge representation from limited training WSIs and corresponding reports to tackle data scarcity.

**Qu et al.** (Qu, Yang, et al. 2024) present multi-instance prompt learning framework integrating image and text prior knowledge into prompts at both patch and slide levels.

**CPLIP** (Javed et al. 2024) by Sajid et al. introduces a new method to fine-tune the CLIP model without any ground truth annotations using a many-to-many contrastive learning method.

Lai et al proposed **CLIPath** (Lai et al. 2023) which introduces Residual Feature Connection (RFC) to effectively fuse the task-specific learning from the target domain and pre-trained CLIP by finetuning with a small amount of trainable parameters.

Shi et al. proposed **ViLa-MIL** (J. Shi et al. 2024) – a dual-scale vision language multiple instance learning framework that adapts the CLIP model for WSI classification in pathology under few-shot settings. The framework introduces learnable prototype vectors to cluster similar patch features in the image branch and employs a context-guided text decoder that incorporates image context in the text branch.

**Zhou et al.** (X. Zhou et al. 2024) curate a pathology knowledge tree – a comprehensive structured pathology knowledge base – having 50,470 informative attributes for 4,718 diseases from 32 human tissues. They adopted a CLIP-based visual-language pretraining approach, to project pathology-specific knowledge into latent embedding space via language model, and use it to guide the visual representation learning.

**Hu et al.** (Hu et al. 2024) present a novel cross-modal framework for histopathology image retrieval, addressing challenges in large-scale WSI analysis. The framework combines a WSI encoder for hierarchical region feature extraction with a prompt-based text encoder to capture fine-grained semantics from paired diagnostic reports.

**Nguyen et al.** (Nguyen, Vuong, and Kwak 2024) introduce TQx, a Text-based Quantitative and Explainable analysis framework that leverages a vision-language model for image-to-text retrieval. By employing a pre-trained vision-language model, TQx extracts relevant words from histopathology images, creating interpretable feature embeddings aligned with text descriptions.

**LILE** (Maleki and Tizhoosh 2022) introduced an iterative approach to retrieve cross-modal information – images and language – outperforming prior models such as IMRAM (H. Chen et al. 2020), CLIP (Radford et al. 2021), and PVSE (Y. Song and Soleymani 2019) on the ARCH (Gamper and Rajpoot 2021) dataset. Building on this, **LILE+H_DINO** (Maleki, Rahnamayan, and Tizhoosh 2024) extended the DINO framework by introducing scale harmonization through a novel patching technique. Evaluated on multiple datasets, this approach outperformed comparable methods in both patch and WSI classification by integrating H-DINO for vision, BioBERT (Jinhyuk Lee et al. 2020) for text feature extraction, and LILE (Maleki and Tizhoosh 2022) for feature alignment.

The **Visual Knowledge Search** (Lv et al. 2024) method focuses on image segmentation and paired text descriptions from 60 renal pathology books. It performs clustering analysis on image and text features and implements a retrieval system based on the semantic features of large models. A knowledge base of 10,317 renal pathology images with paired text was established, consisting of 9,342 pathological images, 975 IHC images, with sources split between Chinese (4,214) and English (6,103), covering tumor (2,038) and non-tumor (8,279) diseases.

**Qiu et al.** (Qiu et al. 2024) present prototype sampling for selecting optimal annotation regions in WSIs to enhance deep learning with limited annotation budgets.

**Ding et al.** (R. Ding et al. 2023) explore improving mitosis detection in cancerous tissue by introducing a tile-level mitosis classification pipeline using BLIP, framing the problem as both an image captioning and a visual question answering (VQA) task, incorporating metadata like tumor and scanner types as context.

**Meseguer et al.** (Meseguer, del Amor, and Naranjo 2024) introduced MI-VisionShot, a method for classifying WSIs in pathology using a few-shot adaptation of vision-language models. During training, MI-VisionShot used **PLIP** to select the most relevant patches based on text descriptions, creating a global representation for each slide. These global representations were then used to build class prototypes, allowing the model to classify new slides by comparing them to these prototypes without relying on text during testing.

### 5.2.2. Text-Guided foundation models for Image Classification

The adaptation of foundation models for image classification in computational pathology marks a significant advancement, particularly through the integration of textual reasoning and visual analysis. Models like **GPT-4V-DPath** (Ferber, Wölflein, et al. 2024) and generative classifiers such as **CITE** (Y. Zhang et al. 2023)**, CAMP** (Nguyen et al. 2024) and **GPC** (Nguyen and Kwak 2023) demonstrate the growing potential of these technologies in enhancing the accuracy and efficiency of pathology image analysis.

GPT-4V-DPath (Ferber, Wölflein, et al. 2024): By incorporating text-guided reasoning, GPT-4V-DPath leverages in-context learning (ICL) to improve performance across various tasks, including zero-shot and few-shot image classification. CITE (Y. Zhang et al. 2023): Another critical study demonstrated the utility of integrating text knowledge into foundation models adaptation to improve image classification in pathology. Integrating textual knowledge into foundation model adaptation can effectively replace traditional classification heads, enabling efficient training with as few as 1 to 16 slides per class.

These findings underscore the value of multimodal approaches in pathology, where combining visual data with textual reasoning can lead to superior performance, especially in data-scarce environments.

CAMP (Nguyen et al. 2024) (Continuous and Adaptive Learning Model in Pathology) and GPC (Nguyen and Kwak 2023) (Generative Pathology Classifier): Both models offer unified frameworks for diverse pathology tasks, including cancer grading, detection, and subtyping. CAMP, in particular, offers substantial reductions in computational time and storage requirements (up to 94% and 85%, respectively), making it a practical solution for large-scale pathology analysis. GPC, on the other hand, combines ConvNets with transformer-based language models to generate relevant labels with manageable computational complexity.

**PathM3** (Q. Zhou et al. 2024) is a multimodal, multi-task MIL framework designed to address challenges in aligning WSIs with diagnostic captions for histopathology. It uses a query-based transformer to align WSIs with captions, addressing redundancy in patch features with a MIL approach that captures correlations. PathM3 also overcomes the scarcity of WSI-level captions through multi-task joint learning.

The Fine-grained Visual-Semantic Interaction (**FiVE) framework** (H. Li et al. 2024) for WSI classification integrates localized visual patterns with a large language model to extract fine-grained pathological descriptions from non-standardized raw reports. This proof-of-concept standardizes pathology report data, significantly improving WSI classification by accurately localizing regions of interest. In few-shot experiments on the TCGA Lung Cancer dataset, FiVE achieved at least 9.19% higher accuracy than its counterparts highlights its potential for broader applications in computational pathology.

**Qu et al.** (Qu, Luo, et al. 2024) propose a two-level prompt learning MIL framework for pathology that integrates language prior knowledge. CLIP extracts instance features, which are aggregated into bag features via a prompt-guided pooling strategy. GPT-4 provides language knowledge at both instance and bag levels, incorporated into prompts for few-shot learning.

**Lin et al.** (Lin et al. 2023) propose a novel method for adapting pre-trained models to histopathology images in multiple instance learning (MIL), addressing domain shift from natural images. By using a prompt component to guide the pre-trained model in distinguishing differences between datasets, the approach enhances MIL performance.

Jaume et al. present **TANGLE** (Jaume, Oldenburg, et al. 2024), a multimodal pre-training strategy that enhances slide representation learning by leveraging gene expression profiles. Using modality-specific encoders aligned through contrastive learning, TANGLE achieves superior few-shot performance across independent breast, lung, and liver WSI datasets.

These models exemplify the potential of adapting foundation models in pathology that maintain high accuracy while managing resources efficiently. The success of these adaptations suggests a promising future for multimodal data integration in specialized pathology tasks.

### 5.2.3. SAM-Based Adaptations in Computational Pathology

The adaptation of the SAM for computational pathology has led to notable advancements in segmentation and classification tasks. Below are the key contributions and findings from three significant SAM-based adaptations:

**SAM-Path** (J. Zhang et al. 2023) – a fine-tuned version of SAM is designed specifically for semantic segmentation in pathology. It introduces trainable class prompts and employs a ViT Small encoder from the HIPT model, achieving significant performance gains on datasets like CRAG and BCSS. SAM-Path (J. Zhang et al. 2023).

**SAM for HistoPath** (Deng et al. 2023) adaptation evaluates SAM's performance in segmenting tumors, non-tumor tissues, and cell nuclei in histopathology images, using datasets like WSIs from skin cancer patients, ROIs related to Minimal Change Disease, and the MoNuSeg dataset. SAM excels in segmenting large, connected tissue structures but shows inconsistent performance in dense instance segmentation, a critical area in histopathology.

Evolving SAM further, **SAM-MIL** (Fang et al. 2024) incorporates spatial context into both segmentation and classification processes. It extracts comprehensive image-level information directly from WSIs and uses a SAM-Guided Group Masking strategy to address class imbalance issues. SAM-MIL (Fang et al. 2024) outperforms several pooling methods and MIL-based approaches, including CLAM (M. Y. Lu et al. 2021), with AUCs of 96.08 and 96.01 and F1-scores of 89.36 and 91.42 on the CAMELYON-16 and TCGA Lung datasets, respectively.

**SegAnyPath** (C. Wang et al. 2024) further extends SAM for pathology image segmentation, addressing the challenges of multi-resolution data, staining inconsistencies and diverse segmentation tasks. To address staining variability, it applies stain augmentation paired with self-distillation, enabling the model to learn robust, stain-invariant features. Additionally, it employs a task-guided Mixture of Experts (MoE) decoder, specializing in distinct segmentation tasks including tissue, tumor and cell segmentation. SegAnyPath incorporates self-supervised pretraining with a Masked Auto-Encoder (MAE) to learn generalizable features, which are further refined during supervised fine-tuning. The model was trained on a dataset of over 1.5 million images and 3.5 million segmentation masks from 39 public sources, spanning diverse organs, magnifications, and staining protocols. SegAnyPath achieved a 29.27% improvement in Dice scores compared to fine-tuned SAM on external datasets, demonstrating its adaptability to domain-specific challenges in pathology.

**Cell-ViT** (Hörst et al. 2024) and Cell-ViT-SAM (Hörst et al. 2024) are ViT-based models for automated cell nuclei segmentation, utilizing SAM and a pre-trained ViT encoder. It achieves superior performance in nuclei detection and segmentation and a faster inference on the PanNuke and MoNuSeg datasets compared to the existing state of the art methods.

**Xu et al.** (K. Xu, Goetz, and Rajpoot 2024) evaluated the domain generalizability of the SAM for nuclear instance segmentation, both with and without fine-tuning the mask decoder. SAM shows strong generalizability in zero-shot learning when provided with a ground truth bounding box prompt. In a clinically relevant task, the fine-tuned SAM using nuclear central points as prompts outperforms HoVer-Net on an external test dataset.

### 5.2.4. Findings and Critical Remarks

The adaptations of SAM for computational pathology, as demonstrated by SAM-Path (J. Zhang et al. 2023), SAM for HistoPath (Deng et al. 2023), and SAM-MIL (Fang et al. 2024), represent significant steps forward in segmentation and classification tasks. Each adaptation brings unique strengths, from SAM-Path's effectiveness in semantic segmentation to SAM-MIL's superior handling of spatial context. Collectively, these models contribute to advancing digital pathology.

However, a critical gap remains in the current research landscape. While the "segment anything" concept is foundational for bottom-up processing of WSIs, crucial for predictive and prognostic analytics—a comprehensive implementation of this idea as a modeling framework in computational pathology is still lacking. No model has been pre-trained exclusively on pathology data, nor has Meta's SAM been extensively fine-tuned on large histopathology datasets comprising millions of slides. This limitation, particularly in handling multiple predictive modeling tasks requiring multi-level aggregation, presents a crucial opportunity for future research. Developing pathology-specific foundation models could truly revolutionize the field.

# 6. Clinical-grade Evaluation of Pathology Foundation Models

To understand the challenges surrounding the clinical adoption of pathology foundation models, we examined three key applications where these models are expected to demonstrate clinical-grade performance. These case studies provide insights into both their current capabilities and areas for further development and evaluation.

## 6.1. Prostate Cancer Detection: Paige Prostate vs. Virchow Foundation Model

The Virchow (Vorontsov et al. 2024) foundation model was initially trained on 1,488,550 whole-slide images (WSIs) in an unsupervised manner and later fine-tuned using supervised learning on 35,387 slide groups, including 2,829 prostate specimens. By comparison, the FDA-approved Paige Prostate model (Campanella et al. 2019) was trained on 66,713 prostate blocks using a multiple-instance, weakly-supervised method with a ResNet34 (K. He et al. 2016) architecture. While the Virchow (Vorontsov et al. 2024) model show better generalization and achieved an impressive AUC of 0.980 on prostate cancer detection, it performed slightly lower than Paige Prostate's more specialized model, which attained an AUC of 0.995 ($P < 0.05$), as reported by authors in Virchow (Vorontsov et al. 2024).

Although the foundation model demonstrated comparable performance, its slightly lower AUC underscores a key challenge to be further investigated: can general-purpose models not always match the performance of task-specific models like Paige Prostate in clinical settings? This invites further exploration into how foundation models can be fine-tuned to close this performance gap while maintaining their generalist advantages. Nevertheless, the narrower performance gap in a single performance measure warrants extending comparative analysis with measures like sensitivity, specificity, and balanced accuracy. Additionally, an important challenge lies in assessing the higher development, training, and operational costs of generic foundation models—both in terms of financial impact on organizations and the environmental footprint due to the extensive computational resources required for their large-scale training.

## 6.2. Microsatellite Stability (MSS) Screening in Colorectal Cancer

Microsatellite instability (MSI) status prediction in colorectal cancer has seen variable performance across different models and cohorts. Studies report AUROC values for MSI status prediction on the TCGA-COADREAD dataset ranging from 0.88 to 0.91 (Amelie Echle et al. 2020; Saillard et al. 2021; Bilal et al. 2022; A. Echle et al. 2022; Bilal, Tsang, et al. 2023). External validation across multiple international cohorts (other than TCGA) shows improved AUROC scores of 0.96, 0.972, and 0.98, indicating that performance can vary based on dataset size and composition (Bilal, Tsang, et al. 2023; Bilal et al. 2022; Amelie Echle et al. 2020; A. Echle et al. 2022; Saillard et al. 2021).

MSIntuit CRC (Saillard et al. 2023), a commercial CE-IVD-marked diagnostic device, achieved sensitivities of 92-95%, ruling out approximately 40% of microsatellite-stable (MSS) patients from further testing. Trained on 859 WSIs from the TCGA-COADREAD cohort and validated on the PAIP cohort, MSIntuit CRC delivers AUROC values of 0.88 (sensitivity 0.98, specificity 0.46) on the MPATH-DP200 cohort and 0.87 (sensitivity 0.96, specificity 0.47) on the MPATH-UFS cohort.

Foundation models such as GigaPath (H. Xu et al. 2024), Phikon (Filiot et al. 2023), and Kaiko (ai et al. 2024) have shown varied results. AUROCs on the PAIP cohort range from 0.66 (Hibou-B (Nechaev, Pchelnikov, and Ivanova 2024)) to 0.978 (Phikon (Filiot et al. 2023)), reflecting both the potential and challenges of using foundation models for MSI status prediction. For MMR classification in colorectal cancer, Myles et al. (Myles et al. 2024) compared CTransPath (Xiyue Wang et al. 2022a), Phikon (Filiot et al. 2023), and UNI (R. J. Chen et al. 2024) against a pretrained ResNet50 (K. He et al. 2016). On test set, UNI (R. J. Chen et al. 2024), Phikon (Filiot et al. 2023), CTransPath (Xiyue Wang et al. 2022a) and ResNet50 (K. He et al. 2016) achieved AUROC of 0.7136, 0.7136, 0.6880, and 0.6709, respectively. OmniScreen (Y. K. Wang et al. 2024), leveraging the Virchow2 model – initially trained on 3,134,922 WSIs in an unsupervised manner – later finetuned further using supervised learning on 47,960 WSIs from 38,984 patients, achieved an AUROC of 0.98, highlighting the possibility of using large-scale, pre-trained models for high-throughput genomic analysis. Recently, Phikon-v2 (Filiot et al. 2024) achieved AUROC of 0.882 and 0.991 and GigaPath (H. Xu et al. 2024) achieved 0.888 and 0.980, on Cy1 and PAIP cohorts, respectively.

The variability in performance across these models points to a need for improved generalization and increasing evaluation metrics. Suboptimal MSI estimation methods and the inherent variability in clinical datasets may explain the performance discrepancies. As the gold standard for MSI testing still falls short of 100% accuracy (Luchini et al.

2019; Stjepanovic et al. 2019), a key question remains: can the next generation of foundation models overcome these persistent challenges?

## 6.3. Colon Biopsy Pre-Screening AI

Colon biopsy pre-screening aims to reduce the pathologist's workload by filtering out normal slides and identifying abnormal or inflammatory (non-neoplastic) and abnormal neoplastic cases with high sensitivity (0.99). Recent models like CAIMAN (Bilal, Tsang, et al. 2023) and IGUANA (Graham, Minhas, et al. 2023) have shown promise, achieving a specificity of 0.56 and 0.55, respectively, at 0.99 sensitivity in cross-validation experiments. These models have demonstrated the capability to automatically rule out more than 55% of normal colonic biopsies, reducing the burden on pathologists.

However, foundation models have yet to provide significant evaluations on this specific task. While models such as UNI (R. J. Chen et al. 2024), REMEDIS (Azizi et al. 2023), and CTransPath (Xiyue Wang et al. 2022a) have demonstrated performance on colorectal cancer screening with balanced accuracies of 65%, 52%, and 39%, respectively, their application to colon biopsy pre-screening remains underexplored. PathAlign (F. Ahmed et al. 2024), which uses a large language model to generate reports distinguishing benign polyps from pre-cancerous adenomas and highly suspicious cancers, has shown promise in workflow efficiency, though quantitative diagnostic evaluations are absent.

Selecting appropriate performance metrics is critical in evaluating the clinical utility of pre-screening AI. The ideal goal is 100% sensitivity in identifying abnormal cases without missing any neoplastic or non-neoplastic abnormalities. While current models approach the average pathologist-level sensitivity of 0·975 (Bilal, Tsang, et al. 2023; K. S. Wang et al. 2021), variability in human interpretation and clinical experience suggests room for improvement. A pressing question for foundation models and generative AI is whether they can further reduce this variability and enhance initial screening for abnormalities.

## 6.4. Findings and Critical Remarks

Pathology foundation models offer significant potential, but challenges remain in clinical adoption. While large models like Virchow (Vorontsov et al. 2024) exhibit competitive performance, specialized models (Campanella et al. 2019) such as Paige Prostate currently deliver superior results in specific tasks (Vorontsov et al. 2024). In MSI status prediction, foundation models show variability, reflecting issues with generalization across different cohorts and assays. In colon biopsy pre-screening, while AI models (Bilal, Tsang, et al. 2023; Graham, Minhas, et al. 2023) have shown promise in reducing workload, foundation models have yet to demonstrate their full potential.

These challenges emphasize the need for continued research, model refinement, and comprehensive benchmarking of foundation models to achieve clinical-grade performance across diverse medical tasks. As advancements in this field continue, a critical question persists: can these models overcome current limitations and unlock new possibilities for precision diagnostics, offering more accurate, efficient, and scalable solutions for clinical practice? In the next section, we explore research gaps and challenges other than clinical-grade performance in diagnostic decision-making that computational pathology faces in its widespread adoptions decides.

# 7. Foundation Pathology Models and Grand Challenges

The integration of AI, particularly foundation models, in digital pathology is reshaping how cancer is diagnosed and treated. We have witnessed a shift from pathologists examining tissue samples under a microscope to the advent of high-resolution WSIs, where AI algorithms can now analyze these images to identify patterns, predict outcomes, and assist in diagnoses (A. H. Song et al. 2023; Snead et al. 2016).

As described in the previous sections, the technical challenges lie in the complexity of WSIs — massive gigapixel images that exhibit significant variability due to differences in sample preparation and scanning. This variability complicates the development of models that can generalize across institutions, which is essential for scaling AI's impact. (Niazi, Parwani, and Gurcan 2019) In clinical practice, AI offers the potential for faster, more accurate, and personalized diagnostics. However, its integration into workflows is not straightforward, raising further challenges such as regulating ethical AI use, protecting patient privacy, and scaling the infrastructure required to support AI systems (Alowais et al. 2023).

The rapid advancement of foundation models discussed in the previous sections highlights the immense potential for modernizing pathology—but also reveals emerging challenges, particularly regarding model interpretability, generalization, and clinical adoption. In this section, we gather the current and future challenges in computational and digital pathology, providing a roadmap for AI researchers, clinicians, and decision-makers as they navigate this evolving landscape.

## 7.1. Data Acquisition and Quality

High-quality data is fundamental to the success of AI models in computational pathology, yet WSIs are often difficult to standardize across institutions. Variations in staining techniques and scanner settings add complexity, while collecting large datasets with expert annotations remains expensive and time-consuming. This lack of standardized, annotated data slows down the development of robust AI models.

Consensus among pathologists reporting is another crucial issue to ensure the quality and reliability of training data, as several assessments emphasize (Bilal, Tsang, et al. 2023; Bulten et al. 2022). Studies involving large language models (LLMs) (Du et al. 2024) underscore the importance of standardized data acquisition. (Sushil, Zack, et al. 2024) For instance, the CORAL (Sushil, Kennedy, et al. 2024) dataset introduced expert-labeled progress notes, improving oncological information extraction and highlighting the benefits of fine-tuning LLMs on high-quality, structured data. Collectively, these studies demonstrate the pivotal role of LLMs in standardizing diagnostic and prognostic reporting in pathology, including extracting structured condition codes reports (Bumgardner et al. 2024; Truhn et al. 2023), enhancing report standardization (Alzaid et al. 2024), and identifying pathological complete response (pCR) (Cheligeer et al. 2024).

Table 7 summarizes key challenges in computational and digital pathology, mapping them to AI solutions, state-of-the-art approaches, and future directions involving Foundation Models.

*Table 7. Grand Challenges in Computational Pathology and Potential of Foundation Models*

| Grand Challenge | Not Solved Effectively Yet | Future Exploration |
|---|---|---|
| **Data Acquisition and Quality** | - Lack of well-annotated, high-quality datasets<br>- Variability in data sources | - Exploration of semi-supervised and supervised counterparts<br>- Balancing annotations from human experts and AI-copilots and improving consensus among pathologists |
| **Data Privacy and Security** | - Handling private data across borders effectively remains unresolved | - Federated learning with Foundation Models trained on diverse data subsets |
| **Computational Infrastructure** | - Real-time processing of large WSIs in clinical settings<br>- Scalability in non-specialized hospitals | - Foundation Models optimized for low-resource settings |
| **Model Interpretability** | - Black-box models lack sufficient transparency for clinical use<br>- Difficulty of evaluating multimodal foundation models<br>- Lack of intrinsic evaluation metrics | - Foundation Models trained with interpretability constraints<br>- Exploring new metrics of evaluation<br>- Leveraging multimodalities to obtain more insights for explanation |
| **Generalization and Robustness** | - Robust performance across variable tissue staining, preparation, and imaging conditions | - Multimodal Foundation Models with cross-institutional and cross-domain learning |
| **Integration into Clinical Workflows** | - Resistance from pathologists<br>- Workflow disruption in transitioning to digital pathology | - Task-agnostic and task-specific foundation models to facilitate easier integration in clinical workflows |
| **Bias in AI Models** | - Comprehensive mitigation of biases in models and datasets remains unsolved | - Pretrained foundation models incorporating fairness objectives from the start |
| **Regulatory and Legal Frameworks** | - Adaptation of AI regulatory frameworks are lagging technological advancements | - Foundation models pre-validated for medical use |
| **Scalability of AI Solutions** | - Scaling foundation models for low-resource environments | - Need for cost-effective AI deployment at scale, particularly in developing regions |
| **Ethical and Societal Concerns** | - Foundation models with built-in ethical constraints | - Accountability and responsibility in autonomous AI systems not fully addressed |
| **Data Sovereignty & Global Collaboration** | - Distributed foundation models collaborating across institutions without centralized data storage | - International legal frameworks for cross-border data-sharing remain unresolved |
| **Adaptation to Novel Medical Discoveries** | - Integrating continuous learning from real-world clinical data remains a challenge | - Foundation models capable of continual fine-tuning based on the latest medical knowledge |
| **Personalized Diagnostics** | - Personalization at scale with limited data availability remains unsolved | - Foundation models fine-tuned for patient-specific diagnostics |
| **Hardware Limitations** | - Computationally efficient AI for pathology tasks is still an ongoing challenge | - Foundation Models optimized for energy-efficient hardware deployment |
| **Multi-modal Data Integration** | - Real-world implementation of multi-modal data integration in pathology still evolving | - Foundation models (e.g., CLIP, BioGPT) designed for multimodal data integration |

## 7.2. Data Privacy and Security

The need for vast amounts of clinical data raises concerns about patient privacy and data security. While regulations like HIPAA and GDPR exist to protect patient information, they make data-sharing across institutions challenging, creating barriers for collaborative AI development. Moreover, there is uncertainty around data ownership—whether it belongs to the patient, hospital, or laboratory (Asif et al. 2023; Mennella et al. 2024; Kiran et al. 2023).

This challenge extends to the development of foundation models, both literally and technically. The majority of foundation models are developed using proprietary data that is often limited to one region and population (M. Y. Lu, Chen, Williamson, Chen, Zhao, et al. 2024; Vorontsov et al. 2024; Zimmermann et al. 2024; Y. K. Wang et al. 2024; Xiyue Wang et al. 2024; H. Xu et al. 2024; Huang et al. 2023). Additionally, the representation of diseases in these models frequently lacks the diversity found in real-world scenarios, raising several concerns that have already been identified (Dippel et al. 2024; H. Xu et al. 2024; Arora et al. 2023; Nakagawa et al. 2023). This limitation can hinder the generalizability and effectiveness of AI applications in diverse clinical settings.

## 7.3. Computational Infrastructure

Processing gigapixel-sized WSIs requires significant computational resources, and not all institutions have access to the high-performance computing (HPC) required for AI analysis. While cloud computing offers some respite, real-time, low-latency analysis essential for clinical decision-making remains difficult to achieve (Kiran et al. 2023; Madabhushi 2009; Lujan, Li, and Parwani 2022).

## 7.4. Model Interpretability

AI models, particularly deep learning models, are often referred to as “black boxes” because they are difficult to interpret (Plass et al. 2023). In healthcare applications, where accountability and transparency are crucial, this lack of interpretability poses a significant barrier to clinical adoption (M. I. Ahmed et al. 2023; Sadeghi et al. 2024; Amann et al. 2022). Regulatory agencies like the FDA require AI models to be explainable, adding another layer of complexity to their deployment in pathology (McNamara, Yi, and Lotter 2024).

A recent study (Le et al. 2024) highlights the importance of addressing the interpretability challenge of foundation models. It employs mechanistic interpretability analysis using features from the embedding space of a ViT Small model from PLUTO (Juyal et al. 2024), identifying interpretable representations of cell and tissue morphology alongside gene expression. The study reveals a connection between morphological properties and gene expression, further supported by a separate spatial transcriptomics dataset. The difficulty of evaluating multimodal pathology models amplified because of the involvement of individual foundation models further necessitates interpretability. The rapid pace of foundation model development has not been matched by interoperability analysis studies for these models (Hart et al. 2023). Addressing these issues is essential for fostering trust and facilitating the clinical adoption of AI technologies in pathology.

## 7.5. Generalization and Robustness

AI models trained in specific environments (e.g., hospitals or laboratories) often struggle to generalize when applied in different settings due to variations in equipment, tissue preparation, or patient populations. This lack of robustness limits the effectiveness of AI across diverse clinical contexts. Gustafsson et al. (Gustafsson and Rantalainen 2024) investigated the robustness of foundation models in computational pathology, under different types of distribution shifts, including both domain shifts (differences in data across institutions) and label shifts (differences in grade distributions between datasets). The authors evaluated two foundation models, UNI and CONCH, for prostate cancer grading using the PANDA dataset, which includes images from two institutions. Results indicated that pathology-specific foundation models significantly outperformed general models (such as ResNet pretrained on ImageNet), even with limited training data.

While foundation models have shown promise in addressing these issues, there is a pressing need for standardized evaluation protocols and benchmarks that assess generalization and robustness (Hart et al. 2023). Methodologies developed for evaluating generative AI in LLMs can serve as a valuable framework; for instance, the GLUE (A. Wang et al. 2018) and SuperGLUE (A. Wang et al. 2020) benchmarks provide comprehensive measures of generalization capabilities across multiple tasks. The evaluation approach in "Language Models are Few-Shot Learners" (Brown et al. 2020) also demonstrates effective assessment methods in diverse tasks. Adapting these standardized evaluation frameworks for foundation models in computational pathology can enhance insights into their performance amidst real-world variability, ultimately improving their clinical applicability and trustworthiness.

## 7.6. Integration into Clinical Workflows

While AI has the potential to revolutionize pathology, integrating it into existing clinical workflows poses significant challenges. Many pathologists remain hesitant to trust AI for critical diagnostic decisions, leading to resistance against disrupting established practices. Furthermore, hospitals and laboratories must invest in new infrastructure, including digital scanners and software, to enable AI-powered pathology (Hart et al. 2023). The successful integration of LLMs has shown promising results, as seen in studies where local training and fine-tuning of models improved task performance, demonstrating that targeted applications of AI can enhance the workflow (Hassan, Kushniruk, and Borycki 2024). By fostering trust through transparency and accessibility, pathology Foundation Models can further encourage their adoption in clinical settings.

## 7.7. Ethical and Societal Concerns

As AI technologies evolve, concerns regarding algorithmic bias—potentially disadvantaging specific patient groups—become increasingly critical (Bilal, Jewsbury, et al. 2023; Agarwal et al. 2023; Bulten et al. 2022)? Additionally, as AI systems gain autonomy in diagnostic decision-making, questions of accountability arise: who is responsible when an AI system makes an error (Novelli, Taddeo, and Floridi 2024; Jackson et al. 2021)? Addressing these ethical and societal concerns is essential to build trust in AI technologies, ensuring their fair and responsible implementation in clinical practice. Foundation Models hold the potential to drive significant societal impact by democratizing access to advanced AI tools, fostering innovation, and improving outcomes in precision medicine and pathology research. By broadening the availability of these models and addressing concerns around bias and accountability, the healthcare industry can work towards more equitable healthcare solutions.

## 7.8. Scalability of AI Solutions

The scalability of AI-powered pathology solutions is constrained by hardware limitations and the energy consumption of AI models (Jia et al. 2023). Deploying AI at a scale across multiple institutions, particularly in resource-limited settings, presents significant challenges. A detailed evaluation of hardware costs and constraints, along with energy consumption, is necessary to understand their impact on research and development expenditures and adoption challenges in the context of foundation model development(Ardon et al. 2024; Hanna et al. 2022).

## 7.9. Grand Challenges and Potential of Foundation Models

AI has made significant progress in addressing challenges in computational pathology, particularly through predictive, contrastive, and generative applications. These techniques enhance data quality, privacy, and computational efficiency. However, many challenges remain unresolved, with large-scale, foundation models that can be adapted for specific tasks—showing the most promise.

Foundation models and AI copilots have the potential to make a significant societal impact, driving a paradigm shift in AI research and deployment. However, their potential remains underutilized due to limited accessibility. Broadening the availability of these tools could democratize their use, fostering innovation and improving outcomes in precision medicine and pathology research.

## 7.10. Findings and critical remarks

Foundation Models have redefined traditional tasks in computational pathology, while their advancements in novel applications promise to revolutionize clinical diagnostics and research, contributing to more equitable and efficient healthcare outcomes. AI has demonstrated significant potential in addressing many challenges in computational pathology, particularly through predictive, contrastive, and generative applications. However, significant obstacles remain, especially in areas like model generalization, privacy, and interpretability. Addressing these challenges is crucial for the transformative impact of AI on pathology practice globally.

# Acknowledgements

The authors extend their appreciation to Prince Sattam bin Abdulaziz University for funding this research work through the project number (PSAU/2024/01/29745).

## References

Agarwal, R., M. Bjarnadottir, L. Rhue, M. Dugas, K. Crowley, J. Clark, and G. Gao. 2023. 'Addressing Algorithmic Bias and the Perpetuation of Health Inequities: An AI Bias Aware Framework'. *Health Policy and Technology* 12 (1): 100702. https://doi.org/10.1016/j.hlpt.2022.100702.

Ahmed, Faruk, Andrew Sellergren, Lin Yang, Shawn Xu, Boris Babenko, Abbi Ward, Niels Olson, et al. 2024. 'PathAlign: A Vision-Language Model for Whole Slide Images in Histopathology'. arXiv. http://arxiv.org/abs/2406.19578.

Ahmed, Molla Imaduddin, Brendan Spooner, John Isherwood, Mark Lane, Emma Orrock, and Ashley Dennison. 2023. 'A Systematic Review of the Barriers to the Implementation of Artificial Intelligence in Healthcare'. *Cureus* 15 (10): e46454. https://doi.org/10.7759/cureus.46454.

ai, kaiko, Nanne Aben, Edwin D. de Jong, Ioannis Gatopoulos, Nicolas Känzig, Mikhail Karasikov, Axel Lagré, Roman Moser, Joost van Doorn, and Fei Tang. 2024. 'Towards Large-Scale Training of Pathology Foundation Models'. arXiv. http://arxiv.org/abs/2404.15217.

Alfasly, Saghir, Peyman Nejat, Sobhan Hemati, Jibran Khan, Isaiah Lahr, Areej Alsaafin, Abubakr Shafique, et al. 2023. 'When Is a Foundation Model a Foundation Model'. arXiv. http://arxiv.org/abs/2309.11510.

Alowais, Shuroug A., Sahar S. Alghamdi, Nada Alsuhebany, Tariq Alqahtani, Abdulrahman I. Alshaya, Sumaya N. Almohareb, Atheer Aldairem, et al. 2023. 'Revolutionizing Healthcare: The Role of Artificial Intelligence in Clinical Practice'. *BMC Medical Education* 23 (1): 689. https://doi.org/10.1186/s12909-023-04698-z.

Alzaid, Ethar, Gabriele Pergola, Harriet Evans, David Snead, and Fayyaz Minhas. 2024. 'Large Multimodal Model Based Standardisation of Pathology Reports with Confidence and Their Prognostic Significance'. arXiv. http://arxiv.org/abs/2405.02040.

Amann, Julia, Dennis Vetter, Stig Nikolaj Blomberg, Helle Collatz Christensen, Megan Coffee, Sara Gerke, Thomas K. Gilbert, et al. 2022. 'To Explain or Not to Explain?—Artificial Intelligence Explainability in Clinical Decision Support Systems'. Edited by Henry Horng-Shing Lu. *PLOS Digital Health* 1 (2): e0000016. https://doi.org/10.1371/journal.pdig.0000016.

Ardon, Orly, Sylvia L. Asa, Mark C. Lloyd, Giovanni Lujan, Anil Parwani, Juan C. Santa-Rosario, Bryan Van Meter, et al. 2024. 'Understanding the Financial Aspects of Digital Pathology: A Dynamic Customizable Return on Investment Calculator for Informed Decision-Making'. *Journal of Pathology Informatics* 15 (December):100376. https://doi.org/10.1016/j.jpi.2024.100376.

Arora, Anmol, Joseph E. Alderman, Joanne Palmer, Shaswath Ganapathi, Elinor Laws, Melissa D. McCradden, Lauren Oakden-Rayner, et al. 2023. 'The Value of Standards for Health Datasets in Artificial Intelligence-Based Applications'. *Nature Medicine* 29 (11): 2929–38. https://doi.org/10.1038/s41591-023-02608-w.

Asif, Amina, Kashif Rajpoot, Simon Graham, David Snead, Fayyaz Minhas, and Nasir Rajpoot. 2023. 'Unleashing the Potential of AI for Pathology: Challenges and Recommendations'. *The Journal of Pathology* 260 (5): 564–77. https://doi.org/10.1002/path.6168.

Atallah, Nehal M., Noorul Wahab, Michael S. Toss, Shorouk Makhlouf, Asmaa Y. Ibrahim, Ayat G. Lashen, Suzan Ghannam, et al. 2023. 'Deciphering the Morphology of Tumor-Stromal Features in Invasive Breast Cancer Using Artificial Intelligence'. *Modern Pathology* 36 (10): 100254. https://doi.org/10.1016/j.modpat.2023.100254.

Azizi, Shekoofeh, Laura Culp, Jan Freyberg, Basil Mustafa, Sebastien Baur, Simon Kornblith, Ting Chen, et al. 2023. 'Robust and Data-Efficient Generalization of Self-Supervised Machine Learning for Diagnostic Imaging'. *Nature Biomedical Engineering* 7 (6): 756–79. https://doi.org/10.1038/s41551-023-01049-7.

Balachandran, Vidhisha, Jingya Chen, Neel Joshi, Besmira Nushi, Hamid Palangi, Eduardo Salinas, Vibhav Vineet, James Woffinden-Luey, and Safoora Yousefi. 2024. 'Eureka: Evaluating and Understanding Large Foundation Models'. arXiv. http://arxiv.org/abs/2409.10566.

Bengio, Yoshua, Aaron Courville, and Pascal Vincent. 2014. 'Representation Learning: A Review and New Perspectives'. arXiv. http://arxiv.org/abs/1206.5538.

Beyer, Lucas, Pavel Izmailov, Alexander Kolesnikov, Mathilde Caron, Simon Kornblith, Xiaohua Zhai, Matthias Minderer, Michael Tschannen, Ibrahim Alabdulmohsin, and Filip Pavetic. 2023. 'FlexiViT: One Model for All Patch Sizes'. In *2023 IEEE/CVF Conference on Computer Vision and Pattern Recognition (CVPR)*, 14496–506. Vancouver, BC, Canada: IEEE. https://doi.org/10.1109/CVPR52729.2023.01393.

Bilal, Mohsin, Robert Jewsbury, Ruoyu Wang, Hammam M. AlGhamdi, Amina Asif, Mark Eastwood, and Nasir Rajpoot. 2023. 'An Aggregation of Aggregation Methods in Computational Pathology'. *Medical Image Analysis* 88 (August):102885. https://doi.org/10.1016/j.media.2023.102885.

Bilal, Mohsin, Mohammed Nimir, David Snead, Graham S. Taylor, and Nasir Rajpoot. 2022. 'Role of AI and Digital Pathology for Colorectal Immuno-Oncology'. *British Journal of Cancer*, October. https://doi.org/10.1038/s41416-022-01986-1.

Bilal, Mohsin, Shan E Ahmed Raza, Ayesha Azam, Simon Graham, Mohammad Ilyas, Ian A Cree, David Snead, Fayyaz Minhas, and Nasir M Rajpoot. 2021. 'Development and Validation of a Weakly Supervised Deep Learning Framework to Predict the Status of Molecular Pathways and Key Mutations in Colorectal Cancer from Routine Histology Images: A Retrospective Study'. *The Lancet Digital Health* 3 (12): e763–72. https://doi.org/10.1016/S2589-7500(21)00180-1.

Bilal, Mohsin, Yee Wah Tsang, Mahmoud Ali, Simon Graham, Emily Hero, Noorul Wahab, Katherine Dodd, et al. 2023. 'Development and Validation of Artificial Intelligence-Based Prescreening of Large-Bowel Biopsies Taken in the UK and Portugal: A Retrospective Cohort Study'. *The Lancet Digital Health* 5 (11): e786–97. https://doi.org/10.1016/S2589-7500(23)00148-6.

Bommasani, Rishi, Drew A. Hudson, Ehsan Adeli, Russ Altman, Simran Arora, Sydney von Arx, Michael S. Bernstein, et al. 2022. 'On the Opportunities and Risks of Foundation Models'. arXiv. http://arxiv.org/abs/2108.07258.

Breen, Jack, Katie Allen, Kieran Zucker, Lucy Godson, Nicolas M. Orsi, and Nishant Ravikumar. 2024. 'A Comprehensive Evaluation of Histopathology Foundation Models for Ovarian Cancer Subtype Classification'. arXiv. https://doi.org/10.48550/arXiv.2405.09990.

Brown, Tom B., Benjamin Mann, Nick Ryder, Melanie Subbiah, Jared Kaplan, Prafulla Dhariwal, Arvind Neelakantan, et al. 2020. 'Language Models Are Few-Shot Learners'. arXiv. http://arxiv.org/abs/2005.14165.

Bulten, Wouter, Kimmo Kartasalo, Po-Hsuan Cameron Chen, Peter Ström, Hans Pinckaers, Kunal Nagpal, Yuannan Cai, et al. 2022. 'Artificial Intelligence for Diagnosis and Gleason Grading of Prostate Cancer: The PANDA Challenge'. *Nature Medicine*, January. https://doi.org/10.1038/s41591-021-01620-2.

Bumgardner, V. K. Cody, Aaron Mullen, Samuel E. Armstrong, Caylin Hickey, Victor Marek, and Jeff Talbert. 2024. 'Local Large Language Models for Complex Structured Tasks'. *AMIA Joint Summits on Translational Science Proceedings. AMIA Joint Summits on Translational Science* 2024:105–14.

Campanella, Gabriele, Shengjia Chen, Ruchika Verma, Jennifer Zeng, Aryeh Stock, Matt Croken, Brandon Veremis, et al. 2024. 'A Clinical Benchmark of Public Self-Supervised Pathology Foundation Models'. arXiv. http://arxiv.org/abs/2407.06508.

Campanella, Gabriele, Eugene Fluder, Jennifer Zeng, Chad Vanderbilt, and Thomas J. Fuchs. 2024. 'Beyond Multiple Instance Learning: Full Resolution All-In-Memory End-To-End Pathology Slide Modeling'. arXiv. http://arxiv.org/abs/2403.04865.

Campanella, Gabriele, Matthew G. Hanna, Luke Geneslaw, Allen P. Miraflor, Vitor Werneck Krauss Silva, Klaus J. Busam, Edi Brogi, Victor E. Reuter, David S. Klimstra, and Thomas J. Fuchs. 2019. 'Clinical-Grade Computational Pathology Using Weakly Supervised Deep Learning on Whole Slide Images'. *Nature Medicine*, 1–9.

Campanella, Gabriele, Ricky Kwan, Eugene Fluder, Jennifer Zeng, Aryeh Stock, Brandon Veremis, Alexandros D. Polydorides, et al. 2023. 'Computational Pathology at Health System Scale -- Self-Supervised Foundation Models from Three Billion Images'. arXiv. http://arxiv.org/abs/2310.07033.

Caron, Mathilde, Hugo Touvron, Ishan Misra, Hervé Jégou, Julien Mairal, Piotr Bojanowski, and Armand Joulin. 2021. 'Emerging Properties in Self-Supervised Vision Transformers'. *arXiv:2104.14294 [Cs]*, April. http://arxiv.org/abs/2104.14294.

Chanda, Dibaloke, Milan Aryal, Nasim Yahya Soltani, and Masoud Ganji. 2024. 'A New Era in Computational Pathology: A Survey on Foundation and Vision-Language Models'. arXiv. https://doi.org/10.48550/arXiv.2408.14496.

Cheligeer, Cheligeer, Guosong Wu, Alison Laws, May Lynn Quan, Andrea Li, Anne-Marie Brisson, Jason Xie, and Yuan Xu. 2024. 'Validating Large Language Models for Identifying Pathologic Complete Responses After Neoadjuvant Chemotherapy for Breast Cancer Using a Population-Based Pathologic Report Data'. https://doi.org/10.21203/rs.3.rs-4004164/v1.

Chen, Hui, Guiguang Ding, Xudong Liu, Zijia Lin, Ji Liu, and Jungong Han. 2020. 'IMRAM: Iterative Matching With Recurrent Attention Memory for Cross-Modal Image-Text Retrieval'. In *2020 IEEE/CVF Conference on Computer Vision and Pattern*

*Recognition (CVPR)*, 12652–60. Seattle, WA, USA: IEEE. https://doi.org/10.1109/CVPR42600.2020.01267.
Chen, Pingyi, Chenglu Zhu, Sunyi Zheng, Honglin Li, and Lin Yang. 2024. 'WSI-VQA: Interpreting Whole Slide Images by Generative Visual Question Answering'. arXiv. https://doi.org/10.48550/arXiv.2407.05603.
Chen, Richard J, Chengkuan Chen, Yicong Li, Tiffany Y Chen, Andrew D Trister, Rahul G Krishnan, and Faisal Mahmood. 2022. 'Scaling Vision Transformers to Gigapixel Images via Hierarchical Self-Supervised Learning'. In *Proceedings of the IEEE/CVF Conference on Computer Vision and Pattern Recognition (CVPR)*, 16144–55.
Chen, Richard J., Tong Ding, Ming Y. Lu, Drew F. K. Williamson, Guillaume Jaume, Andrew H. Song, Bowen Chen, et al. 2024. 'Towards a General-Purpose Foundation Model for Computational Pathology'. *Nature Medicine* 30 (3): 850–62. https://doi.org/10.1038/s41591-024-02857-3.
Chen, Shengjia, Gabriele Campanella, Abdulkadir Elmas, Aryeh Stock, Jennifer Zeng, Alexandros D. Polydorides, Adam J. Schoenfeld, et al. 2024. 'Benchmarking Embedding Aggregation Methods in Computational Pathology: A Clinical Data Perspective'. arXiv. http://arxiv.org/abs/2407.07841.
Chen, Ting, Simon Kornblith, Mohammad Norouzi, and Geoffrey Hinton. 2020. 'A Simple Framework for Contrastive Learning of Visual Representations'. In *Proceedings of the 37th International Conference on Machine Learning*, 1597–1607. PMLR. https://proceedings.mlr.press/v119/chen20j.html.
Chen, Ying, Guoan Wang, Yuanfeng Ji, Yanjun Li, Jin Ye, Tianbin Li, Bin Zhang, et al. 2024. 'SlideChat: A Large Vision-Language Assistant for Whole-Slide Pathology Image Understanding'. arXiv. https://doi.org/10.48550/arXiv.2410.11761.
Chen, Zheyi, Liuchang Xu, Hongting Zheng, Luyao Chen, Amr Tolba, Liang Zhao, Keping Yu, and Hailin Feng. 2024. 'Evolution and Prospects of Foundation Models: From Large Language Models to Large Multimodal Models'. *Computers, Materials & Continua* 80 (2): 1753–1808. https://doi.org/10.32604/cmc.2024.052618.
Chiang, Wei-Lin, Zhuohan Li, Zi Lin, Ying Sheng, Zhanghao Wu, Hao Zhang, Lianmin Zheng, et al. 2023. 'Vicuna: An Open-Source Chatbot Impressing GPT-4 with 90%* ChatGPT Quality'. https://lmsys.org/blog/2023-03-30-vicuna/.
Dankers, Verna, Elia Bruni, and Dieuwke Hupkes. 2022. 'The Paradox of the Compositionality of Natural Language: A Neural Machine Translation Case Study'. arXiv. http://arxiv.org/abs/2108.05885.
Dehghani, Mostafa, Yi Tay, Alexey A. Gritsenko, Zhe Zhao, Neil Houlsby, Fernando Diaz, Donald Metzler, and Oriol Vinyals. 2021. 'The Benchmark Lottery'. arXiv. http://arxiv.org/abs/2107.07002.
Deng, Ruining, Can Cui, Quan Liu, Tianyuan Yao, Lucas W. Remedios, Shunxing Bao, Bennett A. Landman, et al. 2023. 'Segment Anything Model (SAM) for Digital Pathology: Assess Zero-Shot Segmentation on Whole Slide Imaging'. arXiv. http://arxiv.org/abs/2304.04155.
Devlin, Jacob, Ming-Wei Chang, Kenton Lee, and Kristina Toutanova. 2019. 'BERT: Pre-Training of Deep Bidirectional Transformers for Language Understanding'. arXiv. http://arxiv.org/abs/1810.04805.

Ding, Jiayu, Shuming Ma, Li Dong, Xingxing Zhang, Shaohan Huang, Wenhui Wang, Nanning Zheng, and Furu Wei. 2023. 'LongNet: Scaling Transformers to 1,000,000,000 Tokens'. arXiv. https://doi.org/10.48550/arXiv.2307.02486.

Ding, Ruiwen, James Hall, Neil Tenenholtz, and Kristen Severson. 2023. 'Improving Mitosis Detection on Histopathology Images Using Large Vision-Language Models'. arXiv. http://arxiv.org/abs/2310.07176.

Ding, Tong, Sophia J. Wagner, Andrew H. Song, Richard J. Chen, Ming Y. Lu, Andrew Zhang, Anurag J. Vaidya, et al. 2024. 'Multimodal Whole Slide Foundation Model for Pathology'. arXiv. https://doi.org/10.48550/arXiv.2411.19666.

Dippel, Jonas, Barbara Feulner, Tobias Winterhoff, Timo Milbich, Stephan Tietz, Simon Schallenberg, Gabriel Dernbach, et al. 2024. 'RudolfV: A Foundation Model by Pathologists for Pathologists'. arXiv. http://arxiv.org/abs/2401.04079.

Dosovitskiy, Alexey, Lucas Beyer, Alexander Kolesnikov, Dirk Weissenborn, Xiaohua Zhai, Thomas Unterthiner, Mostafa Dehghani, et al. 2020. 'An Image Is Worth 16x16 Words: Transformers for Image Recognition at Scale'. In . https://openreview.net/forum?id=YicbFdNTTy.

Du, Wei, Jaryse Carol Harris, Alessandro Brunetti, Olivia Leung, Xingchen Li, Selemon Walle, Qing Yu, et al. 2024. 'Large Language Models in Pathology: A Comparative Study on Multiple Choice Question Performance with Pathology Trainees'. https://doi.org/10.1101/2024.07.10.24310093.

Echle, A., N. Ghaffari Laleh, P. Quirke, H.I. Grabsch, H.S. Muti, O.L. Saldanha, S.F. Brockmoeller, et al. 2022. 'Artificial Intelligence for Detection of Microsatellite Instability in Colorectal Cancer—a Multicentric Analysis of a Pre-Screening Tool for Clinical Application'. *ESMO Open* 7 (2): 100400. https://doi.org/10.1016/j.esmoop.2022.100400.

Echle, Amelie, Heike Irmgard Grabsch, Philip Quirke, Piet A. van den Brandt, Nicholas P. West, Gordon G.A. Hutchins, Lara R. Heij, et al. 2020. 'Clinical-Grade Detection of Microsatellite Instability in Colorectal Tumors by Deep Learning'. *Gastroenterology* 159 (4): 1406-1416.e11. https://doi.org/10.1053/j.gastro.2020.06.021.

Fang, Heng, Sheng Huang, Wenhao Tang, Luwen Huangfu, and Bo Liu. 2024. 'SAM-MIL: A Spatial Contextual Aware Multiple Instance Learning Approach for Whole Slide Image Classification'. arXiv. http://arxiv.org/abs/2407.17689.

Ferber, Dyke, Omar S. M. El Nahhas, Georg Wölflein, Isabella C. Wiest, Jan Clusmann, Marie-Elisabeth Leßman, Sebastian Foersch, et al. 2024. 'Autonomous Artificial Intelligence Agents for Clinical Decision Making in Oncology'. arXiv. https://doi.org/10.48550/arXiv.2404.04667.

Ferber, Dyke, Georg Wölflein, Isabella C. Wiest, Marta Ligero, Srividhya Sainath, Narmin Ghaffari Laleh, Omar S. M. El Nahhas, et al. 2024. 'In-Context Learning Enables Multimodal Large Language Models to Classify Cancer Pathology Images'. arXiv. http://arxiv.org/abs/2403.07407.

Filiot, Alexandre, Ridouane Ghermi, Antoine Olivier, Paul Jacob, Lucas Fidon, Alice Mac Kain, Charlie Saillard, and Jean-Baptiste Schiratti. 2023. 'Scaling Self-Supervised Learning for Histopathology with Masked Image Modeling'. https://doi.org/10.1101/2023.07.21.23292757.

Filiot, Alexandre, Paul Jacob, Alice Mac Kain, and Charlie Saillard. 2024. 'Phikon-v2, A Large and Public Feature Extractor for Biomarker Prediction'. arXiv. http://arxiv.org/abs/2409.09173.

Firoozi, Roya, Johnathan Tucker, Stephen Tian, Anirudha Majumdar, Jiankai Sun, Weiyu Liu, Yuke Zhu, et al. 2023. 'Foundation Models in Robotics: Applications, Challenges, and the Future'. arXiv. http://arxiv.org/abs/2312.07843.

Gallegos, Isabel O., Ryan A. Rossi, Joe Barrow, Md Mehrab Tanjim, Sungchul Kim, Franck Dernoncourt, Tong Yu, Ruiyi Zhang, and Nesreen K. Ahmed. 2024. 'Bias and Fairness in Large Language Models: A Survey'. arXiv. http://arxiv.org/abs/2309.00770.

Gamper, Jevgenij, and Nasir Rajpoot. 2021. 'Multiple Instance Captioning: Learning Representations from Histopathology Textbooks and Articles'. arXiv. http://arxiv.org/abs/2103.05121.

Garrido, Quentin, Randall Balestriero, Laurent Najman, and Yann Lecun. 2023. 'RankMe: Assessing the Downstream Performance of Pretrained Self-Supervised Representations by Their Rank'. arXiv. http://arxiv.org/abs/2210.02885.

Gildenblat, Jacob, and Ofir Hadar. 2024. 'Segmentation by Factorization: Unsupervised Semantic Segmentation for Pathology by Factorizing Foundation Model Features'. arXiv. http://arxiv.org/abs/2409.05697.

Gou, Jiaxiang, Luping Ji, Pei Liu, and Mao Ye. 2024. 'Queryable Prototype Multiple Instance Learning with Vision-Language Models for Incremental Whole Slide Image Classification'. arXiv. https://doi.org/10.48550/ARXIV.2410.10573.

Graham, Simon, Fayyaz Minhas, Mohsin Bilal, Mahmoud Ali, Yee Wah Tsang, Mark Eastwood, Noorul Wahab, et al. 2023. 'Screening of Normal Endoscopic Large Bowel Biopsies with Interpretable Graph Learning: A Retrospective Study'. *Gut* 72 (9): 1709–21. https://doi.org/10.1136/gutjnl-2023-329512.

Graham, Simon, Quoc Dang Vu, Mostafa Jahanifar, Shan E Ahmed Raza, Fayyaz Minhas, David Snead, and Nasir Rajpoot. 2023. 'One Model Is All You Need: Multi-Task Learning Enables Simultaneous Histology Image Segmentation and Classification'. *Medical Image Analysis* 83 (January):102685. https://doi.org/10.1016/j.media.2022.102685.

Guo, Zhengrui, Jiabo Ma, Yingxue Xu, Yihui Wang, Liansheng Wang, and Hao Chen. 2024. 'HistGen: Histopathology Report Generation via Local-Global Feature Encoding and Cross-Modal Context Interaction'. arXiv. http://arxiv.org/abs/2403.05396.

Guo, Zishan, Renren Jin, Chuang Liu, Yufei Huang, Dan Shi, Supryadi, Linhao Yu, et al. 2023. 'Evaluating Large Language Models: A Comprehensive Survey'. arXiv. http://arxiv.org/abs/2310.19736.

Gustafsson, Fredrik K., and Mattias Rantalainen. 2024. 'Evaluating Computational Pathology Foundation Models for Prostate Cancer Grading under Distribution Shifts'. arXiv. https://doi.org/10.48550/ARXIV.2410.06723.

Hanin, Boris, and David Rolnick. 2019. 'Complexity of Linear Regions in Deep Networks'. arXiv. http://arxiv.org/abs/1901.09021.

Hanna, Matthew G., Orly Ardon, Victor E. Reuter, Sahussapont Joseph Sirintrapun, Christine England, David S. Klimstra, and Meera R. Hameed. 2022. 'Integrating Digital Pathology into Clinical Practice'. *Modern Pathology* 35 (2): 152–64. https://doi.org/10.1038/s41379-021-00929-0.

Hart, Steven N., Noah G. Hoffman, Peter Gershkovich, Chancey Christenson, David S. McClintock, Lauren J. Miller, Ronald Jackups, Vahid Azimi, Nicholas Spies, and Victor Brodsky. 2023. 'Organizational Preparedness for the Use of Large Language Models in Pathology Informatics'. *Journal of Pathology Informatics* 14:100338. https://doi.org/10.1016/j.jpi.2023.100338.

Hassan, Masooma, Andre Kushniruk, and Elizabeth Borycki. 2024. 'Barriers to and Facilitators of Artificial Intelligence Adoption in Health Care: Scoping Review'. *JMIR Human Factors* 11 (August):e48633. https://doi.org/10.2196/48633.

He, Kaiming, Xinlei Chen, Saining Xie, Yanghao Li, Piotr Dollár, and Ross Girshick. 2021. 'Masked Autoencoders Are Scalable Vision Learners'. arXiv. http://arxiv.org/abs/2111.06377.

He, Kaiming, Xinlei Chen, Saining Xie, Yanghao Li, Piotr Dollar, and Ross Girshick. 2022. 'Masked Autoencoders Are Scalable Vision Learners'. In *2022 IEEE/CVF Conference on Computer Vision and Pattern Recognition (CVPR)*, 15979–88. New Orleans, LA, USA: IEEE. https://doi.org/10.1109/CVPR52688.2022.01553.

He, Kaiming, Xiangyu Zhang, Shaoqing Ren, and Jian Sun. 2016. 'Deep Residual Learning for Image Recognition'. In *The IEEE Conference on Computer Vision and Pattern Recognition (CVPR)*.

He, Xuehai, Zhuo Cai, Wenlan Wei, Yichen Zhang, Luntian Mou, Eric Xing, and Pengtao Xie. 2021. 'Towards Visual Question Answering on Pathology Images'. In *Proceedings of the 59th Annual Meeting of the Association for Computational Linguistics and the 11th International Joint Conference on Natural Language Processing (Volume 2: Short Papers)*, 708–18. Online: Association for Computational Linguistics. https://doi.org/10.18653/v1/2021.acl-short.90.

Hoang, Danh-Tai, Gal Dinstag, Eldad D. Shulman, Leandro C. Hermida, Doreen S. Ben-Zvi, Efrat Elis, Katherine Caley, et al. 2024. 'A Deep-Learning Framework to Predict Cancer Treatment Response from Histopathology Images through Imputed Transcriptomics'. *Nature Cancer*, July. https://doi.org/10.1038/s43018-024-00793-2.

Hörst, Fabian, Moritz Rempe, Lukas Heine, Constantin Seibold, Julius Keyl, Giulia Baldini, Selma Ugurel, et al. 2024. 'CellViT: Vision Transformers for Precise Cell Segmentation and Classification'. *Medical Image Analysis* 94 (May):103143. https://doi.org/10.1016/j.media.2024.103143.

Hu, Dingyi, Zhiguo Jiang, Jun Shi, Fengying Xie, Kun Wu, Kunming Tang, Ming Cao, Jianguo Huai, and Yushan Zheng. 2024. 'Histopathology Language-Image Representation Learning for Fine-Grained Digital Pathology Cross-Modal Retrieval'. *Medical Image Analysis* 95 (July):103163. https://doi.org/10.1016/j.media.2024.103163.

Hua, Shengyi, Fang Yan, Tianle Shen, Lei Ma, and Xiaofan Zhang. 2024. 'PathoDuet: Foundation Models for Pathological Slide Analysis of H&E and IHC Stains'. *Medical Image Analysis* 97 (October):103289. https://doi.org/10.1016/j.media.2024.103289.

Huang, Zhi, Federico Bianchi, Mert Yuksekgonul, Thomas J. Montine, and James Zou. 2023. 'A Visual–Language Foundation Model for Pathology Image Analysis Using Medical Twitter'. *Nature Medicine* 29 (9): 2307–16. https://doi.org/10.1038/s41591-023-02504-3.

Ilharco, Gabriel, Mitchell Wortsman, Ross Wightman, Cade Gordon, Nicholas Carlini, Rohan Taori, Achal Dave, et al. 2021. ‘OpenCLIP’. Zenodo. https://doi.org/10.5281/zenodo.5143773.

Ilse, Maximilian, Jakub M. Tomczak, and Max Welling. 2018. ‘Attention-Based Deep Multiple Instance Learning’. https://doi.org/10.48550/ARXIV.1802.04712.

Jackson, Brian R., Ye Ye, James M. Crawford, Michael J. Becich, Somak Roy, Jeffrey R. Botkin, Monica E. de Baca, and Liron Pantanowitz. 2021. ‘The Ethics of Artificial Intelligence in Pathology and Laboratory Medicine: Principles and Practice’. *Academic Pathology* 8:2374289521990784. https://doi.org/10.1177/2374289521990784.

Jaegle, Andrew, Felix Gimeno, Andrew Brock, Andrew Zisserman, Oriol Vinyals, and Joao Carreira. 2021. ‘Perceiver: General Perception with Iterative Attention’. *arXiv:2103.03206 [Cs, Eess]*, March. http://arxiv.org/abs/2103.03206.

Jaume, Guillaume, Lukas Oldenburg, Anurag Vaidya, Richard J. Chen, Drew F. K. Williamson, Thomas Peeters, Andrew H. Song, and Faisal Mahmood. 2024. ‘Transcriptomics-Guided Slide Representation Learning in Computational Pathology’. arXiv. http://arxiv.org/abs/2405.11618.

Jaume, Guillaume, Anurag Vaidya, Andrew Zhang, Andrew H. Song, Richard J. Chen, Sharifa Sahai, Dandan Mo, Emilio Madrigal, Long Phi Le, and Faisal Mahmood. 2024. ‘Multistain Pretraining for Slide Representation Learning in Pathology’. arXiv. http://arxiv.org/abs/2408.02859.

Javed, Sajid, Arif Mahmood, Muhammad Moazam Fraz, Navid Alemi Koohbanani, Ksenija Benes, Yee-Wah Tsang, Katherine Hewitt, David Epstein, David Snead, and Nasir Rajpoot. 2020. ‘Cellular Community Detection for Tissue Phenotyping in Colorectal Cancer Histology Images’. *Medical Image Analysis* 63 (July):101696. https://doi.org/10.1016/j.media.2020.101696.

Javed, Sajid, Arif Mahmood, Iyyakutti Iyappan Ganapathi, Fayaz Ali Dharejo, Naoufel Werghi, and Mohammed Bennamoun. 2024. ‘CPLIP: Zero-Shot Learning for Histopathology with Comprehensive Vision-Language Alignment’. In *Proceedings of the IEEE/CVF Conference on Computer Vision and Pattern Recognition (CVPR)*, 11450–59. IEEE. https://openaccess.thecvf.com/content/CVPR2024/papers/Javed_CPLIP_Zero-Shot_Learning_for_Histopathology_with_Comprehensive_Vision-Language_Alignment_CVPR_2024_paper.pdf.

Jia, Zhenge, Jianxu Chen, Xiaowei Xu, John Kheir, Jingtong Hu, Han Xiao, Sui Peng, Xiaobo Sharon Hu, Danny Chen, and Yiyu Shi. 2023. ‘The Importance of Resource Awareness in Artificial Intelligence for Healthcare’. *Nature Machine Intelligence* 5 (7): 687–98. https://doi.org/10.1038/s42256-023-00670-0.

Jing, Longlong, and Yingli Tian. 2021. ‘Self-Supervised Visual Feature Learning With Deep Neural Networks: A Survey’. *IEEE Transactions on Pattern Analysis and Machine Intelligence* 43 (11): 4037–58. https://doi.org/10.1109/TPAMI.2020.2992393.

Juyal, Dinkar, Harshith Padigela, Chintan Shah, Daniel Shenker, Natalia Harguindeguy, Yi Liu, Blake Martin, et al. 2024. ‘PLUTO: Pathology-Universal Transformer’. arXiv. http://arxiv.org/abs/2405.07905.

Kang, Mingu, Heon Song, Seonwook Park, Donggeun Yoo, and Sérgio Pereira. 2023. ‘Benchmarking Self-Supervised Learning on Diverse Pathology Datasets’. In *2023*

*IEEE/CVF Conference on Computer Vision and Pattern Recognition (CVPR)*, 3344–54. Vancouver, BC, Canada: IEEE. https://doi.org/10.1109/CVPR52729.2023.00326.

Kather, Jakob Nikolas, Lara R. Heij, Heike I. Grabsch, Chiara Loeffler, Amelie Echle, Hannah Sophie Muti, Jeremias Krause, et al. 2020. 'Pan-Cancer Image-Based Detection of Clinically Actionable Genetic Alterations'. *Nature Cancer* 1 (August):789–99. https://doi.org/10.1101/833756.

Kather, Jakob Nikolas, Alexander T. Pearson, Niels Halama, Dirk Jäger, Jeremias Krause, Sven H. Loosen, Alexander Marx, et al. 2019. 'Deep Learning Can Predict Microsatellite Instability Directly from Histology in Gastrointestinal Cancer'. *Nature Medicine* 25 (7): 1054–56. https://doi.org/10.1038/s41591-019-0462-y.

Kather, Jakob Nikolas, Cleo-Aron Weis, Francesco Bianconi, Susanne M. Melchers, Lothar R. Schad, Timo Gaiser, Alexander Marx, and Frank Gerrit Zöllner. 2016. 'Multi-Class Texture Analysis in Colorectal Cancer Histology'. *Scientific Reports* 6 (1): 27988. https://doi.org/10.1038/srep27988.

Khan, Asifullah, Anabia Sohail, Mustansar Fiaz, Mehdi Hassan, Tariq Habib Afridi, Sibghat Ullah Marwat, Farzeen Munir, et al. 2024. 'A Survey of the Self Supervised Learning Mechanisms for Vision Transformers'. arXiv. http://arxiv.org/abs/2408.17059.

Kiran, Nfn, Fnu Sapna, Fnu Kiran, Deepak Kumar, Fnu Raja, Sheena Shiwlani, Antonella Paladini, et al. 2023. 'Digital Pathology: Transforming Diagnosis in the Digital Age'. *Cureus* 15 (9): e44620. https://doi.org/10.7759/cureus.44620.

Kirillov, Alexander, Eric Mintun, Nikhila Ravi, Hanzi Mao, Chloe Rolland, Laura Gustafson, Tete Xiao, et al. 2023. 'Segment Anything'. arXiv. http://arxiv.org/abs/2304.02643.

Kolesnikov, Alexander, Lucas Beyer, Xiaohua Zhai, Joan Puigcerver, Jessica Yung, Sylvain Gelly, and Neil Houlsby. 2020. 'Big Transfer (BiT): General Visual Representation Learning'. In *Computer Vision – ECCV 2020*, edited by Andrea Vedaldi, Horst Bischof, Thomas Brox, and Jan-Michael Frahm, 12350:491–507. Lecture Notes in Computer Science. Cham: Springer International Publishing. https://doi.org/10.1007/978-3-030-58558-7_29.

Lai, Zhengfeng, Zhuoheng Li, Luca Cerny Oliveira, Joohi Chauhan, Brittany N. Dugger, and Chen-Nee Chuah. 2023. 'CLIPath: Fine-Tune CLIP with Visual Feature Fusion for Pathology Image Analysis Towards Minimizing Data Collection Efforts'. In *2023 IEEE/CVF International Conference on Computer Vision Workshops (ICCVW)*, 2366–72. Paris, France: IEEE. https://doi.org/10.1109/ICCVW60793.2023.00251.

Lazard, Tristan, Marvin Lerousseau, Sophie Gardrat, Anne Vincent-Salomon, Marc-Henri Stern, Manuel Rodrigues, Etienne Decencière, and Thomas Walter. 2023. 'Democratizing Computational Pathology: Optimized Whole Slide Image Representations for The Cancer Genome Atlas'. https://doi.org/10.1101/2023.12.04.569894.

Le, Nhat, Ciyue Shen, Chintan Shah, Blake Martin, Daniel Shenker, Harshith Padigela, Jennifer Hipp, et al. 2024. 'Interpretability Analysis on a Pathology Foundation Model Reveals Biologically Relevant Embeddings across Modalities'. arXiv. http://arxiv.org/abs/2407.10785.

Lee, Jeaung, Jeewoo Lim, Keunho Byeon, and Jin Tae Kwak. 2024. ‘Benchmarking Pathology Foundation Models: Adaptation Strategies and Scenarios’. arXiv. https://doi.org/10.48550/arXiv.2410.16038.

Lee, Jinhyuk, Wonjin Yoon, Sungdong Kim, Donghyeon Kim, Sunkyu Kim, Chan Ho So, and Jaewoo Kang. 2020. ‘BioBERT: A Pre-Trained Biomedical Language Representation Model for Biomedical Text Mining’. Edited by Jonathan Wren. *Bioinformatics* 36 (4): 1234–40. https://doi.org/10.1093/bioinformatics/btz682.

Lenz, Tim, Peter Neidlinger, Marta Ligero, Georg Wölflein, Marko van Treeck, and Jakob Nikolas Kather. 2024. ‘Unsupervised Foundation Model-Agnostic Slide-Level Representation Learning’. arXiv. https://doi.org/10.48550/arXiv.2411.13623.

Li, Hao, Ying Chen, Yifei Chen, Wenxian Yang, Bowen Ding, Yuchen Han, Liansheng Wang, and Rongshan Yu. 2024. ‘Generalizable Whole Slide Image Classification with Fine-Grained Visual-Semantic Interaction’. arXiv. http://arxiv.org/abs/2402.19326.

Li, Junnan, Dongxu Li, Silvio Savarese, and Steven Hoi. 2023. ‘BLIP-2: Bootstrapping Language-Image Pre-Training with Frozen Image Encoders and Large Language Models’. In *Proceedings of the 40th International Conference on Machine Learning*, 19730–42. PMLR. https://proceedings.mlr.press/v202/li23q.html.

Li, Wenwen, Chia-Yu Hsu, Sizhe Wang, Yezhou Yang, Hyunho Lee, Anna Liljedahl, Chandi Witharana, et al. 2024. ‘Segment Anything Model Can Not Segment Anything: Assessing AI Foundation Model’s Generalizability in Permafrost Mapping’. arXiv. http://arxiv.org/abs/2401.08787.

Lin, Yi, Zhongchen Zhao, Zhengjie ZHU, Lisheng Wang, Kwang-Ting Cheng, and Hao Chen. 2023. ‘Exploring Visual Prompts for Whole Slide Image Classification with Multiple Instance Learning’. arXiv. http://arxiv.org/abs/2303.13122.

Liu, Haotian, Chunyuan Li, Qingyang Wu, and Yong Jae Lee. 2023. ‘Visual Instruction Tuning’. *Advances in Neural Information Processing Systems* 36 (December):34892–916.

Liu, Jiawei, Cheng Yang, Zhiyuan Lu, Junze Chen, Yibo Li, Mengmei Zhang, Ting Bai, et al. 2024. ‘Towards Graph Foundation Models: A Survey and Beyond’. arXiv. http://arxiv.org/abs/2310.11829.

Liu, Pei, Luping Ji, Jiaxiang Gou, Bo Fu, and Mao Ye. 2024. ‘Interpretable Vision-Language Survival Analysis with Ordinal Inductive Bias for Computational Pathology’. arXiv. http://arxiv.org/abs/2409.09369.

Liu, Ze, Yutong Lin, Yue Cao, Han Hu, Yixuan Wei, Zheng Zhang, Stephen Lin, and Baining Guo. 2021. ‘Swin Transformer: Hierarchical Vision Transformer Using Shifted Windows’. In *2021 IEEE/CVF International Conference on Computer Vision (ICCV)*, 9992–10002. Montreal, QC, Canada: IEEE. https://doi.org/10.1109/ICCV48922.2021.00986.

Liu, Zhuang, Hanzi Mao, Chao-Yuan Wu, Christoph Feichtenhofer, Trevor Darrell, and Saining Xie. 2022. ‘A ConvNet for the 2020s’. In , 11976–86. https://openaccess.thecvf.com/content/CVPR2022/html/Liu_A_ConvNet_for_the_2020s_CVPR_2022_paper.html.

Longpre, Shayne, Stella Biderman, Alon Albalak, Hailey Schoelkopf, Daniel McDuff, Sayash Kapoor, Kevin Klyman, et al. 2024. ‘The Responsible Foundation Model Development Cheatsheet: A Review of Tools & Resources’. arXiv. http://arxiv.org/abs/2406.16746.

Lu, Cheng, Hongming Xu, Mingkang Wang, Duanbo Shi, Hua-Min Qin, Anant Madabhushi, Peng Gao, and Fengyu Cong. 2024. 'When Multiple Instance Learning Meets Foundation Models: Advancing Histological Whole Slide Image Analysis'. https://doi.org/10.21203/rs.3.rs-4704418/v1.

Lu, Jiasen, Dhruv Batra, Devi Parikh, and Stefan Lee. 2019. 'ViLBERT: Pretraining Task-Agnostic Visiolinguistic Representations for Vision-and-Language Tasks'. arXiv. http://arxiv.org/abs/1908.02265.

Lu, Ming Y., Bowen Chen, Drew F. K. Williamson, Richard J. Chen, Ivy Liang, Tong Ding, Guillaume Jaume, et al. 2024. 'A Visual-Language Foundation Model for Computational Pathology'. *Nature Medicine* 30 (3): 863–74. https://doi.org/10.1038/s41591-024-02856-4.

Lu, Ming Y., Bowen Chen, Drew F. K. Williamson, Richard J. Chen, Melissa Zhao, Aaron K. Chow, Kenji Ikemura, et al. 2024. 'A Multimodal Generative AI Copilot for Human Pathology'. *Nature*, June. https://doi.org/10.1038/s41586-024-07618-3.

Lu, Ming Y., Bowen Chen, Andrew Zhang, Drew F. K. Williamson, Richard J. Chen, Tong Ding, Long Phi Le, Yung-Sung Chuang, and Faisal Mahmood. 2023. 'Visual Language Pretrained Multiple Instance Zero-Shot Transfer for Histopathology Images'. arXiv. http://arxiv.org/abs/2306.07831.

Lu, Ming Y., Drew F. K. Williamson, Tiffany Y. Chen, Richard J. Chen, Matteo Barbieri, and Faisal Mahmood. 2021. 'Data-Efficient and Weakly Supervised Computational Pathology on Whole-Slide Images'. *Nature Biomedical Engineering* 5 (6): 555–70. https://doi.org/10.1038/s41551-020-00682-w.

Lu, Wenqi, Ayat G Lashen, Noorul Wahab, Islam M Miligy, Mostafa Jahanifar, Michael Toss, Simon Graham, et al. 2023. 'AI-based Intra-tumor Heterogeneity Score of Ki67 Expression as a Prognostic Marker for Early-stage ER+/HER2− Breast Cancer'. *The Journal of Pathology: Clinical Research*, October, cjp2.346. https://doi.org/10.1002/cjp2.346.

Luchini, C., F. Bibeau, M.J.L. Ligtenberg, N. Singh, A. Nottegar, T. Bosse, R. Miller, et al. 2019. 'ESMO Recommendations on Microsatellite Instability Testing for Immunotherapy in Cancer, and Its Relationship with PD-1/PD-L1 Expression and Tumour Mutational Burden: A Systematic Review-Based Approach'. *Annals of Oncology* 30 (8): 1232–43. https://doi.org/10.1093/annonc/mdz116.

Lujan, Giovani, Zaibo Li, and Anil V. Parwani. 2022. 'Challenges in Implementing a Digital Pathology Workflow in Surgical Pathology'. *Human Pathology Reports* 29 (September):300673. https://doi.org/10.1016/j.hpr.2022.300673.

Luo, Renqian, Liai Sun, Yingce Xia, Tao Qin, Sheng Zhang, Hoifung Poon, and Tie-Yan Liu. 2022. 'BioGPT: Generative Pre-Trained Transformer for Biomedical Text Generation and Mining'. *Briefings in Bioinformatics* 23 (6): bbac409. https://doi.org/10.1093/bib/bbac409.

Lv, Xiaomin, Chong Lai, Liya Ding, Maode Lai, and Qingrong Sun. 2024. 'Renal Digital Pathology Visual Knowledge Search Platform Based on Language Large Model and Book Knowledge'. arXiv. http://arxiv.org/abs/2406.18556.

Ma, Jiabo, Zhengrui Guo, Fengtao Zhou, Yihui Wang, Yingxue Xu, Yu Cai, Zhengjie Zhu, et al. 2024. 'Towards A Generalizable Pathology Foundation Model via Unified Knowledge Distillation'. arXiv. http://arxiv.org/abs/2407.18449.

Madabhushi, Anant. 2009. 'Digital Pathology Image Analysis: Opportunities and Challenges'. *Imaging in Medicine* 1 (1): 7–10. https://doi.org/10.2217/iim.09.9.

Maleki, Danial, Shahryar Rahnamayan, and H. R. Tizhoosh. 2024. ‘A Self-Supervised Framework for Cross-Modal Search in Histopathology Archives Using Scale Harmonization’. *Scientific Reports* 14 (1): 9724. https://doi.org/10.1038/s41598-024-60256-7.

Maleki, Danial, and H. R. Tizhoosh. 2022. ‘LILE: Look In-Depth before Looking Elsewhere -- A Dual Attention Network Using Transformers for Cross-Modal Information Retrieval in Histopathology Archives’. arXiv. http://arxiv.org/abs/2203.01445.

Mallya, Mayur, Ali Khajegili Mirabadi, Hossein Farahani, and Ali Bashashati. 2024. ‘Benchmarking Histopathology Foundation Models for Ovarian Cancer Bevacizumab Treatment Response Prediction from Whole Slide Images’. arXiv. http://arxiv.org/abs/2407.20596.

Marini, Niccolò, Stefano Marchesin, Marek Wodzinski, Alessandro Caputo, Damian Podareanu, Bryan Cardenas Guevara, Svetla Boytcheva, et al. 2024. ‘Multimodal Representations of Biomedical Knowledge from Limited Training Whole Slide Images and Reports Using Deep Learning’. *Medical Image Analysis* 97 (October):103303. https://doi.org/10.1016/j.media.2024.103303.

Mayall, Frederick George, Mark David Goodhead, Louis De Mendonça, Sarah Eleanor Brownlie, Azka Anees, and Stephen Perring. 2023. ‘Artificial Intelligence-Based Triage of Large Bowel Biopsies Can Improve Workflow’. *Journal of Pathology Informatics* 14:100181. https://doi.org/10.1016/j.jpi.2022.100181.

McNamara, Stephanie L., Paul H. Yi, and William Lotter. 2024. ‘The Clinician-AI Interface: Intended Use and Explainability in FDA-Cleared AI Devices for Medical Image Interpretation’. *Npj Digital Medicine* 7 (1): 80. https://doi.org/10.1038/s41746-024-01080-1.

Mennella, Ciro, Umberto Maniscalco, Giuseppe De Pietro, and Massimo Esposito. 2024. ‘Ethical and Regulatory Challenges of AI Technologies in Healthcare: A Narrative Review’. *Heliyon* 10 (4): e26297. https://doi.org/10.1016/j.heliyon.2024.e26297.

Meseguer, Pablo, Rocío del Amor, and Valery Naranjo. 2024. ‘MI-VisionShot: Few-Shot Adaptation of Vision-Language Models for Slide-Level Classification of Histopathological Images’. arXiv. https://doi.org/10.48550/ARXIV.2410.15881.

Mieko, Ochi, Daisuke Komura, and Shumpei Ishikawa. 2025. ‘Pathology Foundation Models’. *JMA Journal* 8 (1): 121–30. https://doi.org/10.31662/jmaj.2024-0206.

Montúfar, Guido, Razvan Pascanu, Kyunghyun Cho, and Yoshua Bengio. 2014. ‘On the Number of Linear Regions of Deep Neural Networks’. arXiv. http://arxiv.org/abs/1402.1869.

Myles, Craig, In Hwa Um, David J. Harrison, and David Harris-Birtill. 2024. ‘Leveraging Foundation Models for Enhanced Detection of Colorectal Cancer Biomarkers in Small Datasets’. In *Medical Image Understanding and Analysis*, edited by Moi Hoon Yap, Connah Kendrick, Ardhendu Behera, Timothy Cootes, and Reyer Zwiggelaar, 14859:329–43. Lecture Notes in Computer Science. Cham: Springer Nature Switzerland. https://doi.org/10.1007/978-3-031-66955-2_23.

Nakagawa, Keisuke, Lama Moukheiber, Leo A. Celi, Malhar Patel, Faisal Mahmood, Dibson Gondim, Michael Hogarth, and Richard Levenson. 2023. ‘AI in Pathology: What Could Possibly Go Wrong?’ *Seminars in Diagnostic Pathology* 40 (2): 100–108. https://doi.org/10.1053/j.semdp.2023.02.006.

Nechaev, Dmitry, Alexey Pchelnikov, and Ekaterina Ivanova. 2024. 'Hibou: A Family of Foundational Vision Transformers for Pathology'. arXiv. http://arxiv.org/abs/2406.05074.

Neidlinger, Peter, Omar S. M. El Nahhas, Hannah Sophie Muti, Tim Lenz, Michael Hoffmeister, Hermann Brenner, Marko van Treeck, et al. 2024. 'Benchmarking Foundation Models as Feature Extractors for Weakly-Supervised Computational Pathology'. arXiv. http://arxiv.org/abs/2408.15823.

Nguyen, Anh Tien, Keunho Byeon, Kyungeun Kim, Boram Song, Seoung Wan Chae, and Jin Tae Kwak. 2024. 'CAMP: Continuous and Adaptive Learning Model in Pathology'. arXiv. http://arxiv.org/abs/2407.09030.

Nguyen, Anh Tien, and Jin Tae Kwak. 2023. 'GPC: Generative and General Pathology Image Classifier'. In *Medical Image Computing and Computer Assisted Intervention – MICCAI 2023 Workshops*, edited by M. Emre Celebi, Md Sirajus Salekin, Hyunwoo Kim, Shadi Albarqouni, Catarina Barata, Allan Halpern, Philipp Tschandl, et al., 14393:203–12. Lecture Notes in Computer Science. Cham: Springer Nature Switzerland. https://doi.org/10.1007/978-3-031-47401-9_20.

Nguyen, Anh Tien, Trinh Thi Le Vuong, and Jin Tae Kwak. 2024. 'Towards a Text-Based Quantitative and Explainable Histopathology Image Analysis'. arXiv. http://arxiv.org/abs/2407.07360.

Niazi, Muhammad Khalid Khan, Anil V Parwani, and Metin N Gurcan. 2019. 'Digital Pathology and Artificial Intelligence'. *The Lancet Oncology* 20 (5): e253–61. https://doi.org/10.1016/S1470-2045(19)30154-8.

Nicke, Till, Jan Raphael Schaefer, Henning Hoefener, Friedrich Feuerhake, Dorit Merhof, Fabian Kiessling, and Johannes Lotz. 2024. 'Tissue Concepts: Supervised Foundation Models in Computational Pathology'. arXiv. http://arxiv.org/abs/2409.03519.

Novelli, Claudio, Mariarosaria Taddeo, and Luciano Floridi. 2024. 'Accountability in Artificial Intelligence: What It Is and How It Works'. *AI & SOCIETY* 39 (4): 1871–82. https://doi.org/10.1007/s00146-023-01635-y.

OpenAI. 2023a. 'GPT-4V(Ision) System Card'. https://cdn.openai.com/papers/GPTV_System_Card.pdf.

———. 2023b. 'GPT-4V(Ision) Technical Work and Authors'. https://openai.com/contributions/gpt-4v/.

Oquab, Maxime, Timothée Darcet, Théo Moutakanni, Huy Vo, Marc Szafraniec, Vasil Khalidov, Pierre Fernandez, et al. 2024. 'DINOv2: Learning Robust Visual Features without Supervision'. arXiv. https://doi.org/10.48550/arXiv.2304.07193.

Peng, Zhiliang, Li Dong, Hangbo Bao, Qixiang Ye, and Furu Wei. 2022. 'BEiT v2: Masked Image Modeling with Vector-Quantized Visual Tokenizers'. arXiv. https://doi.org/10.48550/arXiv.2208.06366.

Perez-Lopez, Raquel, Narmin Ghaffari Laleh, Faisal Mahmood, and Jakob Nikolas Kather. 2024. 'A Guide to Artificial Intelligence for Cancer Researchers'. *Nature Reviews Cancer* 24 (6): 427–41. https://doi.org/10.1038/s41568-024-00694-7.

Plass, Markus, Michaela Kargl, Tim-Rasmus Kiehl, Peter Regitnig, Christian Geißler, Theodore Evans, Norman Zerbe, Rita Carvalho, Andreas Holzinger, and Heimo Müller. 2023. 'Explainability and Causability in Digital Pathology'. *The Journal of Pathology: Clinical Research* 9 (4): 251–60. https://doi.org/10.1002/cjp2.322.

Qiu, Jingna, Marc Aubreville, Frauke Wilm, Mathias Öttl, Jonas Utz, Maja Schlereth, and Katharina Breininger. 2024. ‘Leveraging Image Captions for Selective Whole Slide Image Annotation’. arXiv. http://arxiv.org/abs/2407.06363.

Qu, Linhao, Xiaoyuan Luo, Kexue Fu, Manning Wang, and Zhijian Song. 2024. ‘The Rise of AI Language Pathologists: Exploring Two-Level Prompt Learning for Few-Shot Weakly-Supervised Whole Slide Image Classification’. arXiv. http://arxiv.org/abs/2305.17891.

Qu, Linhao, Dingkang Yang, Dan Huang, Qinhao Guo, Rongkui Luo, Shaoting Zhang, and Xiaosong Wang. 2024. ‘Pathology-Knowledge Enhanced Multi-Instance Prompt Learning for Few-Shot Whole Slide Image Classification’. arXiv. http://arxiv.org/abs/2407.10814.

Raciti, Patricia, Jillian Sue, Juan A. Retamero, Rodrigo Ceballos, Ran Godrich, Jeremy D. Kunz, Adam Casson, et al. 2023. ‘Clinical Validation of Artificial Intelligence–Augmented Pathology Diagnosis Demonstrates Significant Gains in Diagnostic Accuracy in Prostate Cancer Detection’. *Archives of Pathology & Laboratory Medicine* 147 (10): 1178–85. https://doi.org/10.5858/arpa.2022-0066-oa.

Radford, Alec, Jong Wook Kim, Chris Hallacy, Aditya Ramesh, Gabriel Goh, Sandhini Agarwal, Girish Sastry, et al. 2021. ‘Learning Transferable Visual Models From Natural Language Supervision’. arXiv. http://arxiv.org/abs/2103.00020.

Raghu, Maithra, Ben Poole, Jon Kleinberg, Surya Ganguli, and Jascha Sohl-Dickstein. 2017. ‘On the Expressive Power of Deep Neural Networks’. arXiv. http://arxiv.org/abs/1606.05336.

Ray, Partha Pratim. 2023. ‘Benchmarking, Ethical Alignment, and Evaluation Framework for Conversational AI: Advancing Responsible Development of ChatGPT’. *BenchCouncil Transactions on Benchmarks, Standards and Evaluations* 3 (3): 100136. https://doi.org/10.1016/j.tbench.2023.100136.

Sadeghi, Zahra, Roohallah Alizadehsani, Mehmet Akif Cifci, Samina Kausar, Rizwan Rehman, Priyakshi Mahanta, Pranjal Kumar Bora, et al. 2024. ‘A Review of Explainable Artificial Intelligence in Healthcare’. *Computers and Electrical Engineering* 118 (August):109370. https://doi.org/10.1016/j.compeleceng.2024.109370.

Saillard, Charlie, Olivier Dehaene, Tanguy Marchand, Olivier Moindrot, Aurélie Kamoun, Benoit Schmauch, and Simon Jegou. 2021. ‘Self Supervised Learning Improves dMMR/MSI Detection from Histology Slides across Multiple Cancers’. *arXiv:2109.05819 [Cs, Eess]*, September. http://arxiv.org/abs/2109.05819.

Saillard, Charlie, Rémy Dubois, Oussama Tchita, Nicolas Loiseau, Thierry Garcia, Aurélie Adriansen, Séverine Carpentier, et al. 2023. ‘Validation of MSIntuit as an AI-Based Pre-Screening Tool for MSI Detection from Colorectal Cancer Histology Slides’. *Nature Communications* 14 (1): 6695. https://doi.org/10.1038/s41467-023-42453-6.

Saillard, Charlie, Rodolphe Jenatton, Felipe Llinares-López, Zelda Mariet, David Cahané, Eric Durand, and Jean-Philippe Vert. 2024. ‘H-Optimus-0’. https://github.com/bioptimus/releases/tree/main/models/h-optimus/v0.

Saldanha, Oliver Lester, Chiara M. L. Loeffler, Jan Moritz Niehues, Marko Van Treeck, Tobias P. Seraphin, Katherine Jane Hewitt, Didem Cifci, et al. 2023. ‘Self-Supervised Attention-Based Deep Learning for Pan-Cancer Mutation Prediction from Histopathology'. *Npj Precision Oncology* 7 (1): 35. https://doi.org/10.1038/s41698-023-00365-0.

Shaikovski, George, Adam Casson, Kristen Severson, Eric Zimmermann, Yi Kan Wang, Jeremy D. Kunz, Juan A. Retamero, et al. 2024. 'PRISM: A Multi-Modal Generative Foundation Model for Slide-Level Histopathology'. arXiv. http://arxiv.org/abs/2405.10254.

Shao, Zhuchen, Hao Bian, Yang Chen, Yifeng Wang, Jian Zhang, Xiangyang Ji, and yongbing zhang. 2021. 'TransMIL: Transformer Based Correlated Multiple Instance Learning for Whole Slide Image Classification'. In *Advances in Neural Information Processing Systems*, 34:2136–47. Curran Associates, Inc. https://proceedings.neurips.cc/paper/2021/hash/10c272d06794d3e5785d5e7c5356e9ff-Abstract.html.

Shen, Yiqing, Yulin Luo, Dinggang Shen, and Jing Ke. 2022. 'RandStainNA: Learning Stain-Agnostic Features from Histology Slides by Bridging Stain Augmentation and Normalization'. In *Medical Image Computing and Computer Assisted Intervention – MICCAI 2022*, edited by Linwei Wang, Qi Dou, P. Thomas Fletcher, Stefanie Speidel, and Shuo Li, 212–21. Cham: Springer Nature Switzerland. https://doi.org/10.1007/978-3-031-16434-7_21.

Shi, Congzhen, Ryan Rezai, Jiaxi Yang, Qi Dou, and Xiaoxiao Li. 2024. 'A Survey on Trustworthiness in Foundation Models for Medical Image Analysis'. arXiv. http://arxiv.org/abs/2407.15851.

Shi, Jiangbo, Chen Li, Tieliang Gong, Yefeng Zheng, and Huazhu Fu. 2024. 'ViLa-MIL: Dual-Scale Vision-Language Multiple Instance Learning for Whole Slide Image Classification'. In *Proceedings of the IEEE/CVF Conference on Computer Vision and Pattern Recognition (CVPR)*, 11248–58.

Snead, David RJ, Yee-Wah Tsang, Aisha Meskiri, Peter K Kimani, Richard Crossman, Nasir M Rajpoot, Elaine Blessing, et al. 2016. 'Validation of Digital Pathology Imaging for Primary Histopathological Diagnosis'. *Histopathology* 68 (7): 1063–72.

Song, Andrew H., Guillaume Jaume, Drew F. K. Williamson, Ming Y. Lu, Anurag Vaidya, Tiffany R. Miller, and Faisal Mahmood. 2023. 'Artificial Intelligence for Digital and Computational Pathology'. *Nature Reviews Bioengineering* 1 (12): 930–49. https://doi.org/10.1038/s44222-023-00096-8.

Song, Yale, and Mohammad Soleymani. 2019. 'Polysemous Visual-Semantic Embedding for Cross-Modal Retrieval'. In *2019 IEEE/CVF Conference on Computer Vision and Pattern Recognition (CVPR)*, 1979–88. Long Beach, CA, USA: IEEE. https://doi.org/10.1109/CVPR.2019.00208.

Steyaert, Sandra, Marija Pizurica, Divya Nagaraj, Priya Khandelwal, Tina Hernandez-Boussard, Andrew J. Gentles, and Olivier Gevaert. 2023. 'Multimodal Data Fusion for Cancer Biomarker Discovery with Deep Learning'. *Nature Machine Intelligence* 5 (4): 351–62. https://doi.org/10.1038/s42256-023-00633-5.

Stjepanovic, N., L. Moreira, F. Carneiro, F. Balaguer, A. Cervantes, J. Balmaña, and E. Martinelli. 2019. 'Hereditary Gastrointestinal Cancers: ESMO Clinical Practice Guidelines for Diagnosis, Treatment and Follow-Up†'. *Annals of Oncology* 30 (10): 1558–71. https://doi.org/10.1093/annonc/mdz233.

Sun, Yuxuan, Hao Wu, Chenglu Zhu, Sunyi Zheng, Qizi Chen, Kai Zhang, Yunlong Zhang, et al. 2024. 'PathMMU: A Massive Multimodal Expert-Level Benchmark for Understanding and Reasoning in Pathology'. arXiv. http://arxiv.org/abs/2401.16355.

Sun, Yuxuan, Yunlong Zhang, Yixuan Si, Chenglu Zhu, Zhongyi Shui, Kai Zhang, Jingxiong Li, Xingheng Lyu, Tao Lin, and Lin Yang. 2024. 'PathGen-1.6M: 1.6 Million Pathology Image-Text Pairs Generation through Multi-Agent Collaboration'. arXiv. http://arxiv.org/abs/2407.00203.

Sun, Yuxuan, Chenglu Zhu, Sunyi Zheng, Kai Zhang, Lin Sun, Zhongyi Shui, Yunlong Zhang, Honglin Li, and Lin Yang. 2024. 'PathAsst: A Generative Foundation AI Assistant towards Artificial General Intelligence of Pathology'. *Proceedings of the AAAI Conference on Artificial Intelligence* 38 (5): 5034–42. https://doi.org/10.1609/aaai.v38i5.28308.

Sushil, Madhumita, Vanessa E. Kennedy, Divneet Mandair, Brenda Y. Miao, Travis Zack, and Atul J. Butte. 2024. 'CORAL: Expert-Curated Oncology Reports to Advance Language Model Inference'. *NEJM AI* 1 (4). https://doi.org/10.1056/AIdbp2300110.

Sushil, Madhumita, Travis Zack, Divneet Mandair, Zhiwei Zheng, Ahmed Wali, Yan-Ning Yu, Yuwei Quan, Dmytro Lituiev, and Atul J Butte. 2024. 'A Comparative Study of Large Language Model-Based Zero-Shot Inference and Task-Specific Supervised Classification of Breast Cancer Pathology Reports'. *Journal of the American Medical Informatics Association*, June, ocae146. https://doi.org/10.1093/jamia/ocae146.

Tamkin, Alex, Vincent Liu, Rongfei Lu, Daniel Fein, Colin Schultz, and Noah Goodman. 2023. 'DABS: A Domain-Agnostic Benchmark for Self-Supervised Learning'. arXiv. http://arxiv.org/abs/2111.12062.

Tamkin, Alex, Mike Wu, and Noah Goodman. 2021. 'Viewmaker Networks: Learning Views for Unsupervised Representation Learning'. arXiv. http://arxiv.org/abs/2010.07432.

Touvron, Hugo, Louis Martin, Kevin Stone, Peter Albert, Amjad Almahairi, Yasmine Babaei, Nikolay Bashlykov, et al. 2023. 'Llama 2: Open Foundation and Fine-Tuned Chat Models'. arXiv. https://doi.org/10.48550/arXiv.2307.09288.

Tran, Manuel, Paul Schmidle, Sophia J. Wagner, Valentin Koch, Brenna Novotny, Valerio Lupperger, Annette Feuchtinger, et al. 2024. 'Generating Clinical-Grade Pathology Reports from Gigapixel Whole Slide Images with HistoGPT'. https://doi.org/10.1101/2024.03.15.24304211.

Truhn, Daniel, Chiara Ml Loeffler, Gustav Müller-Franzes, Sven Nebelung, Katherine J Hewitt, Sebastian Brandner, Keno K Bressem, Sebastian Foersch, and Jakob Nikolas Kather. 2023. 'Extracting Structured Information from Unstructured Histopathology Reports Using Generative Pre-trained Transformer 4 ( GPT -4)'. *The Journal of Pathology*, December, path.6232. https://doi.org/10.1002/path.6232.

Vaidya, Anurag, Andrew Zhang, Guillaume Jaume, Andrew H. Song, Tong Ding, Sophia J. Wagner, Ming Y. Lu, et al. 2025. 'Molecular-Driven Foundation Model for Oncologic Pathology'. arXiv. https://doi.org/10.48550/arXiv.2501.16652.

Vaswani, Ashish, Noam Shazeer, Niki Parmar, Jakob Uszkoreit, Llion Jones, Aidan N. Gomez, Lukasz Kaiser, and Illia Polosukhin. 2023. 'Attention Is All You Need'. arXiv. http://arxiv.org/abs/1706.03762.

Verwimp, Eli, Rahaf Aljundi, Shai Ben-David, Matthias Bethge, Andrea Cossu, Alexander Gepperth, Tyler L. Hayes, et al. 2024. 'Continual Learning: Applications and the Road Forward'. arXiv. http://arxiv.org/abs/2311.11908.
Vorontsov, Eugene, Alican Bozkurt, Adam Casson, George Shaikovski, Michal Zelechowski, Kristen Severson, Eric Zimmermann, et al. 2024. 'A Foundation Model for Clinical-Grade Computational Pathology and Rare Cancers Detection'. *Nature Medicine*, July. https://doi.org/10.1038/s41591-024-03141-0.
Wahab, Noorul, Michael Toss, Islam M. Miligy, Mostafa Jahanifar, Nehal M. Atallah, Wenqi Lu, Simon Graham, et al. 2023. 'AI-Enabled Routine H&E Image Based Prognostic Marker for Early-Stage Luminal Breast Cancer'. *Npj Precision Oncology* 7 (1): 122. https://doi.org/10.1038/s41698-023-00472-y.
Wang, Alex, Yada Pruksachatkun, Nikita Nangia, Amanpreet Singh, Julian Michael, Felix Hill, Omer Levy, and Samuel R. Bowman. 2020. 'SuperGLUE: A Stickier Benchmark for General-Purpose Language Understanding Systems'. arXiv. http://arxiv.org/abs/1905.00537.
Wang, Alex, Amanpreet Singh, Julian Michael, Felix Hill, Omer Levy, and Samuel Bowman. 2018. 'GLUE: A Multi-Task Benchmark and Analysis Platform for Natural Language Understanding'. In *Proceedings of the 2018 EMNLP Workshop BlackboxNLP: Analyzing and Interpreting Neural Networks for NLP*, 353–55. Brussels, Belgium: Association for Computational Linguistics. https://doi.org/10.18653/v1/W18-5446.
Wang, Chong, Yajie Wan, Shuxin Li, Kaili Qu, Xuezhi Zhou, Junjun He, Jing Ke, Yi Yu, Tianyun Wang, and Yiqing Shen. 2024. 'SegAnyPath: A Foundation Model for Multi-Resolution Stain-Variant and Multi-Task Pathology Image Segmentation'. *IEEE Transactions on Medical Imaging*, 1–1. https://doi.org/10.1109/TMI.2024.3501352.
Wang, K. S., G. Yu, C. Xu, X. H. Meng, J. Zhou, C. Zheng, Z. Deng, et al. 2021. 'Accurate Diagnosis of Colorectal Cancer Based on Histopathology Images Using Artificial Intelligence'. *BMC Medicine* 19 (1): 76. https://doi.org/10.1186/s12916-021-01942-5.
Wang, Xiaosong, Xiaofan Zhang, Guotai Wang, Junjun He, Zhongyu Li, Wentao Zhu, Yi Guo, et al. 2024. 'OpenMEDLab: An Open-Source Platform for Multi-Modality Foundation Models in Medicine'. arXiv. http://arxiv.org/abs/2402.18028.
Wang, Xiyue, Sen Yang, Jun Zhang, Minghui Wang, Jing Zhang, Wei Yang, Junzhou Huang, and Xiao Han. 2022a. 'Transformer-Based Unsupervised Contrastive Learning for Histopathological Image Classification'. *Medical Image Analysis* 81 (October):102559. https://doi.org/10.1016/j.media.2022.102559.
———. 2022b. 'Transformer-Based Unsupervised Contrastive Learning for Histopathological Image Classification'. *Medical Image Analysis* 81 (October):102559. https://doi.org/10.1016/j.media.2022.102559.
Wang, Xiyue, Junhan Zhao, Eliana Marostica, Wei Yuan, Jietian Jin, Jiayu Zhang, Ruijiang Li, et al. 2024. 'A Pathology Foundation Model for Cancer Diagnosis and Prognosis Prediction'. *Nature*, September. https://doi.org/10.1038/s41586-024-07894-z.
Wang, Yi Kan, Ludmila Tydlitatova, Jeremy D. Kunz, Gerard Oakley, Ran A. Godrich, Matthew C. H. Lee, Chad Vanderbilt, et al. 2024. 'Screen Them All: High-Throughput Pan-Cancer Genetic and Phenotypic Biomarker Screening from H&E Whole Slide Images'. arXiv. http://arxiv.org/abs/2408.09554.
Wang, Yuxuan, Yueqian Wang, Dongyan Zhao, Cihang Xie, and Zilong Zheng. 2024. 'VideoHallucer: Evaluating Intrinsic and Extrinsic Hallucinations in Large Video-Language Models'. arXiv. http://arxiv.org/abs/2406.16338.
Wu, Jiayang, Wensheng Gan, Zefeng Chen, Shicheng Wan, and Philip S. Yu. 2023. 'Multimodal Large Language Models: A Survey'. arXiv. http://arxiv.org/abs/2311.13165.
Wu, Tongtong, Linhao Luo, Yuan-Fang Li, Shirui Pan, Thuy-Trang Vu, and Gholamreza Haffari. 2024. 'Continual Learning for Large Language Models: A Survey'. arXiv. http://arxiv.org/abs/2402.01364.
Wu, Yuanfeng, Shaojie Li, Zhiqiang Du, and Wentao Zhu. 2023a. 'BROW: Better featuRes fOr Whole Slide Image Based on Self-Distillation'. arXiv. http://arxiv.org/abs/2309.08259.
———. 2023b. 'BROW: Better featuRes fOr Whole Slide Image Based on Self-Distillation'. arXiv. https://doi.org/10.48550/arXiv.2309.08259.
Xiang, Jinxi, Xiyue Wang, Xiaoming Zhang, Yinghua Xi, Feyisope Eweje, Yijiang Chen, Yuchen Li, et al. 2025. 'A Vision–Language Foundation Model for Precision Oncology'. *Nature*, January. https://doi.org/10.1038/s41586-024-08378-w.
Xu, Hanwen, Naoto Usuyama, Jaspreet Bagga, Sheng Zhang, Rajesh Rao, Tristan Naumann, Cliff Wong, et al. 2024. 'A Whole-Slide Foundation Model for Digital Pathology from Real-World Data'. *Nature* 630 (8015): 181–88. https://doi.org/10.1038/s41586-024-07441-w.
Xu, Kesi, Lea Goetz, and Nasir Rajpoot. 2024. 'On Generalisability of Segment Anything Model for Nuclear Instance Segmentation in Histology Images'. arXiv. http://arxiv.org/abs/2401.14248.
Xu, Yingxue, Yihui Wang, Fengtao Zhou, Jiabo Ma, Shu Yang, Huangjing Lin, Xin Wang, et al. 2024. 'A Multimodal Knowledge-Enhanced Whole-Slide Pathology Foundation Model'. arXiv. http://arxiv.org/abs/2407.15362.
Yang, Weikai, Mengchen Liu, Zheng Wang, and Shixia Liu. 2024. 'Foundation Models Meet Visualizations: Challenges and Opportunities'. *Computational Visual Media* 10 (3): 399–424. https://doi.org/10.1007/s41095-023-0393-x.
Yang, Zhaochang, Ting Wei, Ying Liang, Xin Yuan, Ruitian Gao, Yujia Xia, Jie Zhou, Yue Zhang, and Zhangsheng Yu. 2024. 'A Foundation Model for Generalizable Cancer Diagnosis and Survival Prediction from Histopathological Images'. https://doi.org/10.1101/2024.05.16.594499.
Yu, Jiahui, Zirui Wang, Vijay Vasudevan, Legg Yeung, Mojtaba Seyedhosseini, and Yonghui Wu. 2022. 'CoCa: Contrastive Captioners Are Image-Text Foundation Models'. arXiv. https://doi.org/10.48550/arXiv.2205.01917.
Zhang, Jingwei, Ke Ma, Saarthak Kapse, Joel Saltz, Maria Vakalopoulou, Prateek Prasanna, and Dimitris Samaras. 2023. 'SAM-Path: A Segment Anything Model for Semantic Segmentation in Digital Pathology'. In *Medical Image Computing and Computer Assisted Intervention – MICCAI 2023 Workshops*, edited by M. Emre Celebi, Md Sirajus Salekin, Hyunwoo Kim, Shadi Albarqouni, Catarina Barata, Allan Halpern, Philipp Tschandl, et al., 14393:161–70. Lecture Notes in Computer Science. Cham: Springer Nature Switzerland. https://doi.org/10.1007/978-3-031-47401-9_16.

Zhang, Xingxuan, Jiansheng Li, Wenjing Chu, Junjia Hai, Renzhe Xu, Yuqing Yang, Shikai Guan, Jiazheng Xu, and Peng Cui. 2024. ‘On the Out-Of-Distribution Generalization of Multimodal Large Language Models’. arXiv. http://arxiv.org/abs/2402.06599.

Zhang, Yunkun, Jin Gao, Mu Zhou, Xiaosong Wang, Yu Qiao, Shaoting Zhang, and Dequan Wang. 2023. ‘Text-Guided Foundation Model Adaptation for Pathological Image Classification’. In *Medical Image Computing and Computer Assisted Intervention – MICCAI 2023*, edited by Hayit Greenspan, Anant Madabhushi, Parvin Mousavi, Septimiu Salcudean, James Duncan, Tanveer Syeda-Mahmood, and Russell Taylor, 14224:272–82. Lecture Notes in Computer Science. Cham: Springer Nature Switzerland. https://doi.org/10.1007/978-3-031-43904-9_27.

Zhao, Zehui, Laith Alzubaidi, Jinglan Zhang, Ye Duan, and Yuantong Gu. 2024. ‘A Comparison Review of Transfer Learning and Self-Supervised Learning: Definitions, Applications, Advantages and Limitations’. *Expert Systems with Applications* 242 (May):122807. https://doi.org/10.1016/j.eswa.2023.122807.

Zheng, Sunyi, Xiaonan Cui, Yuxuan Sun, Jingxiong Li, Honglin Li, Yunlong Zhang, Pingyi Chen, Xueping Jing, Zhaoxiang Ye, and Lin Yang. 2024. ‘Benchmarking PathCLIP for Pathology Image Analysis’. *Journal of Imaging Informatics in Medicine*, July. https://doi.org/10.1007/s10278-024-01128-4.

Zhong, Wanjun, Ruixiang Cui, Yiduo Guo, Yaobo Liang, Shuai Lu, Yanlin Wang, Amin Saied, Weizhu Chen, and Nan Duan. 2023. ‘AGIEval: A Human-Centric Benchmark for Evaluating Foundation Models’. arXiv. http://arxiv.org/abs/2304.06364.

Zhou, Jinghao, Chen Wei, Huiyu Wang, Wei Shen, Cihang Xie, Alan Yuille, and Tao Kong. 2021. ‘Image BERT Pre-Training with Online Tokenizer’. In . https://openreview.net/forum?id=ydopy-e6Dg.

Zhou, Qifeng, Wenliang Zhong, Yuzhi Guo, Michael Xiao, Hehuan Ma, and Junzhou Huang. 2024. ‘PathM3: A Multimodal Multi-Task Multiple Instance Learning Framework for Whole Slide Image Classification and Captioning’. arXiv. http://arxiv.org/abs/2403.08967.

Zhou, Xiao, Xiaoman Zhang, Chaoyi Wu, Ya Zhang, Weidi Xie, and Yanfeng Wang. 2024. ‘Knowledge-Enhanced Visual-Language Pretraining for Computational Pathology’. arXiv. http://arxiv.org/abs/2404.09942.

Zimmermann, Eric, Eugene Vorontsov, Julian Viret, Adam Casson, Michal Zelechowski, George Shaikovski, Neil Tenenholtz, et al. 2024. ‘Virchow 2: Scaling Self-Supervised Mixed Magnification Models in Pathology’. arXiv. http://arxiv.org/abs/2408.00738.